\documentstyle[jair,twoside,11pt,theapa,graphicx,amsthm,subfigure,times,epstopdf]{article}

\jairheading{27}{2006}{335-380}{05/06}{11/06}
\ShortHeadings{Anytime Point-Based Approximations for Large POMDPs}
{Pineau, Gordon \& Thrun}
\firstpageno{335}

\newtheorem{lemma}{Lemma}

\begin{document}

\title{Anytime Point-Based Approximations for Large POMDPs}

\author{\name Joelle Pineau \email jpineau@cs.mcgill.ca \\
       \addr School of Computer Science\\ McGill University\\ Montr\'{e}al QC, H3A 2A7 CANADA\\
       \name Geoffrey Gordon \email ggordon@cs.cmu.edu \\
       \addr Machine Learning Department\\ Carnegie Mellon University\\ Pittsburgh PA,  15232 USA\\
       \name Sebastian Thrun \email thrun@stanford.edu \\
       \addr Computer Science Department\\ Stanford University\\ Stanford CA, 94305 USA\\
}

\maketitle

\def\argmax{\mathop{\rm argmax}}
\def\argmin{\mathop{\rm argmin}}
\def\suchthat{\mathop{\rm such\;that}}
\newtheorem{theorem}{Theorem}[section]
\newtheorem{definition}{Definition}[section]
\newtheorem{example}{Example}[section]
\newtheorem*{remark}{Remark}
\sloppy

\begin{abstract}
The Partially Observable Markov Decision Process has long been recognized as a rich framework for real-world planning and control problems, especially in robotics. However exact solutions in this framework are typically computationally intractable for all but the smallest problems.  A well-known technique for speeding up POMDP solving involves performing value backups at specific belief points, rather than over the entire belief simplex. The efficiency of this approach, however, depends greatly on the selection of points.  This paper presents a set of novel techniques for selecting informative belief points which work well in practice. The point selection procedure is combined with point-based value backups to form an effective anytime POMDP algorithm called \textit{Point-Based Value Iteration} (PBVI). The first aim of this paper is to introduce this algorithm and present a theoretical analysis justifying the choice of belief selection technique. The second aim of this paper is to provide a thorough empirical comparison between PBVI and other state-of-the-art POMDP methods, in particular the Perseus algorithm, in an effort to highlight their similarities and differences. Evaluation is performed using both standard POMDP domains and realistic robotic tasks.

\end{abstract}

\section{Introduction}
\label{introduction}

The concept of planning has a long tradition in the AI literature~\cite{fikes71,chapman87,mcallester91,penberthy92,blum97}.  Classical planning is generally concerned with agents which operate in environments that are fully observable, deterministic, finite, static, and discrete. While these techniques are able to solve increasingly large state-space problems, the basic assumptions of classical planning---full observability, static environment, deterministic actions---make these unsuitable for most robotic applications.

Planning under uncertainty aims to improve robustness by explicitly reasoning about the type of uncertainty that can arise.
The Partially Observable Markov Decision Process (POMDP)~\cite{astrom65,sondik71,monahan82,white91,lovejoy91a,kaelbling98,boutilier99} has emerged as possibly the most general representation for (single-agent) planning under uncertainty. The POMDP supersedes other frameworks in terms of representational power simply because it combines the most essential features for planning under uncertainty.

First, POMDPs handle uncertainty in both \textit{action effects} and \textit{state observability}, whereas many other frameworks handle neither of these, and some handle only stochastic action effects. To handle partial state observability, plans are expressed over \textit{information states}, instead of world states, since the latter ones are not directly observable. The space of information states is the space of all beliefs a system might have regarding the world state. Information states are easily calculated from the measurements of noisy and imperfect sensors. In POMDPs, information states are typically represented by probability distributions over world states.

Second, many POMDP algorithms form plans by optimizing a \textit{value function}. This is a powerful approach to plan optimization, since it allows one to numerically trade off between alternative ways to satisfy a goal, compare actions with different costs/rewards, as well as plan for multiple interacting goals. While value function optimization is used in other planning approaches---for example Markov Decision Processes (MDPs)~\cite{bellman57}---POMDPs are unique in expressing the value function over information states, rather than world states.

Finally, whereas classical and conditional planners produce a sequence of actions, POMDPs produce a full \textit{policy} for action selection, which prescribes the choice of action for any possible information state. By producing a universal plan, POMDPs alleviate the need for re-planning, and allow fast execution. Naturally, the main drawback of optimizing a universal plan is the computational complexity of doing so. This is precisely what we seek to alleviate with the work described in this paper

Most known algorithms for \textit{exact} planning in POMDPs operate by optimizing the value function over \textit{all} possible information states (also known as \textit{beliefs}). These algorithms can run into the well-known curse of dimensionality, where the dimensionality of planning problem is directly related to the number of states~\cite{kaelbling98}. But they can also suffer from the lesser known curse of history, where the number of belief-contingent plans increases exponentially with the planning horizon. In fact, exact POMDP planning is known to be PSPACE-complete, whereas propositional planning is only NP-complete~\cite{littman96}. As a result, many POMDP domains with only a few states, actions and sensor observations are computationally intractable.

A commonly used technique for speeding up POMDP solving involves selecting a finite set of belief points and performing value backups on this set~\cite{sondik71,cheng88,lovejoy91,hauskrecht00,zhang01}. While the usefulness of belief point updates is well acknowledged, how and when these backups should be applied has not been thoroughly explored.

This paper describes a class of \textit{Point-Based Value Iteration} (PBVI) POMDP approximations where the value function is estimated based strictly on point-based updates. In this context, the choice of points is an integral part of the algorithm, and our approach interleaves value backups with steps of belief point selection. One of the key contributions of this paper is the presentation and analysis of a set of heuristics for selecting informative belief points. These range from a naive version that combines point-based value updates with random belief point selection, to a sophisticated algorithm that combines the standard point-based value update with an estimate of the error bound between the approximate and exact solutions to select belief points. Empirical and theoretical evaluation of these techniques reveals the importance of taking distance between points into consideration when selecting belief points. The result is an approach which exhibits good performance with very few belief points (sometimes less than the number of states), thereby overcoming the curse of history.

The PBVI class of algorithms has a number of important properties, which are discussed at greater length in the paper:
\begin{itemize}
\item \textbf{Theoretical guarantees}. We present a bound on the error of the value function obtained by point-based approximation, with respect to the exact solution. This bound applies to a number of point-based approaches, including our own PBVI, Perseus~\cite{spaan05}, and others.
\item \textbf{Scalability}. We are able to handle problems on the order of $10^3$ states, which is an order of magnitude larger than problems solved by more traditional POMDP techniques. The empirical performance is evaluated extensively in realistic robot tasks, including a search-for-missing-person scenario.
\item \textbf{Wide applicability}. The approach makes few assumptions about the nature or structure of the domain. The PBVI framework does assume known discrete state/ action/observation spaces and a known model (i.e., state-to-state transitions, observation probabilities, costs/rewards), but no additional specific structure (e.g., constrained policy class, factored model).
\item \textbf{Anytime performance}. An anytime solution can be achieved by gradually alternating phases of belief point selection and phases of point-based value updates. This allows for an effective trade-off between planning time and solution quality. 
\end{itemize}

While PBVI has many important properties, there are a number of other recent POMDP approaches which exhibit competitive performance~\cite{braziunas04,poupart03,smith04,spaan05}. We provide an overview of these techniques in the later part of the paper. We also provide a comparative evaluation of these algorithms and PBVI using standard POMDP domains, in an effort to guide practitioners in their choice of algorithm. One of the algorithms, Perseus~\cite{spaan05}, is most closely related to PBVI both in design and in performance. We therefore provide a direct comparison of the two approaches using a realistic robot task, in an effort to shed further light on the comparative strengths and weaknesses of these two approaches.

The paper is organized as follows. Section~\ref{sec_pomdp} begins by exploring the basic concepts in POMDP solving, including representation, inference, and exact planning. Section~\ref{sec_pbvi} presents the general anytime PBVI algorithm and its theoretical properties. Section~\ref{sec_belief} discusses novel strategies to select good belief points. Section~\ref{sec_results} presents an empirical comparison of POMDP algorithms using standard simulation problems. Section~\ref{sec_robot} pursues the empirical evaluation by tackling complex robot domains and directly comparing PBVI with Perseus. Finally, Section~\ref{sec_related} surveys a number of existing POMDP approaches that are closely related to PBVI.

\section{Review of POMDPs}
\label{sec_pomdp}

Partially Observable Markov Decision Processes provide a general planning and decision-making framework for acting optimally in partially observable domains. They are well-suited to a great number of real-world problems where decision-making is required despite prevalent uncertainty. They generally assume a complete and correct world model, with stochastic state transitions, imperfect state tracking, and a reward structure. Given this information, the goal is to find an action strategy which maximizes expected reward gains. This section first establishes the basic terminology and essential concepts pertaining to POMDPs, and then reviews optimal techniques for POMDP planning.

\subsection{Basic POMDP Terminology}
\label{sec_pomdp_components}

Formally, a POMDP is defined by six distinct quantities,
denoted $\{S, A, Z, T, O, R\}$. The first three of these are:
\begin{itemize}
\item \emph{States}. The state of the world is denoted $s$, with
the finite set of all states denoted by $S=\{s_0, s_1, \ldots\}$. The state
at time $t$ is denoted $s_t$, where $t$ is a discrete time index.
The state is not directly observable in POMDPs, where an agent can
only compute a belief over the state space $S$.
\item \emph{Observations}. To infer a belief regarding the world's
state $s$, the agent can take sensor measurements.  The set of all
measurements, or observations, is denoted $Z=\{z_0, z_1,
\ldots\}$. The observation at time $t$ is denoted
$z_t$. Observation $z_t$ is usually an incomplete projection of the
world state $s_t$, contaminated by sensor noise.
\item \emph{Actions}. To act in the world, the agent is given a finite set of
actions, denoted $A=\{a_0, a_1, \ldots\}$. Actions stochastically affect the state
of the world.  Choosing the right action as a function of history is the core problem in
POMDPs.
\end{itemize}

Throughout this paper, we assume that states, actions and observations are discrete and finite.
For mathematical convenience, we also assume that actions and
observations are alternated over time.

To fully define a POMDP, we have to specify the probabilistic laws
that describe state transitions and observations. These laws are given
by the following distributions:
\begin{itemize}
\item The \emph{state transition probability distribution},
\begin{eqnarray}
T(s,a,s') &:=& Pr(s_t=s' \mid  s_{t-1}=s, a_{t-1}=a) \;\; \forall t,
\end{eqnarray}
is the probability of transitioning to state $s'$, given that the
agent is in state $s$ and selects action $a$, for any $(s,a,s')$.
Since $T$ is a conditional probability distribution, we have $\sum_{s'
\in S}T(s,a,s')=1, \forall (s,a)$. As our notation suggests, $T$ is
time-invariant.
\item The \emph{observation probability distribution},
\begin{eqnarray}
O(s,a,z) &:=& Pr(z_t=z \mid  s_{t-1}=s, a_{t-1}=a) \;\; \forall t,
\end{eqnarray}
is the probability that the agent will perceive observation $z$ upon
executing action $a$ in state $s$. This conditional probability is
defined for all $(s,a,z)$ triplets, for which $\sum_{z \in
Z}O(s,a,z)=1, \forall (s,a)$. The probability function $O$ is also time-invariant.
\end{itemize}

Finally, the objective of POMDP planning is to optimize
action selection, so the agent is given a reward function describing its performance:
\begin{itemize}
\item The \emph{reward function}.  $R(s,a):S\times A\longrightarrow \Re$,
assigns a numerical value quantifying the utility of performing action $a$ when in state $s$. We assume the reward is bounded, $R_{min}<R<R_{max}$. The goal of the
agent is to collect as much reward as possible over time.  More precisely, it wants to maximize the sum:
\begin{eqnarray}
E[\sum_{t = t_0}^T \gamma^{t-t_0} r_t],
\label{eqn_expectation}
\end{eqnarray}
where $r_t$ is the reward at time $t$, $E[\;]$ is the mathematical
expectation, and $\gamma$ where $0\leq\gamma<1$ is a \textit{discount
factor}, which ensures that the sum in Equation~\ref{eqn_expectation}
is finite.
\end{itemize}

These items together, the states $S$, actions $A$, observations
$Z$, reward $R$, and the probability distributions, $T$
and $O$, define the probabilistic world model that underlies
each POMDP\@.

\subsection{Belief Computation}
\label{sec_pomdp_tracking}

POMDPs are instances of Markov processes, which implies that
the current world state, $s_t$, is sufficient to predict the future, independent of the past $\{s_0, s_1, ..., s_{t-1}\}$.
The key characteristic that sets POMDPs apart from many other
probabilistic models (such as MDPs) is the fact that the state $s_t$ is
not directly observable. Instead, the agent can only perceive
observations $\{z_1,\ldots,z_t\}$, which convey incomplete information
about the world's state.

Given that the state is not directly observable, the agent can instead maintain a
complete trace of all observations and all actions it ever
executed, and use this to select its actions. The action/observation trace is known as a \textit{history}. We formally define
\begin{eqnarray}
h_t &:=& \{a_0, z_1, \ldots, z_{t-1}, a_{t-1}, z_{t}\}
\end{eqnarray}
to be the history at time $t$.

This history trace can get very long as time goes on.
A well-known fact is that this history does not need to be represented explicitly, but can
instead be summarized via a \textit{belief distribution}~\cite{astrom65},
which is the following posterior probability distribution:
\begin{eqnarray}
b_t(s)& :=& Pr(s_t=s\mid z_t,a_{t-1},z_{t-1}, \ldots , a_0, b_0).
\end{eqnarray}

This of course requires knowing the \emph{initial state probability distribution}:
\begin{eqnarray}
b_0(s) &:=& Pr(s_0=s),
\end{eqnarray}
which defines the probability that the domain is in state $s$ at time $t=0$.
It is common either to specify this initial belief as
part of the model, or to give it only to the runtime system which tracks
beliefs and selects actions.  For our work, we will assume that this
initial belief (or a set of possible initial beliefs) are available to
the planner.

Because the belief distribution $b_t$ is a sufficient statistic for the history, 
it suffices to condition the selection of actions on $b_t$, instead of on the ever-growing sequence of past
observations and actions.  Furthermore, the belief $b_t$ at time $t$
is calculated \textit{recursively}, using only the belief one time step
earlier, $b_{t-1}$, along with the most recent action $a_{t-1}$ and
observation $z_t$.

We define the belief update equation, $\tau()$, as:
\begin{eqnarray}
\tau(b_{t-1}, a_{t-1}, z_t) & = & b_t(s') \nonumber\\
 &=& \frac{\displaystyle \sum_{s'} O(s',a_{t-1},z_t) \;T(s,a_{t-1},s')\;b_{t-1}(s) }{ Pr(z_t | b_{t-1}, a_{t-1}) }
\label{eqn_belief}
\end{eqnarray}
where the denominator is a normalizing constant.

This equation is equivalent to the decades-old Bayes filter~\cite{jazwinski70},
and is commonly applied in the context of hidden Markov
models~\cite{rabiner89}, where it is known as the forward algorithm. Its continuous generalization forms the basis
of Kalman filters~\cite{kalman60}.

It is interesting to consider the nature of belief distributions.  Even for
finite state spaces, the belief is a continuous quantity. It is defined over a simplex describing
the space of all distributions over the state space $S$.  For very
large state spaces, calculating the belief update (Eqn~\ref{eqn_belief}) can
be computationally challenging.  Recent research has led to efficient
techniques for belief state computation that exploit structure of the
domain~\cite{dean88,boyen98,poupart00,thrun00a}. However, by far the
most complex aspect of POMDP planning is the generation of a policy
for action selection, which is described next.  For
example in robotics, calculating beliefs over state spaces with $10^6$
states is easily done in real-time~\shortcite{burgard99}. In contrast, calculating
optimal action selection policies exactly appears to be infeasible for
environments with more than a few dozen states~\cite{kaelbling98}, not directly because of the size of the state space, but because of the complexity of the optimal policies. Hence we assume throughout this paper that the belief can be computed accurately, and instead focus on the problem of finding good approximations to the optimal policy.

\subsection{Optimal Policy Computation}
\label{sec_pomdp_VI}

The central objective of the POMDP perspective is to compute a \emph{policy} for
selecting actions. A policy is of the form:
\begin{eqnarray}
\pi(b)&\longrightarrow& a,
\end{eqnarray}
where $b$ is a belief distribution and $a$ is the action chosen by
the policy $\pi$.

Of particular interest is the notion of
\emph{optimal policy}, which is a policy that maximizes the expected
future discounted cumulative reward:
\begin{eqnarray}
\pi^*(b_{t_0}) &=& \argmax_{\pi} E_{\pi} \left[ \sum_{t = t_0}^{T} \gamma^{t-t_0} r_t \Bigg | b_{t_0} \right] .
\end{eqnarray}

There are two distinct but interdependent reasons why computing an optimal policy is challenging.
The more widely-known reason is the so-called curse of dimensionality: in a problem with $n$
physical states, $\pi$ is defined over all belief states in an
$(n-1)$-dimensional continuous space.
The less-well-known reason is the curse of history: POMDP solving is in many ways like
a search through the space of possible POMDP histories. It starts by searching over short histories (through which it can select the best short policies),
and gradually considers increasingly long histories.  Unfortunately the number of distinct
possible action-observation histories grows exponentially with the planning horizon.

The two curses---dimensionality and history---often act independently: planning complexity can
grow exponentially with horizon even in problems with only a few
states, and problems with a large number of physical states may still
only have a small number of relevant histories. Which curse is predominant depends both on the
problem at hand, and the solution technique. For example, the belief point methods that are the focus
of this paper specifically target the curse of history, leaving themselves vulnerable to the curse
of dimensionality. Exact algorithms on the other hand typically suffer far more from the curse of
history. The goal is therefore to find techniques that offer the best balance between both.

We now describe a straightforward approach to finding optimal policies by~\citeA{sondik71}.
The overall idea is to apply multiple iterations of dynamic programming,
to compute increasingly more accurate values for each
belief state $b$. Let $V$ be a value function that maps
belief states to values in $\Re$. Beginning with
the initial value function:
\begin{eqnarray}
V_{0}(b)&=&\max_a  \sum_{s \in S} R(s,a)b(s),
\end{eqnarray}
then the $t$-th value function is constructed  from the
$(t-1)$-th by the following recursive equation:
\begin{eqnarray}
V_{t}(b)&=&\max_a \left[ \sum_{s \in S} R(s,a)b(s) + \gamma \sum_{z \in Z} Pr(z \mid a,b) V_{t-1}(\tau(b,a,z)) \right],
\label{eqn_valIt}
\end{eqnarray}
where $\tau(b,a,z)$ is the belief
updating function defined in Equation~\ref{eqn_belief}.
This value function update maximizes the expected sum of all
(possibly discounted) future pay-offs the agent receives in
the next $t$ time steps, for any belief state $b$.
Thus, it produces a policy that is optimal under the planning horizon $t$.
The optimal policy can also be directly extracted from the previous-step
value function:
\begin{eqnarray}
\pi_{t}^*(b) &=& \argmax_a  \left[ \sum_{s \in S} R(s,a)b(s) + \gamma \sum_{z \in Z} Pr(z \mid a,b) V_{t-1}(\tau(b,a,z)) \right].
\end{eqnarray}

\citeA{sondik71} showed that the value function at any finite horizon $t$ can be expressed by a set of vectors:
$\Gamma_t=\{\alpha_0, \alpha_1, \ldots, \alpha_m\}$. Each $\alpha$-vector represents an $|S|$-dimensional
hyper-plane, and defines the value function over a
bounded region of the belief:
\begin{eqnarray}
V_t(b) &=& \max_{\alpha \in \Gamma_t}\sum_{s \in S} \alpha(s)b(s).
\label{eqn_max_alpha}
\end{eqnarray}

In addition, each $\alpha$-vector is associated
with an action, defining the best immediate policy assuming optimal
behavior for the following $(t-1)$ steps (as defined respectively by
the sets $\{V_{t-1},...,V_{0}\}$).

The $t$-horizon solution set, $\Gamma_t$, can be computed as follows. First, we rewrite Equation~\ref{eqn_valIt} as:
\begin{eqnarray}
\hspace{-8mm}
V_t(b) & = & \max_{a \in A} \left[ \sum_{s \in S} R(s,a) b(s) +  \gamma
\sum_{z \in Z} \max_{\alpha \in \Gamma_{t-1}} \sum_{s \in S} \sum_{s' \in S}
T(s,a,s') O(s', a, z) \alpha(s') b(s)  \right]. \label{eqn_backup}
\end{eqnarray}
Notice that in this representation of $V_t(b)$, the nonlinearity in the term $P(z|a,b)$ from Equation~\ref{eqn_valIt}
cancels out the nonlinearity in the term $\tau(b,a,z)$, leaving a linear
function of b(s) inside the max operator.

The value $V_t(b)$ cannot be computed directly for each belief $b \in B$ (since there are infinitely many beliefs), but the corresponding set $\Gamma_t$ can be generated through a sequence of operations on the set $\Gamma_{t-1}$.

The first operation is to generate intermediate sets $\Gamma_t^{a,*}$ and $\Gamma_t^{a,z}, \forall a \in A, \forall z \in Z$ \emph{(Step 1)}:
\begin{eqnarray}
\Gamma_t^{a,*} & \leftarrow & \alpha^{a,*}(s) = R(s,a) \label{eqn_projection}\\
\Gamma_t^{a,z} & \leftarrow & \alpha^{a,z}_i(s) = \gamma \sum_{s' \in S} T(s,a,s') O(s',a,z) \alpha_i(s'), \forall \alpha_i \in \Gamma_{t-1} \nonumber
\end{eqnarray}
where each $\alpha^{a,*}$ and $\alpha_i^{a,z}$ is once again an $|S|$-dimensional hyper-plane.

Next we create $\Gamma_t^{a}$ ($\forall a \in A$), the cross-sum over observations\footnote{The symbol $\oplus$ denotes the cross-sum operator. A cross-sum operation is defined over two sets, $A=\{a_1, a_2, \ldots, a_m\}$ and $B=\{b_1, b_2, \ldots, b_n\}$, and produces a third set, $C=\{a_1+b_1, a_1+b_2, \ldots, a_1+b_n, a_2+b_1, a_2+b_2, \ldots, \ldots, a_m+b_n\}$.}, which includes one $\alpha^{a,z}$ from each $\Gamma_t^{a,z}$ \emph{(Step 2)}:
\begin{eqnarray}
\Gamma_t^a & = & \Gamma_t^{a,*} + \Gamma_t^{a,z_1} \oplus \Gamma_t^{a,z_2} \oplus \ldots
\label{eqn_crosssum}
\end{eqnarray}

Finally we take the union of $\Gamma_t^a$ sets \emph{(Step 3)}:
\begin{eqnarray}
\Gamma_t & = & \cup_{a \in A} \; \Gamma_t^a.
\label{eqn_union}
\end{eqnarray}

This forms the pieces of the backup solution at horizon $t$.
The actual value function $V_t$ is extracted from the set $\Gamma_t$ as described in Equation~\ref{eqn_max_alpha}.

Using this approach, bounded-time POMDP problems with finite state,
action, and observation spaces can be solved exactly given a choice of the horizon $T$.
If the environment is such that the agent might not be able to bound the planning horizon in
advance, the policy $\pi_{t}^*(b)$ is an approximation to the
optimal one whose quality improves in expectation with the planning
horizon $t$ (assuming $0 \leq \gamma < 1$).

As mentioned above, the value function $V_t$ can be extracted directly from the set $\Gamma_t$.
An important aspect of this algorithm (and of all optimal finite-horizon POMDP solutions) is that the value function is guaranteed to be a piecewise
linear, convex, and continuous function of the belief~\cite{sondik71}. The piecewise-linearity and continuous properties are a direct result of the fact that $V_t$ is composed of finitely many linear $\alpha$-vectors. The convexity property is a result of the maximization operator (Eqn~\ref{eqn_max_alpha}).  It is worth pointing out that the intermediate sets $\Gamma_t^{a,z}$, $\Gamma_t^{a,*}$ and $\Gamma_t^{a}$ also represent functions of the belief which are composed entirely of linear segments.  This property holds for the intermediate representations because they incorporate the expectation over observation probabilities (Eqn~\ref{eqn_projection}).

In the worst case, the exact value update procedure described could require time doubly
exponential in the planning horizon $T$~\cite{kaelbling98}.
To better understand the complexity of the exact update, let $|S|$ be the number of states, $|A|$ the number of 
actions, $|Z|$ the number of observations,
and $|\Gamma_{t-1}|$ the number of $\alpha$-vectors in the previous
solution set. Then Step 1 creates $|A|\,|Z|\,|\Gamma_{t-1}|$
projections and Step 2 generates $|A|\,|\Gamma_{t-1}|^{|Z|}$
cross-sums. So, in the worst case, the new solution requires:
\begin{eqnarray} 
|\Gamma_{t}|=O(|A||\Gamma_{t-1}|^{|Z|})
\label{eqn_complexity}
\end{eqnarray}
$\alpha$-vectors to represent the value function at horizon $t$; these can be
computed in time $O(|S|^2|A|\,|\Gamma_{t-1}|^{|Z|})$.

It is often the case that a vector in $\Gamma_t$ will be
completely dominated by another vector over the entire belief simplex:
\begin{equation}
\alpha_i\cdot b<\alpha_j \cdot b, \;\;\; \forall b.
\label{eqn_prune}
\end{equation}

\begin{figure}[ht]
\centerline{\includegraphics[height=3.5cm]{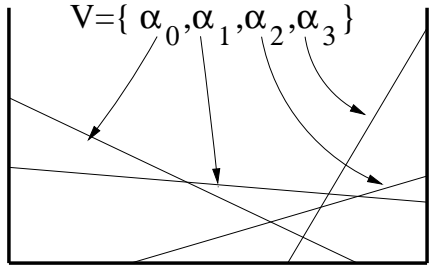}}
\caption{POMDP value function representation}
\label{fig_alpha_prune}
\end{figure}

Similarly, a vector may be fully dominated by a set of other vectors (e.g., $\alpha_2$ in Fig.~\ref{fig_alpha_prune} is dominated by the combination of $\alpha_1$ and $\alpha_3$).
This vector can then be pruned away without affecting the solution. 
Finding dominated vectors can be expensive. Checking whether a single vector
is dominated requires solving a linear program with $|S|$ variables and $|\Gamma_t|$ constraints. Nonetheless it can be
time-effective to apply pruning after each iteration to prevent an explosion of the solution size.
In practice, $|\Gamma_{t}|$ often appears to grow singly exponentially in $t$,
given clever mechanisms for pruning unnecessary linear functions.
This enormous computational complexity has long been a key impediment toward applying POMDPs to
practical problems. 

\subsection{Point-Based Value Backup}
\label{sec_pomdp_point}

Exact POMDP solving, as outlined above, optimizes the value function over all beliefs.  Many approximate POMDP solutions, including the PBVI approach proposed in this paper, gain computational advantage by applying value updates at specific (and few) belief points, rather than over all beliefs~\cite{cheng88,zhang01,poon01}. These approaches differ significantly (and to great consequence) in how they select the belief points, but once a set of points is selected, the procedure for updating their value is standard.  We now describe the procedure for updating the value function at a set of known belief points.

As in Section~\ref{sec_pomdp_VI}, the value function update is implemented as a sequence of operations on a set of $\alpha$-vectors. If we assume that we are only interested in updating the value function at a fixed set of belief points, $B=\{b_0, b_1, ..., b_q\}$, then it follows that the value function will contain at most one $\alpha$-vector for each belief point.  The point-based value function is therefore represented by the corresponding set $\{\alpha_0, \alpha_1, \ldots, \alpha_q\}$.

Given a solution set $\Gamma_{t-1}$, we simply modify the exact backup operator (Eqn~\ref{eqn_backup}) such that only one $\alpha$-vector per belief point is maintained.
The point-based backup now gives an $\alpha$-vector which is valid over a region around $b$. It assumes that the other belief points in that region have the same action
choice and lead to the same facets of $V_{t-1}$ as the point $b$. This is the key idea behind all algorithms presented in this paper, and the reason for the large computational savings associated with this class of algorithms.

To obtain solution set $\Gamma_t$ from the previous set $\Gamma_{t-1}$, we begin once again by generating intermediate sets $\Gamma_t^{a,*}$ and $\Gamma_t^{a,z}, \forall a \in A, \forall z \in Z$ (exactly as in Eqn~\ref{eqn_projection}) \emph{(Step 1)}:
\begin{eqnarray}
\Gamma_t^{a,*} & \leftarrow & \alpha^{a,*}(s) = R(s,a) \label{eqn_projection2} \\
\Gamma_t^{a,z} & \leftarrow & \alpha^{a,z}_i(s) = \gamma \sum_{s' \in S} T(s,a,s') O(s',a,z) \alpha_i(s'), \forall \alpha_i \in \Gamma_{t-1}. \nonumber
\end{eqnarray}

Next, whereas performing an exact value update requires a cross-sum operation (Eqn~\ref{eqn_crosssum}), by operating over a finite set of points, we can instead use a simple summation. We construct $\Gamma_t^{a}, \forall a \in A$ \emph{(Step 2)}:
\begin{eqnarray}
\Gamma_t^{a} & \leftarrow & \alpha^a_b = \Gamma_t^{a,*} + \sum_{z \in Z} \argmax_{\alpha \in \Gamma_t^{a,z}} (\sum_{s \in S}\alpha(s) b(s)), \forall b \in B.
\label{eqn_crosssum2}
\end{eqnarray}

Finally, we find the best action for each belief point \emph{(Step 3)}:
\begin{eqnarray}
\alpha_b & = & \argmax_{\Gamma_t^{a}, \forall a \in A} (\sum_{s \in S} \Gamma_t^{a}(s) b(s)), \;\;\; \forall b \in B.\\
\Gamma_t & = & \cup_{b \in B} \; \alpha_b
\label{eqn_union2}
\end{eqnarray}

While these operations preserve only the best $\alpha$-vector at each belief point $b \in B$, an estimate of the value function at \textit{any} belief in the simplex (including $b \notin B$) can be extracted from the set $\Gamma_t$ just as before:
\begin{eqnarray}
V_t(b) &=& \max_{\alpha \in \Gamma_t}\sum_{s \in S} \alpha(s)b(s).
\end{eqnarray}

To better understand the complexity of updating the value of a set of points $B$, let $|S|$ be the number of states, $|A|$ the number of actions, $|Z|$ the number of observations, and $|\Gamma_{t-1}|$ the number of $\alpha$-vectors in the previous solution set. As with an exact update, Step 1 creates $|A|\,|Z|\,|\Gamma_{t-1}|$ projections (in time $|S|^2\,|A|\,|Z|\,|\Gamma_{t-1}|$). Steps 2 and 3 then reduce this set to at most $|B|$ components (in time $|S|\,|A|\,|\Gamma_{t-1}|\,|Z|\,|B|$). Thus, a full point-based value update takes only polynomial time, and even more crucially, the size of the solution set $\Gamma_t$ remains constant at every iteration.  The point-based value backup algorithm is summarized in Table~\ref{table_backup}.

\begin{table}[ht]
\begin{center}
\begin{tabular}{lr}
\hline
$\Gamma_t$=BACKUP($B$, $\Gamma_{t-1}$) & 1\\
\hspace{4mm} For each action $a \in A$ & 2\\
\hspace{8mm}   For each observation $z \in Z$ & 3\\
\hspace{12mm}     For each solution vector $\alpha_i \in \Gamma_{t-1}$ & 4\\
\hspace{16mm}       $\alpha_i^{a,z}(s)= \gamma \sum_{s' \in S} T(s,a,s') O(s',a,z) \alpha_i(s'), \forall s \in S$ & 5\\
\hspace{12mm}     End & 6\\
\hspace{12mm}     $\Gamma_t^{a,z} = \cup_i \; \alpha_i^{a,z}$ & 7\\
\hspace{8mm}   End & 8\\
\hspace{4mm} End & 9\\
\hspace{4mm} $\Gamma_t = \emptyset$ & 10\\
\hspace{4mm} For each belief point $b \in B$ & 11\\
\hspace{8mm}   $\alpha_b = \argmax_{a \in A} \left[ \sum_{s \in S} R(s,a)b(s) + \sum_{z \in Z} \max_{\alpha \in \Gamma_t^{a,z}} \left[ \sum_{s \in S} \alpha(s) b(s) \right] \right] $ & 12\\
\hspace{8mm}   If($\alpha_b \notin \Gamma_t$) & 13\\
\hspace{12mm}    $\Gamma_t = \Gamma_t \cup \alpha_b$ & 14\\
\hspace{4mm} End & 15\\
\hspace{4mm} Return $\Gamma_t$ & 16\\
\hline
\end{tabular}
\caption{Point-based value backup}
\label{table_backup}
\end{center}
\end{table}

Note that the algorithm as outlined in Table~\ref{table_backup} includes a trivial pruning step (lines 13-14), whereby we refrain from adding to $\Gamma_t$ any vector already included in it. As a result, it is often the case that $|\Gamma_{t}| \leq |B|$. This situation arises whenever multiple nearby belief points support the same vector. This pruning step can be computed rapidly (without solving linear programs) and is clearly advantageous in terms of reducing the set $\Gamma_t$.

The point-based value backup is found in many POMDP solvers, and in general serves to improve estimates of the value function.  It is also an integral part of the PBVI framework.

\section{Anytime Point-Based Value Iteration}
\label{sec_pbvi}

We now describe the algorithmic framework for our new class of fast approximate POMDP algorithms called \textit{Point-Based Value Iteration} (PBVI).
PBVI-class algorithms offer an anytime solution to large-scale discrete POMDP domains.  The key to achieving an anytime solution is to interleave two main components: the point-based update described in Table~\ref{table_backup} and steps of belief set selection. The approximate value function we find is guaranteed to have bounded error (compared to the optimal) for any discrete POMDP domain.

The current section focuses on the overall anytime algorithm and its theoretical properties, independent of the belief point selection process.  Section~\ref{sec_belief} then discusses in detail various novel techniques for belief point selection.

The overall PBVI framework is simple. We start with a (small) initial set of belief points to
which are applied a first series of backup operations. The set of belief points is then grown,
a new series of backup operations are applied to all belief points (old and new), and so on, until a satisfactory solution is obtained. By interleaving value backup iterations with expansions of the
belief set, PBVI offers a range of solutions, gradually trading off computation time and solution quality.

The full algorithm is presented in Table~\ref{table_pbvi}. The algorithm accepts as input an initial belief point set ($B_{Init}$), an initial value ($\Gamma_0$), the number of desired expansions ($N$), and the planning horizon ($T$).  A common choice for $B_{Init}$ is the initial belief $b_0$; alternately, a larger set could be used, especially in cases where sample trajectories are available. The initial value, $\Gamma_0$, is typically set to be purposefully low (e.g., $\alpha_0(s)=\frac{R_{min}}{1-\gamma}, \forall s \in S$). When we do this, we can show that the point-based solution is always be a lower-bound on the exact solution~\cite{lovejoy91}. This follows from the simple observation that failing to compute an $\alpha$-vector can only lower the value function.

For problems with a finite horizon, we run $T$ value backups between each expansion of the belief set.  In infinite-horizon problems, we select the horizon $T$ so that
\begin{equation}
\gamma^T \left[R_{\rm max}-R_{\rm min} \right] < \epsilon, \nonumber
\end{equation}
where $R_{\rm max} = \max_{s,a}R(s,a)$ and $ R_{\rm min} = \min_{s,a}R(s,a)$.

The complete algorithm terminates once a fixed number of expansions ($N$) have been completed. Alternately, the algorithm could terminate once the value function approximation reaches a given performance criterion. This is discussed further below.

The algorithm uses the BACKUP routine described in Table~\ref{table_backup}. We can assume for the moment that the EXPAND subroutine (line 8) selects belief points at random.  This performs reasonably well for small problems where it is easy to achieve good coverage of the entire belief simplex. However it scales poorly to larger domains where exponentially many points are needed to guarantee good coverage of the belief simplex. More sophisticated approaches to selecting belief points are presented in Section~\ref{sec_belief}. Overall, the PBVI framework described here offers a simple yet flexible approach to solving large-scale POMDPs.

\begin{table}[htb!]
\begin{center}
\begin{tabular}{lr}
\hline
$\Gamma$=PBVI-MAIN($B_{Init}$, $\Gamma_0$, $N$, $T$) & 1\\
\hspace{4mm} $B$=$B_{Init}$ & 2\\
\hspace{4mm} $\Gamma=\Gamma_0$ & 3\\
\hspace{4mm} For $N$ expansions & 4\\
\hspace{8mm}   For $T$ iterations & 5\\
\hspace{12mm}     $\Gamma=$BACKUP($B$,$\Gamma$) & 6\\
\hspace{8mm}   End & 7\\
\hspace{8mm}   $B_{new}=$EXPAND($B$,$\Gamma$) & 8\\
\hspace{8mm}   $B=B \cup B_{new}$ & 9\\
\hspace{4mm} End & 10\\
\hspace{4mm} Return $\Gamma$ & 11\\
\hline
\end{tabular}
\caption{Algorithm for Point-Based Value Iteration (PBVI)}
\label{table_pbvi}
\end{center}
\end{table}

For any belief set $B$ and horizon $t$, the algorithm in Table~\ref{table_pbvi} will produce an estimate of the value function, denoted $V^B_t$. We now show that the error between $V^B_t$ and the optimal value function $V^*$ is bounded. The bound depends on how densely $B$ samples the belief simplex $\Delta$; with denser sampling, $V^B_t$ converges to $V^*_t$, the $t$-horizon optimal solution, which in turn has bounded error with respect to $V^*$, the optimal solution.
So cutting off the PBVI iterations at any sufficiently large horizon, we can show that the difference between
$V^B_t$ and the optimal infinite-horizon $V^*$ is not too large. The overall error in PBVI is bounded, according to the triangle inequality, by:
\begin{equation}
\|V^B_t-V^*\|_{\infty} \leq \|V^B_t-V^*_t\|_\infty + \|V^*_t-V^*\|_\infty. \nonumber
\end{equation}
The second term is bounded by $\gamma^t\|V^*_0-V^*\|$~\cite{bertsekas96}.
The remainder of this section states and proves a bound on the first term, which we denote $\epsilon_t$.

Begin by assuming that $H$ denotes an exact value backup, and $\tilde H$ denotes the PBVI backup. Now define $\epsilon(b)$ to be the error introduced at a specific belief $b \in \Delta$ by performing one iteration of point-based backup:
$$\epsilon(b) = |\tilde HV^B(b)-HV^B(b)|_{\infty}.$$
Next define $\epsilon$ to be the maximum total error introduced by doing one iteration of point-based backup:
\begin{eqnarray}
\epsilon & = & |\tilde HV^B-HV^B|_{\infty} \nonumber\\
 & = & \max_{b \in \Delta} \epsilon(b).\nonumber
\end{eqnarray}
Finally define the density $\delta_B$ of a set of belief points $B$ to be
the maximum distance from any belief in the simplex $\Delta$ to a belief in set $B$.  More precisely:
\begin{equation}
\delta_B = \max_{b'\in\Delta}\min_{b\in B} \|b-b'\|_1. \nonumber
\end{equation}
Now we can prove the following lemma:
\begin{lemma}
The error introduced in PBVI when performing \textbf{one iteration} of value backup over $B$, instead of over $\Delta$, is bounded by
$$\epsilon \leq \frac{(R_{\rm max}-R_{\rm min})\delta_B}{1-\gamma}$$

\label{lemma_pruning}
\end{lemma}
\textbf{Proof:} Let $b'\in\Delta$ be the point where PBVI makes its
worst error in value update, and $b\in B$ be the closest (1-norm) sampled belief
to $b'$.  Let $\alpha$ be the vector that is maximal at
$b$, and $\alpha'$ be the vector that would be maximal at $b'$.  By failing to include
$\alpha'$ in its solution set, PBVI makes an error of at most $\alpha' \cdot b' - \alpha
\cdot b'$.  On the other hand, since $\alpha$ is maximal at $b$, then
$\alpha'\cdot b \leq \alpha\cdot b$.  So,
\[
\begin{array}{rcll}
\epsilon &\leq& \alpha' \cdot b' - \alpha \cdot b'\vspace{.5ex}
& \vspace{.5ex}\\
&=& \alpha' \cdot b' - \alpha \cdot b' + (\alpha'\cdot b -
\alpha'\cdot b) & \mbox{Add zero} \vspace{.5ex}\\
&\leq& \alpha' \cdot b' - \alpha \cdot b' + \alpha\cdot b -
\alpha'\cdot b & \mbox{Assume $\alpha$ is optimal at $b$} \vspace{.5ex}\\
&=& (\alpha'-\alpha)\cdot(b'-b)\vspace{.5ex} & \mbox {Re-arrange the terms}\\
&\leq& \|\alpha'-\alpha\|_\infty \|b'-b\|_1 & \mbox{By H\"older inequality}
\vspace{.5ex}\\
&\leq& \|\alpha'-\alpha\|_\infty \delta_B & \mbox{By definition of
$\delta_B$} \vspace{.5ex}\\
&\leq& \frac{(R_{\rm max}-R_{\rm min}) \delta_B}{1-\gamma} & \mbox{}
\end{array}
\]

The last inequality holds because each $\alpha$-vector represents the reward achievable starting from some
state and following some sequence of actions and observations.  
Therefore the sum of rewards must fall between $\frac{R_{\rm min}}{1-\gamma}$ and $\frac{R_{\rm max}}{1-\gamma}$.
\qed

Lemma~\ref{lemma_pruning} states a bound on the approximation error introduced by one iteration of point-based value updates within the PBVI framework. We now look at the bound over multiple value updates.

\begin{theorem}
For any belief set $B$ and any horizon $t$, the error of the PBVI
algorithm $\epsilon_t = \|V^B_t - V^*_t\|_\infty$ is bounded by
$$\epsilon_t \leq \frac{(R_{\rm max}-R_{\rm min})\delta_B}{(1-\gamma)^2}$$
\label{thm_bound}
\end{theorem}
\textbf{Proof:}

\[
\begin{array}{rcll}
\epsilon_t & = & ||V^B_t-V^*_t||_{\infty} & 
\vspace{.5ex}\\
 & = & ||\tilde HV^B_{t-1}-HV^*_{t-1}||_{\infty} & \mbox{By definition of $\tilde H$} \vspace{.5ex}\\
 & \leq & ||\tilde HV^B_{t-1}-HV^B_{t-1}||_{\infty} + ||HV^B_{t-1}-HV^*_{t-1}||_{\infty} & \mbox{By triangle inequality}
\vspace{.5ex}\\
 & \leq & \frac{(R_{\rm max}-R_{\rm min})\delta_B}{1-\gamma} + ||HV^B_{t-1}-HV^*_{t-1}||_{\infty} &
 \mbox{By lemma~\ref{lemma_pruning}} \vspace{.5ex}\\
 & \leq & \frac{(R_{\rm max}-R_{\rm min})\delta_B}{1-\gamma}+ \gamma ||V^B_{t-1}-V^*_{t-1}||_{\infty} &
\mbox{By contraction of exact value backup} \vspace{.5ex}\\
 & = & \frac{(R_{\rm max}-R_{\rm min})\delta_B}{1-\gamma} + \gamma \epsilon_{t-1} & \mbox{By definition of
$\epsilon_{t-1}$} \vspace{.5ex}\\
 & \leq & \frac{(R_{\rm max}-R_{\rm min})\delta_B}{(1-\gamma)^2} &
 \mbox{By sum of a geometric series} \qed
\end{array}
\]

The bound described in this section depends on how densely $B$ samples the
belief simplex $\Delta$. In the case where not all beliefs are reachable,
PBVI does not need to sample all of $\Delta$ densely, but can
replace $\Delta$ by the set of reachable beliefs $\bar\Delta$ (Fig.~\ref{fig_beliefTree}).
The error bounds and convergence results hold on $\bar\Delta$. We simply need to re-define $b' \in \bar\Delta$ in lemma~\ref{lemma_pruning}.

As a side note, it is worth pointing out that because PBVI makes no assumption regarding the initial value function $V^B_0$, the point-based solution $V^B$ is not guaranteed to improve with the addition of belief points.  Nonetheless, the theorem presented in this section shows that the \textit{bound on the error} between $V_t^B$ (the point-based solution) and $V^*$ (the optimal solution) is guaranteed to decrease (or stay the same) with the addition of belief points. In cases where $V_t^B$ is initialized pessimistically (e.g., $V_0^B(s)=\frac{R_{min}}{1-\gamma}, \forall s \in S$, as suggested above), then $V_t^B$ will improve (or stay the same) with each value backup and addition of belief points.

This section has thus far skirted the issue of belief point selection, however the bound presented in this section clearly argues in favor of dense sampling over the belief simplex.  While randomly selecting points according to a uniform distribution may eventually accomplish this, it is generally inefficient, in particular for high dimensional cases. Furthermore, it does not take advantage of the fact that the error bound holds for dense sampling over \textit{reachable} beliefs.  Thus we seek more efficient ways to generate belief points than at random over the entire simplex. This is the issue explored in the next section.

\section{Belief Point Selection}
\label{sec_belief}

In section~\ref{sec_pbvi}, we outlined the prototypical PBVI algorithm, while conveniently avoiding the question of how and when belief points should be selected.
There is a clear trade-off between including fewer beliefs (which would favor fast planning over good performance), versus including many beliefs (which would slow down planning, but ensure a better bound on performance).  This brings up the question of \textit{how many} belief points should be included. However the number of points is not the only consideration. It is likely that some collections of belief points (e.g., those frequently encountered) are more likely to produce a good value function than others. This brings up the question of \textit{which} beliefs should be included.

A number of approaches have been proposed in the literature. For example, some exact value function approaches use linear programs to identify points where the value function needs to be further improved~\cite{cheng88,littman96,zhang01}, however this is typically very expensive. The value function can also be approximated by learning the value at regular points, using a fixed-resolution~\cite{lovejoy91}, or variable-resolution~\cite{zhou01} grid. This is less expensive than solving LPs, but can scales poorly as the number of states increases. Alternately, one can use heuristics to generate grid-points~\cite{hauskrecht00,poon01}. This tends to be more scalable, though significant experimentation is required to establish which heuristics are most useful.

This section presents five heuristic strategies for selecting belief points, from fast and naive random sampling, to increasingly more sophisticated stochastic simulation techniques.  The most effective strategy we propose is one that carefully selects points that are likely to have the largest impact in reducing the error bound (Theorem~\ref{thm_bound}).

Most of the strategies we consider focus on selecting \textit{reachable} beliefs, rather than getting uniform coverage over the entire belief simplex. Therefore it is useful to begin this discussion by looking at how reachability is assessed.

\begin{figure}[!ht]
\centerline{\includegraphics[height=5.5cm]{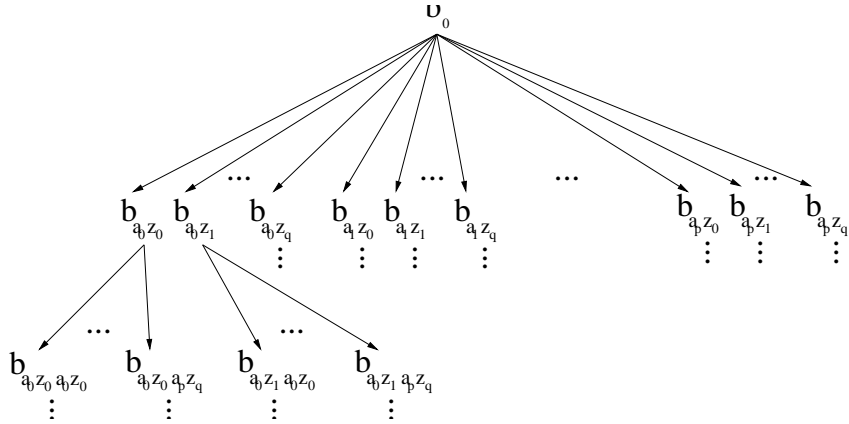}}
\caption{The set of reachable beliefs}
\label{fig_beliefTree}
\end{figure}

While some exact POMDP value iteration solutions are optimal for \textit{any} initial belief, PBVI (and other related techniques) assume a known initial belief $b_0$. As shown in Figure~\ref{fig_beliefTree}, we can use the initial belief to build a tree of reachable beliefs. In this representation, each path through the tree corresponds to a sequence in belief space, and increasing depth corresponds to an increasing plan horizon. When selecting a set of belief points for PBVI, including all reachable beliefs would guarantee optimal performance (conditioned on the initial belief), but at the expense of computational tractability, since the set of reachable beliefs, $\bar\Delta$, can grow exponentially with the planning horizon. Therefore, it is best to select a subset $B \subset \bar\Delta$ which is sufficiently small for computational tractability, but sufficiently large for good value function approximation.\footnote{All strategies discussed below assume that the belief point set, $B$, approximately doubles in size on each belief expansion. This ensures that the number of rounds of value iteration is logarithmic (in the final number of belief points needed). Alternately, each strategy could be used (with very little modification) to add a fixed number of new belief points, but this may require many more rounds of value iteration. Since value iteration is much more expensive than belief computation, it seems appropriate to double the size of $B$ at each expansion.}

In domains where the initial belief is not known (or not unique), it is still possible to use reachability analysis by sampling a few initial beliefs (or using a set of known initial beliefs) to seed multiple reachability trees.

We now discuss five strategies for selecting belief points, each of which can be used within the PBVI framework to perform expansion of the belief set.

\subsection{Random Belief Selection (\textbf{RA})}
\label{sec_belief_ra}

The first strategy is also the simplest. It consists of sampling belief points from a uniform distribution over the entire belief simplex. To sample over the simplex, we cannot simply sample each $b(s)$ independently over $[0,1]$ (this would violate the constraint that $\sum_s b(s)=1$). Instead, we use the algorithm described in Table~\ref{table_expansionRA}~\cite<see>[for more details including proof of uniform coverage]{devroye86}.

\begin{table}[!ht]
\begin{center}
\begin{tabular}{lr}
\hline
$B_{new}$=EXPAND$_{\textrm{RA}}$($B$, $\Gamma$) & 1\\
\hspace{4mm} $B_{new}$= $B$ & 2\\
\hspace{4mm} Foreach $b \in B$ & 3 \\
\hspace{8mm}     $S$ := number of states & 4 \\
\hspace{8mm}     For $i=0:S$ & 5 \\
\hspace{12mm}        $b_{tmp}[i]$=rand$_{\textrm{uniform}}$(0,1) & 6 \\
\hspace{8mm}     End & 7 \\
\hspace{8mm}     Sort $b_{tmp}$ in ascending order & 8 \\
\hspace{8mm}     For $i=1:S-1$ & 9 \\
\hspace{12mm}        $b_{new}[i]$=$b_{tmp}[i+1]-b_{tmp}[i]$ & 10 \\
\hspace{8mm}     End & 11 \\
\hspace{8mm}     $B_{new} = B_{new} \cup {b_{new}}$ & 12 \\
\hspace{4mm} End & 13 \\
\hspace{4mm} Return $B_{new}$ & 14 \\
\hline
\end{tabular}
\caption{Algorithm for belief expansion with random action selection}
\label{table_expansionRA}
\end{center}
\end{table}

This random point selection strategy, unlike the other strategies presented below, does not focus on reachable beliefs. For this reason, we do not necessarily advocate this approach. However we include it because it is an obvious choice, it is by far the simplest to implement, and it has been used in related work by~\citeA{hauskrecht00} and~\citeA{poon01}.
In smaller domains (e.g., $<$20 states), it performs reasonably well, since the belief simplex is relatively low-dimensional. In large domains (e.g., 100+ states), it cannot provide good coverage of the belief simplex with a reasonable number of points, and therefore exhibits poor performance.  This is demonstrated in the experimental results presented in Section~\ref{sec_results}.

All of the remaining belief selection strategies make use of the belief tree (Figure~\ref{fig_beliefTree}) to focus on \textit{reachable} beliefs, rather than trying to cover the entire belief simplex.

\subsection{Stochastic Simulation with Random Action (\textbf{SSRA})}
\label{sec_belief_ssra}

To generate points along the belief tree, we use a technique called stochastic simulation. It involves running single-step forward trajectories from belief points already in $B$. Simulating a single-step forward trajectory for a given $b \in B$ requires selecting an action and observation pair $(a,z)$, and then computing the new belief $\tau(b,a,z)$ using the Bayesian update rule (Eqn~\ref{eqn_belief}).  In the case of \textit{Stochastic Simulation with Random Action} (SSRA), the action selected for forward simulation is picked (uniformly) at random from the full action set. Table~\ref{table_expansionSSRA} summarizes the belief expansion procedure for SSRA.  First, a state $s$ is drawn from the belief distribution $b$. Second, an action $a$ is drawn at random from the full action set. Next, a posterior state $s'$ is drawn from the transition model $T(s,a,s')$. Finally, an observation $z$ is drawn from the observation model $O(s',a,z)$.  Using the triple $(b,a,z)$, we can calculate the new belief $b_{new}=\tau(b,a,z)$ (according to Equation~\ref{eqn_belief}), and add to the set of belief points $B_{new}$.

\begin{table}[!ht]
\begin{center}
\begin{tabular}{lr}
\hline
$B_{new}$=EXPAND$_{\textrm{SSRA}}$($B$, $\Gamma$) & 1\\
\hspace{4mm} $B_{new}$= $B$ & 2\\
\hspace{4mm} Foreach $b \in B$ & 3\\
\hspace{8mm}     $s$=rand$_{\textrm{multinomial}}$($b$) & 4\\
\hspace{8mm}	 $a$=rand$_{\textrm{uniform}}$($A$) & 5\\
\hspace{8mm}     $s'$=rand$_{\textrm{multinomial}}$($T(s,a,\cdot)$) & 6\\
\hspace{8mm}     $z$=rand$_{\textrm{multinomial}}$($O(s',a,\cdot)$) & 7\\
\hspace{8mm}     $b_{new}$ = $\tau(b, a, z)$ (see Eqn~\ref{eqn_belief})& 8\\
\hspace{8mm}     $B_{new}$ = $B_{new} \cup b_{new}$ & 9\\
\hspace{4mm} End & 10\\
\hspace{4mm} Return $B_{new}$ & 11\\
\hline
\end{tabular}
\caption{Algorithm for belief expansion with random action selection}
\label{table_expansionSSRA}
\end{center}
\end{table}

This strategy is better than picking points at random (as described above), because it restricts $B_{new}$ to the belief tree (Fig.~\ref{fig_beliefTree}). However this belief tree is still very large, especially when the branching factor is high, due to large numbers of actions/observations.  By being more selective about which paths in the belief tree are explored, one can hope to effectively restrict the belief set further.

A similar technique for stochastic simulation was discussed by~\citeA{poon01}, however the belief set was initialized differently (not using $b_0$), and therefore the stochastic simulations were not restricted to the set of reachable beliefs.

\subsection{Stochastic Simulation with Greedy Action (\textbf{SSGA})}
\label{sec_belief_ssga}

The procedure for generating points using \textit{Stochastic Simulation with Greedy Action} (SSGA) is based on the well-known $\epsilon$\textit{-greedy} exploration strategy used in reinforcement learning~\cite{sutton98}.  This strategy is similar to the SSRA procedure, except that rather than choosing an action randomly, SSEA will choose the \textit{greedy} action (i.e., the current best action at the given belief $b$) with probability $1-\epsilon$, and will chose a random action with probability $\epsilon$ (we use $\epsilon=0.1$).  Once the action is selected, we perform a single-step forward simulation as in SSRA to yield a new belief point.
Table~\ref{table_expansionSSGA} summarizes the belief expansion procedure for SSGA.

\begin{table}[!ht]
\begin{center}
\begin{tabular}{lr}
\\
\hline
$B_{new}$=EXPAND$_{\textrm{SSGA}}$($B$, $\Gamma$) & 1\\
\hspace{4mm} $B_{new}$= $B$ & 2\\
\hspace{4mm} Foreach $b \in B$ & 3\\
\hspace{8mm}     $s$=rand$_{\textrm{multinomial}}$($b$) & 4\\
\hspace{8mm}     If rand$_{\textrm{uniform}}[0,1] < \epsilon$ & 5\\ 
\hspace{12mm}	   $a$=rand$_{\textrm{uniform}}$($A$) & 6\\
\hspace{8mm}     Else & 7\\
\hspace{12mm}	   $a$=$\argmax_{\alpha \in \Gamma} \sum_{s \in S} \alpha(s) b(s)$ & 8\\
\hspace{8mm}     End & 9  \\
\hspace{8mm}     $s'$=rand$_{\textrm{multinomial}}$($T(s,a,\cdot)$) & 10\\
\hspace{8mm}     $z$=rand$_{\textrm{multinomial}}$($O(s',a,\cdot)$) & 11\\
\hspace{8mm}     $b_{new}$ = $\tau(b, a, z)$ (see Eqn~\ref{eqn_belief})& 12\\
\hspace{8mm}     $B_{new}$ = $B_{new} \cup b_{new}$ & 13\\
\hspace{4mm} End & 14\\
\hspace{4mm} Return $B_{new}$ & 15\\
\hline
\end{tabular}
\caption{Algorithm for belief expansion with greedy action selection}
\label{table_expansionSSGA}
\end{center}
\end{table}

A similar technique, featuring stochastic simulation using greedy actions, was outlined by~\citeA
{hauskrecht00}. However in that case, the belief set included all extreme points of the belief simplex, and stochastic simulation was done from those extreme points, rather than from the initial belief.

\subsection{Stochastic Simulation with Exploratory Action (\textbf{SSEA})}
\label{sec_belief_ssea}

The error bound in Section~\ref{sec_pbvi} suggests that PBVI performs
best when its belief set is uniformly dense in the set of reachable
beliefs. The belief point strategies proposed thus far ignore this information.
The next approach we propose gradually expands $B$ by greedily choosing new reachable beliefs that improve the worst-case density.

Unlike SSRA and SSGA which select a single action to simulate the forward trajectory for a given $b \in B$, \textit{Stochastic Sampling with Exploratory Action} (SSEA) does a one step forward simulation with \textit{each action}, thus producing new beliefs $\{b_{a_0}, b_{a_1}, ...\}$.
However it does not accept all new beliefs $\{b_{a_0}, b_{a_1}, ...\}$, but rather calculates the $L_1$ distance between each $b_{a}$ and its closest neighbor in $B$. We then keep only that point $b_{a}$ that is farthest away from any point already in $B$. We use the $L_1$ norm to calculate distance between belief points to be consistent with the error bound in Theorem~\ref{thm_bound}. Table~\ref{table_expansionSSEA} summarizes the SSEA expansion procedure.

\begin{table}[!ht]
\begin{center}
\begin{tabular}{lr}
\hline
$B_{new}$=EXPAND$_{\textrm{SSEA}}$($B$, $\Gamma$) & 1\\
\hspace{4mm} $B_{new}$= $B$ & 2\\
\hspace{4mm} Foreach $b \in B$ & 3\\
\hspace{8mm}   Foreach $a \in A$ & 4\\
\hspace{12mm}     $s$=rand$_{\textrm{multinomial}}$($b$) & 5\\
\hspace{12mm}     $s'$=rand$_{\textrm{multinomial}}$($T(s,a,\cdot)$) & 6\\
\hspace{12mm}     $z$=rand$_{\textrm{multinomial}}$($O(s',a,\cdot)$) & 7\\
\hspace{12mm}     $b_a$=$\tau(b, a, z)$ (see Eqn~\ref{eqn_belief})& 8\\
\hspace{8mm}   End & 9\\
\hspace{8mm}   $b_{new}$ = $max_{a \in A}$ $min_{b' \in B_{new}}$ $\sum_{s \in S}$ $|b_a(s)-b'(s)|$ & 10\\
\hspace{8mm}   $B_{new}$ = $B_{new} \cup b_{new}$ (see Eqn~\ref{eqn_belief})& 11\\
\hspace{4mm} End & 12\\
\hspace{4mm} Return $B_{new}$ & 13\\
\hline
\end{tabular}
\caption{Algorithm for belief expansion with exploratory action selection}
\label{table_expansionSSEA}
\end{center}
\end{table}

\subsection{Greedy Error Reduction (\textbf{GER})}
\label{sec_belief_ger}

While the SSEA strategy above is able to improve the worst-case density of reachable beliefs, it does not directly minimize the expected error.  And while we would like to directly minimize the error, all we can measure is a bound on the error (Lemma~\ref{lemma_pruning}).  We therefore propose a final strategy which greedily adds the candidate beliefs that will most effectively reduce this error bound.  Our empirical results, as presented below, show that this strategy is the most successful one discovered thus far.

\begin{figure}[!ht]
\centerline{\includegraphics[height=5.5cm]{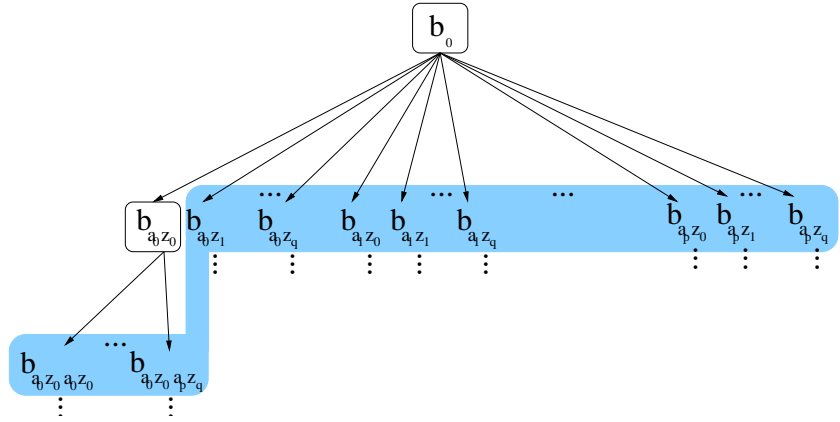}}
\caption{The set of reachable beliefs}
\label{fig_envelope}
\end{figure}

To understand how we expand the belief set in the GER strategy, it is useful to re-consider the belief tree, which we reproduce in Figure~\ref{fig_envelope}. 
Each node in the tree corresponds to a specific belief. We can divide these nodes into three sets. Set 1 includes those belief points already in $B$, in this case $b_0$ and $b_{a_0z_0}$. Set 2 contains those belief points that are immediate descendants of the points in $B$ (i.e., the nodes in the grey zone). These are the candidates from which we will select the new points to be added to $B$. We call this set the envelope (denoted $\bar{B}$). Set 3 contains all other reachable beliefs.

We need to decide which belief $b$ should be removed from the envelope $\bar{B}$ and added to the set of active belief points $B$.  Every point that is added to $B$ will improve our estimate of the value function.  The new point will reduce the error bounds (as defined in Section~\ref{sec_pbvi} for points that were already in $B$; however, the error bound for the new point itself might be quite large.  That means that the largest error bound for points in $B$ will not monotonically decrease; however, for a particular point in $B$ (such as the initial belief $b_0$) the error bound will be decreasing.

To find the point which will most reduce our error bound, we can look at the analysis of Lemma~\ref{lemma_pruning}.  Lemma~\ref{lemma_pruning} bounds the amount of additional error that a single point-based backup introduces.  Write $b'$ for the new belief which we are considering adding, and write $b$ for some belief which is already in $B$.  Write $\alpha$ for the value hyper-plane at $b$, and write $\alpha'$ for $b'$.  As the lemma points out, we have
$$\epsilon(b') \leq (\alpha'-\alpha) \cdot (b'-b)$$
When evaluating this error, we need to minimize over all $b \in B$. Also, since we do not know what $\alpha'$ will be until we have done some backups at $b'$, we make a conservative assumption and choose the worst-case value of $\alpha' \in [R_{\rm min}/(1-\gamma),R_{\rm max}/(1-\gamma)]^{|S|}$.  Thus, we can evaluate:
\begin{equation}
\epsilon(b') \leq \min_{b \in B} \sum_{s \in S} \left\{ \label{eqn_error}
\begin{array}{cc}
(\frac{R_{\rm max}}{1-\gamma}-\alpha(s))(b'(s)-b(s)) & b'(s)\geq b(s) \\
(\frac{R_{\rm min}}{1-\gamma}-\alpha(s))(b'(s)-b(s)) & b'(s) < b(s)
\end{array}
\right.
\end{equation}

While one could simply pick the candidate $b' \in \bar{B}$ which currently has the largest error bound,\footnote{We tried this, however it did not perform as well empirically as what we suggest in Equation~\ref{eqn_envelope1}, because it did not consider the probability of reaching that belief.} $\epsilon(b')$, this would ignore reachability considerations. Rather, we evaluate the error at each $b \in B$, by weighing the error of the fringe nodes by their reachability probability:
\begin{eqnarray}
\epsilon(b) & = & \max_{a \in A} \sum_{z \in Z} O(b,a,z) \;\; \epsilon(\tau(b,a,z)) \label{eqn_envelope1}\\
 & = & \max_{a \in A} \sum_{z \in Z} \left( \sum_{s \in S} \sum_{s' \in S} T(s,a,s') O(s',a,z) b(s) \right) \epsilon(\tau(b,a,z)), \nonumber
\end{eqnarray}
noting that $\tau(b,a,z) \in \bar{B}$, and $\epsilon(\tau(b,a,z))$ can be evaluated according to Equation~\ref{eqn_error}.

Using Equation~\ref{eqn_envelope1}, we find the existing point $b \in B$ with the largest error bound. We can now directly reduce its error by adding to our set one of its descendants.  We select the next-step belief $\tau(b,a,z)$ which maximizes error bound reduction:
\begin{eqnarray}
B & = & B \cup \tau(\tilde{b},\tilde{a},\tilde{z}), \label{eqn_envelope3}\\
& \textrm{where} &  \tilde{b}, \tilde{a}  :=  \argmax_{b \in B, a \in A} \sum_{z \in Z} O(b,a,z) \; \epsilon(\tau(b,a,z)) \label{eqn_envelope4}\\
& \hspace{4mm} & \tilde{z} := \argmax_{z \in Z} O(\tilde{b},\tilde{a},z) \; \epsilon(\tau(\tilde{b},\tilde{a},z)) \label{eqn_envelope5}
\end{eqnarray}

Table~\ref{table_expansionGER} summarizes the GER approach to belief point selection.

\begin{table}[!ht]
\begin{center}
\begin{tabular}{lr}
\hline
$B_{new}$=EXPAND$_{\textrm{GER}}$($B$, $\Gamma$) & 1\\
\hspace{4mm} $B_{new}$= $B$ & 2\\
\hspace{4mm} $N$=$|B|$ & 3\\
\hspace{4mm} For $i=1:N$ & 4\\
\hspace{8mm}   $\tilde{b}, \tilde{a}  :=  \argmax_{b \in B, a \in A} \sum_{z \in Z} O(b,a,z) \; \epsilon(\tau(b,a,z)) $ & 5\\
\hspace{8mm}   $\tilde{z} := \argmax_{z \in Z} O(\tilde{b},\tilde{a},z) \; \epsilon(\tau(\tilde{b},\tilde{a},z))$ & 6\\
\hspace{8mm}   $b_{new} = \tau(\tilde{b},\tilde{a},\tilde{z}) $ & 7\\
\hspace{8mm}   $B_{new} = B_{new} \cup b_{new}$ & 8 \\
\hspace{4mm} End & 9\\
\hspace{4mm} Return $B_{new}$ & 10\\
\hline
\end{tabular}
\caption{Algorithm for belief expansion}
\label{table_expansionGER}
\end{center}
\end{table}

The complexity of adding one new points with GER is $O(SAZB)$ (where $S$=\#states, $A$=\#actions, $Z$=\#observations, $B$=\#beliefs already selected). In comparison, a value backup (for one point) is $O(S^2AZB)$, and each point typically needs to be updated several times. As we point out in empirical results below, belief selection (even with GER) takes minimal time compared to value backup.

This concludes our presentation of belief selection techniques for the PBVI framework.
In summary, there are three factors to consider when picking a belief point: (1) how likely is it to occur? (2) how far is it from other belief points already selected? (3) what is the current approximate value for that point?  The simplest heuristic (RA) accounts for none of these, whereas some of the others (SSRA, SSGA, SSEA) account for one, and GER incorporates all three factors.

\subsection{Belief Expansion Example}
\label{sec_belief_example}

\begin{figure}[!ht]
\centerline{\includegraphics[height=1.5cm]{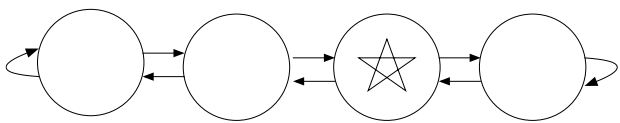}}
\caption{1D POMDP}
\label{fig_1d}
\end{figure}

We consider a simple example, shown in Figure~\ref{fig_1d}, to illustrate the difference between the various belief expansion techniques outlined above.  This 1D POMDP~\cite{littman96} has four states, one of which is the goal (indicated by the star). The two actions, left and right, have the expected (deterministic) effect. The goal state is fully observable (\textit{observation=goal}), while the other three states are aliased (\textit{observation=none}). A reward of $+1$ is received for being in the goal state, otherwise the reward is zero. We assume a discount factor of $\gamma=0.75$. The initial distribution is uniform over non-goal states, and the system resets to that distribution whenever the goal is reached.

\begin{figure}[!ht]
\centerline{\includegraphics[height=6cm]{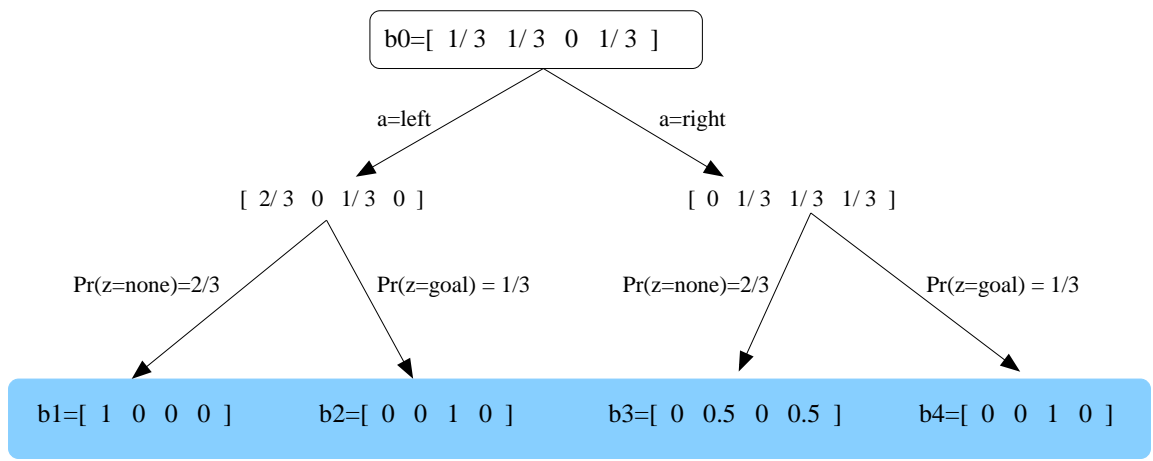}}
\caption{1D POMDP belief tree}
\label{fig_1d_envelope1}
\end{figure}

The belief set $B$ is always initialized to contain the initial belief $b_0$. Figure~\ref{fig_1d_envelope1} shows part of the belief tree, including the original belief set (top node), and its envelope (leaf nodes).  We now consider what each belief expansion method might do.

The \textbf{Random} heuristic can pick any belief point (with equal probability) from the entire belief simplex. It does not directly expand any branches of the belief tree, but it will eventually put samples nearby.

The \textbf{Stochastic Simulation with Random Action} has a $50\%$ chance of picking each action. Then, regardless of which action was picked, there's a 2/3 chance of seeing observation \textit{none}, and a 1/3 chance of seeing observation \textit{goal}.  As a result, the \textbf{SSRA} will select: $Pr(b_{new} = b1)=0.5*\frac{2}{3}$, $Pr(b_{new} = b2)=0.5*\frac{1}{3}$, $Pr(b_{new} = b3)=0.5*\frac{2}{3}$, $Pr(b_{new} = b4)=0.5*\frac{1}{3}$.

The \textbf{Stochastic Simulation with Greedy Action} first needs to know the policy at $b_0$.  A few iterations of point-based updates (Section~\ref{sec_pomdp_point}) applied to this initial (single point) belief set reveal that $\pi(b_0)=left$.\footnote{This may not be obvious to the reader, but it follows directly from the repeated application of equations~\ref{eqn_projection2}--\ref{eqn_union2}.} As a result, expansion of the belief will greedily select action $left$ with probability $1-\epsilon+\frac{\epsilon}{|A|} = 0.95$ (assuming $\epsilon=0.1$ and $|A|=2$). Action $right$ will be selected for belief expansion with probability $\frac{\epsilon}{|A|}=0.05$. Combining this along with the observation probabilities, we can tell that \textbf{SSGA} will expand as follows: $Pr(b_{new} = b1)=0.95*\frac{2}{3}$, $Pr(b_{new} = b2)=0.95*\frac{1}{3}$, $Pr(b_{new} = b3)=0.05*\frac{2}{3}$, $Pr(b_{new} = b4)=0.05*\frac{1}{3}$.

Predicting the choice of \textbf{Stochastic Simulation with Exploratory Action} is slightly more complicated.  Four cases can occur, depending on the outcomes of random forward simulation from $b_0$:
\begin{enumerate}
\item If action \textit{left} goes to $b_1$ ($Pr=2/3$) and action \textit{right} goes to $b_3$ ($Pr=2/3$), then $b_1$ will be selected because $||b_0-b_1||_1=4/3$ whereas $||b_0-b_3||_1=2/3$. This case will occur with $Pr=4/9$.
\item If action \textit{left} goes to $b_1$ ($Pr=2/3$) and action \textit{right} goes to $b_4$ ($Pr=1/3$), then $b_4$ will be selected because $||b_0-b_4||_1=2$. This case will occur with $Pr=2/9$.
\item If action \textit{left} goes to $b_2$ ($Pr=1/3$) and action \textit{right} goes to $b_3$ ($Pr=2/3$), then $b_2$ will be selected because $||b_0-b_2||_1=2$. This case will occur with $Pr=2/9$.
\item If action \textit{left} goes to $b_2$ ($Pr=1/3$) and action \textit{right} goes to $b_4$ ($Pr=1/3$), then either can be selected (since they are equidistant to $b_0$). In this case each $b_2$ and $b_4$ has $Pr=1/18$ of being selected.
\end{enumerate}
All told, $Pr(b_{new} = b1)=4/9$, $Pr(b_{new} = b2)=5/18$, $Pr(b_{new} = b3)=0$, $Pr(b_{new} = b4)=5/18$.

Now looking at belief expansion using \textbf{Greedy Error Reduction}, we need to compute the error $\epsilon(\tau(b_0,a,z)), \forall a,z$.  We consider Equation~\ref{eqn_error}: since $B$ has only one point, $b_0$, then necessarily $b=b_0$. To estimate $\alpha$, we apply multiple steps of value backup at $b_0$ and obtain $\alpha=[0.94 \; 0.94 \; 0.92 \; 1.74]$.  Using $b$ and $\alpha$ as such, we can now estimate the error at each candidate belief: $\epsilon(b_1)=2.93$, $\epsilon(b_2)=4.28$, $\epsilon(b_3)=1.20$, $\epsilon(b_4)=4.28$. Note that because $B$ has only one point, the dominating factor is their distance to $b_0$. Next, we factor in the observation probabilities, as in Eqns~\ref{eqn_envelope4}-\ref{eqn_envelope5}, which allows us to determine that $\tilde{a}=left$ and $\tilde{z}=none$, and therefore we should select $b_{new} = b_1$.

In summary, we note that SSGA, SSEA and GER all favor selecting $b_1$, whereas SSRA picks each option with equal probability (considering that $b_2$ and $b_4$ are actually the same).  In general, for a problem of this size, it is reasonable to expand the entire belief tree. Any of the techniques discussed here will be do this quickly, except RA which will not pick the exact nodes in the belief tree, but will select equally good nearby beliefs. This example is provided simply to illustrate the different choices made by each strategy.  

\section{A Review of Point-Based Approaches for POMDP Solving}
\label{sec_related}

The previous section describes a new class of point-based algorithms for POMDP solving. The idea of using point-based updates in POMDPs has been explored previously in the literature, and in this section we summarize the main results. For most of the approaches discussed below, the procedure for updating the value function at a given point remains unchanged (as outlined in Section~\ref{sec_pomdp_point}). Rather, the approaches are mainly differentiated by how the belief points are selected, and by how the updates are ordered.

\subsection{Exact Point-Based Algorithms}

Some of the earlier exact POMDP techniques use point-based backups to optimize the value function over limited regions of the belief simplex~\cite{sondik71,cheng88}. These techniques typically require solving multiple linear programs to find candidate belief points where the value function is sub-optimal, which can be an expensive operation. Furthermore, to guarantee that an exact solution is found, relevant beliefs must be generated systematically, meaning that all reachable beliefs must be considered. As a result, these methods typically cannot scale beyond a handful of states/actions/observations.

In work by~\citeA{zhang01}, point-based updates are interleaved with standard dynamic programming updates to further accelerate planning. In this case the points are not generated systematically, but rather backups are applied to both a set of \textit{witness} points and \textit{LP} points. The witness points are identified as a result of the standard dynamic programming updates, whereas the LP points are identified by solving linear programs to identify beliefs where the value has not yet been improved. Both of these procedures are significantly more expensive than the belief selection heuristics presented in this paper and results are limited to domains with at most a dozen states/actions/observations. Nonetheless this approach is guaranteed to converge to the optimal solution.

\subsection{Grid-Based Approximations}

There exists many approaches that approximate the value function using a finite set of belief points along with their values. These points are often distributed according to a grid pattern over the belief space, thus the name \textit{grid-based approximation}.  An interpolation-extrapolation rule specifies the value at non-grid points as a function of the value of neighboring grid-points. These approaches ignore the convexity of the POMDP value function.

Performing value backups over grid-points is relatively straightforward: dynamic programming updates as specified in Equation~\ref{eqn_valIt} can be adapted to grid-points for a simple polynomial-time algorithm. Given a set of grid points $G$, the value at each $b^G \in G$ is defined by:
\begin{eqnarray}
V(b^G)&=&\max_a \left[ \sum_{s \in S} b^G(s)R(s,a) + \gamma \sum_{z \in Z} Pr(z \mid a,b) V(\tau(b,a,z)) \right].
\end{eqnarray}
If $\tau(b,a,z)$ is part of the grid, then $V(\tau(b,a,z))$ is defined by the value backups. Otherwise, $V(\tau(b,a,z))$ is approximated using an interpolation rule such as:
\begin{eqnarray}
V(\tau(b,a,z) & = & \sum_{i=1}^{|G|}\lambda(i)V(b_i^G),
\end{eqnarray}
where $\lambda(i) \geq 0$ and $\sum_{i=1}^{|G|} \lambda(i) = 1$. This produces a convex combination over grid-points.
The two more interesting questions with respect to grid-based approximations are (1) how to calculate the interpolation function; and (2) how to select grid points.

In general, to find the interpolation that leads to the best value function approximation at a point $b$ requires solving the following linear program:
\begin{eqnarray}
\textrm{Minimize} & & \sum_{i=1}^{|G|}\lambda(i)V(b_i^G) \\
\textrm{Subject to} & & b=\sum_{i=1}^{|G|}\lambda(i)b_i^G \\
& & \sum_{i=1}^{|G|}\lambda(i)=1\\
& & 0 \leq \lambda(i) \leq 1, 1 \leq i \leq |G|.
\end{eqnarray}

Different approaches have been proposed to select grid points.
\citeA{lovejoy91} constructs a fixed-resolution regular grid over the entire belief space. A benefit is that value interpolations can be calculated quickly by considering only neighboring grid-points.  The disadvantage is that the number of grid points grows exponentially with the dimensionality of the belief (i.e., with the number of states).  A simpler approach would be to select random points over the belief space \cite{hauskrecht97a}. But this requires slower interpolation for estimating the value of the new points.  Both of these methods are less than ideal when the beliefs encountered are not uniformly distributed. In particular, many problems are characterized by dense beliefs at the edges of the simplex (i.e., probability mass focused on a few states, and most other states have zero probability), and low belief density in the middle of the simplex.  A distribution of grid-points that better reflects the actual distribution over belief points is therefore preferable.

Alternately,~\citeA{hauskrecht97a} also proposes using the corner points of the belief simplex (e.g., [1 0 0 \ldots ], [0 1 0 \ldots], \ldots, [0 0 0 \ldots 1]), and generating additional successor belief points through one-step stochastic simulations (Eqn~\ref{eqn_belief}) from the corner points. He also proposes an approximate interpolation algorithm that uses the values at $|S|-1$ critical points plus one non-critical point in the grid. An alternative approach is that by~\citeA{brafman97}, which builds a grid by also starting with the critical points of the belief simplex, but then uses a heuristic to estimate the usefulness of gradually adding intermediate points (e.g., $b_k=0.5b_i+0.5b_j$, for any pair of points). Both Hauskrecht's and Brafman's methods---generally referred to as \textit{non-regular grid approximations}---require fewer points than Lovejoy's regular grid approach. However the interpolation rule used to calculate the value at non-grid points is typically more expensive to compute, since it involves searching over all grid points, rather than just the neighboring sub-simplex.

\citeA{zhou01} propose a grid-based approximation that combines advantages from both regular and non-regular grids.  The idea is to sub-sample the regular fixed-resolution grid proposed by Lovejoy.  This gives a variable resolution grid since some parts of the beliefs can be more densely sampled than others and by restricting grid points to lie on the fixed-resolution grid the approach can guarantee fast value interpolation for non-grid points. Nonetheless, the algorithm often requires a large number of grid points to achieve good performance.

Finally,~\citeA{bonet02} proposes the first grid-based algorithm for POM\-DPs with $\epsilon$-optimality (for any $\epsilon>0$). This approach requires thorough coverage of the belief space such that every point is within $\delta$ of a grid-point. The value update for each grid point is fast to implement, since the interpolation rule depends only on the nearest neighbor of the one-step successor belief for each grid point (which can be pre-computed).  The main limitation is the fact that $\epsilon$-coverage of the belief space can only be attained by using exponentially many grid points. Furthermore, this method requires good coverage of the entire belief space, as opposed to the algorithms of Section~\ref{sec_belief}, which focus on coverage of the reachable beliefs.

\subsection{Approximate Point-Based Algorithms}

More similar to the PBVI-class of algorithms are those approaches that update both the value
and gradient at each grid point~\cite{lovejoy91,hauskrecht00,poon01}. These methods are able to preserve the piecewise linearity and convexity of the value function, and define a value function over the entire belief simplex.
Most of these methods use random beliefs, and/or require the inclusion of a large number of fixed beliefs such as the corners of the probability simplex.
In contrast, the PBVI-class algorithms we propose (with the exception of PBVI+RA) select only reachable beliefs, and in particular those belief points that improve the error bounds as quickly as possible.
The idea of using reachability analysis (also known as \textit{stochastic simulation}) to generate new points was explored by some of the earlier approaches~\cite{hauskrecht00,poon01}. However their analysis indicated that stochastic simulation was \textit{not} superior to random point placements. We re-visit this question (and conclude otherwise) in the empirical evaluation presented below.

More recently, a technique closely related to PBVI called Perseus has been proposed~\cite{vlassis04,spaan05}. Perseus uses point-based backups similar to the ones used in PBVI, but the two approaches differ in two ways. First, Perseus uses randomly generated trajectories through the belief space to select a set of belief points. This is in contrast to the belief-point selection heuristics outlined above for PBVI. Second, whereas PBVI systematically updates the value at all belief points at every epoch of value iteration, Perseus selects a subset of points to update at every epoch. The method used to select points is the following: points are randomly sampled one at a time and their value is updated. This continues until the value of \textit{all} points has been improved. The insight resides in observing that updating the $\alpha$-vector at one point often also improves the value estimate of other nearby points (which are then removed from the sampling set). This approach is conceptually simple and empirically effective.

The HSVI algorithm~\cite{smith04} is another point-based algorithm, which differs from PBVI both in how it picks belief points, and in how it orders value updates. It maintains a lower and an upper bound on the value function approximation, and uses it to select belief points. The updating of the upper bound requires solving linear programs and is generally the most expensive step. The ordering of value update is as follows: whenever a belief point is expanded from the belief tree, HSVI updates only the value of its direct ancestors (parents, grand-parents, etc., all the way back to the initial belief in the head node). This is in contrast to PBVI which performs a batch of belief point expansions, followed by a batch of value updates over all points. In other respects, HSVI and PBVI share many similarities: both offer anytime performance, theoretical guarantees, and scalability; finally the HSVI also takes reachability into account.  We will evaluate empirical differences between HSVI and PBVI in the next section.

Finally, the RTBSS algorithm~\cite{paquet05} offers an online version of point-based algorithms.  The idea is to construct a belief reachability tree similar to Figure~\ref{fig_beliefTree}, but using the current belief as the top node, and terminating the tree at some fixed depth $d$. The value at each node can be computed recursively over the finite planning horizon $d$. The algorithm can eliminate some subtrees by calculating a bound on their value, and comparing it to the value of other computed subtrees. RTBSS can in fact be combined with an offline algorithms such as PBVI, where the offline algorithm is used to pre-compute a lower bound on the exact value function; this can be used to increase subtree pruning, thereby increasing the depth of the online tree construction and thus also the quality of the solution. This online algorithm can yield fast results in very large POMDP domains. However the overall solution quality does not achieve the same error guarantees as the offline approaches.

\section{Experimental Evaluation}
\label{sec_results}

This section looks at a variety of simulated POMDP domains to evaluate the empirical performance of PBVI.
The first three domains---Tiger-grid, Hallway, Hallway2---are extracted from the established POMDP literature~\cite{cassandra99}. The fourth---Tag---was introduced in some of our earlier work as a new challenge for POMDP algorithms.

The first goal of these experiments is to establish the scalability of the PBVI framework; this is accomplished by showing that PBVI-type algorithms can successfully solve problems in excess of 800 states.  We also demonstrate that PBVI algorithms compare favorably to alternative approximate value iteration methods. Finally, following on the example of Section~\ref{sec_belief_example}, we study at a larger scale the impact of the belief selection strategy, which confirms the superior performance of the GER strategy.

\subsection{Maze Problems}
\label{sec_results_simulation}

There exists a set of benchmark problems commonly used to evaluate POMDP planning algorithms~\cite{cassandra99}. This section presents results demonstrating the performance of PBVI-class algorithms on some of those problems. While these benchmark problems are relatively small (at most 92 states, 5 actions, and 17 observations) compared to most robotics planning domains, they are useful from an analysis point of view and for comparison to previous work.

The initial performance analysis focuses on three well-known problems from the POMDP literature: Tiger-grid (also known as Maze33), Hallway, and Hallway2. All three are maze navigation problems of various sizes. The problems are fully described by~\citeA{littman95c}; parameterization is available from~\citeA{cassandra99}.

\begin{figure}[!ht]
\centering
\subfigure[Tiger-grid]{\includegraphics[height=5.5cm]{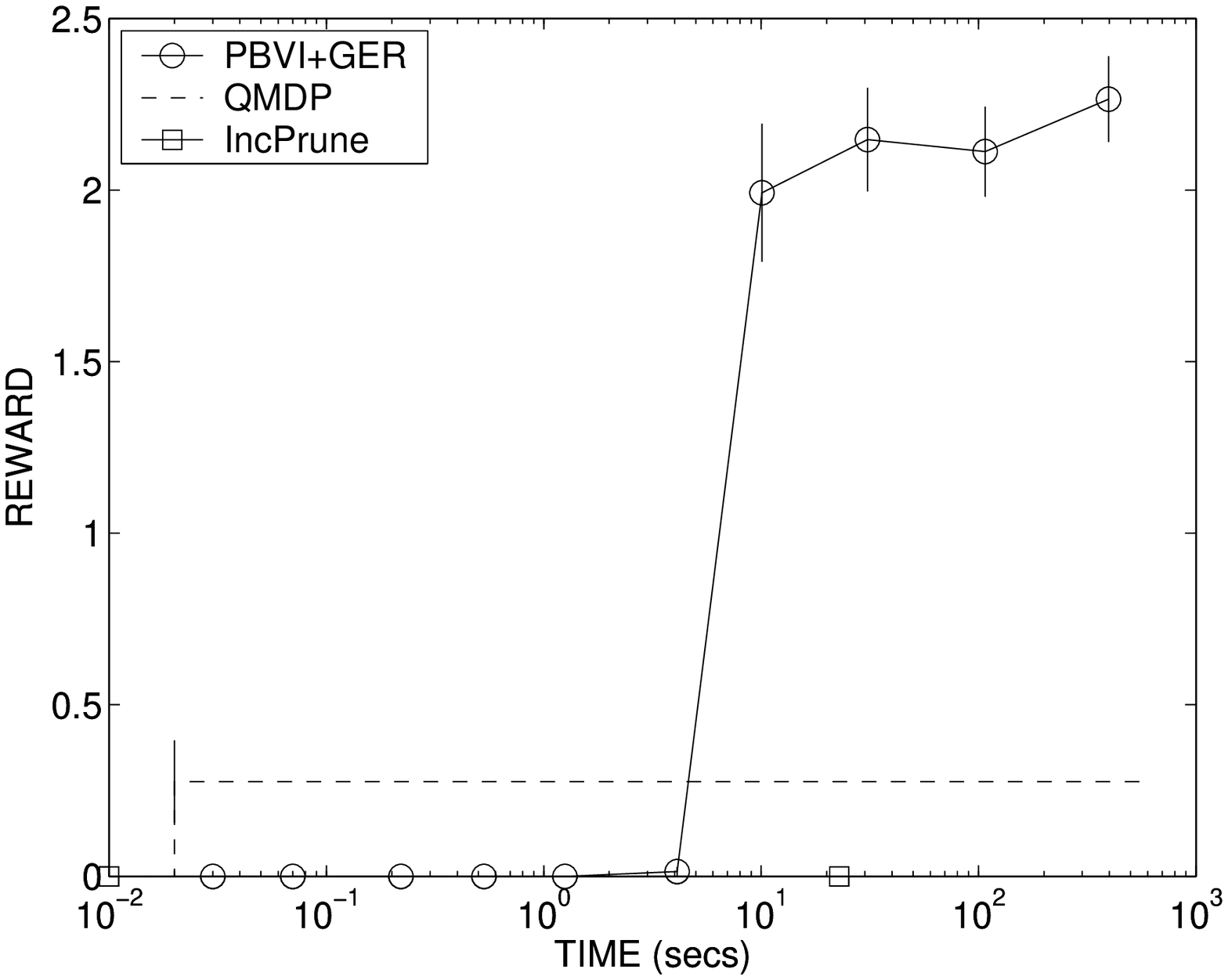}}
\hspace{0cm}
\subfigure[Hallway]{\includegraphics[height=5.5cm]{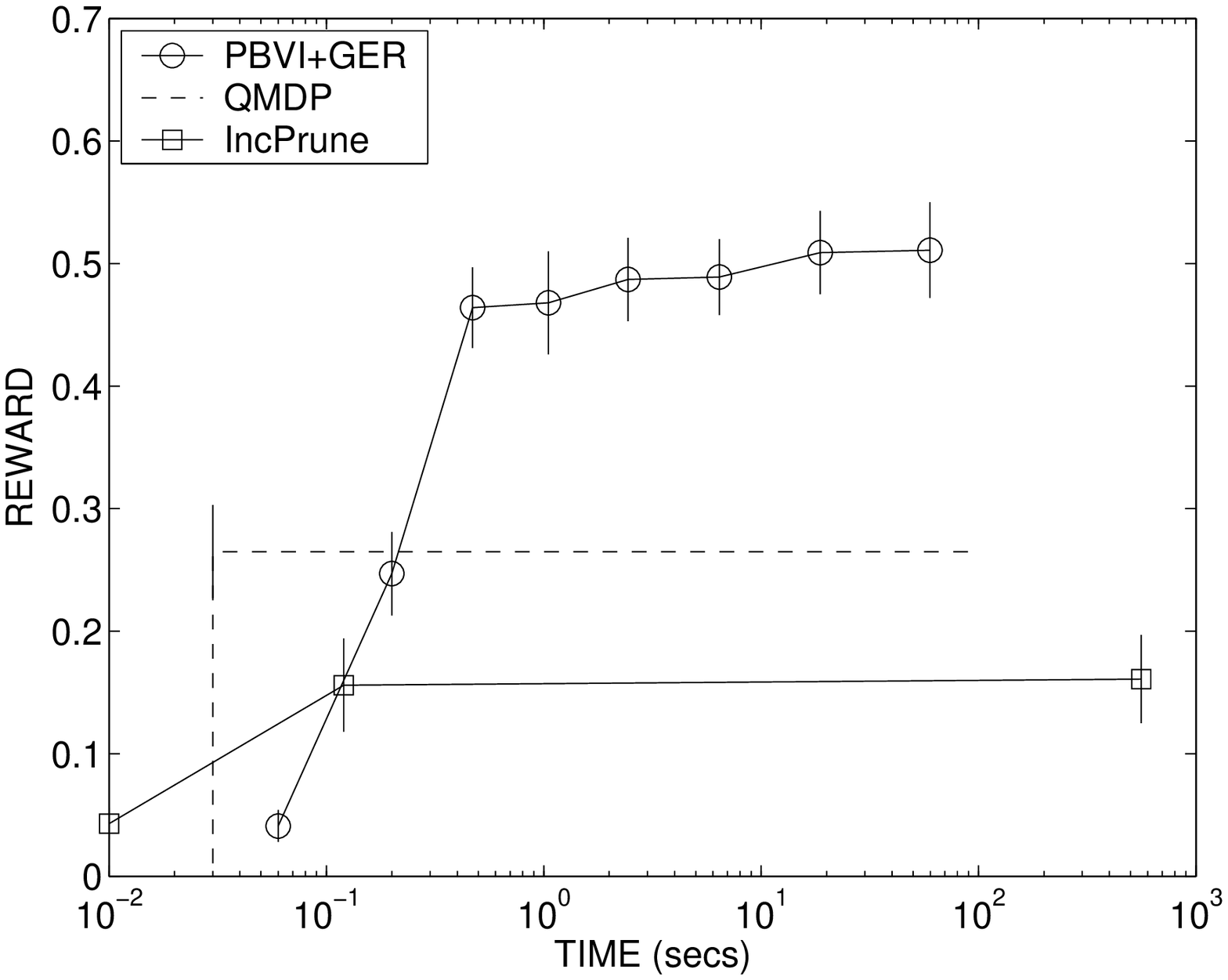}}
\hspace{0cm}
\subfigure[Hallway2]{\includegraphics[height=5.5cm]{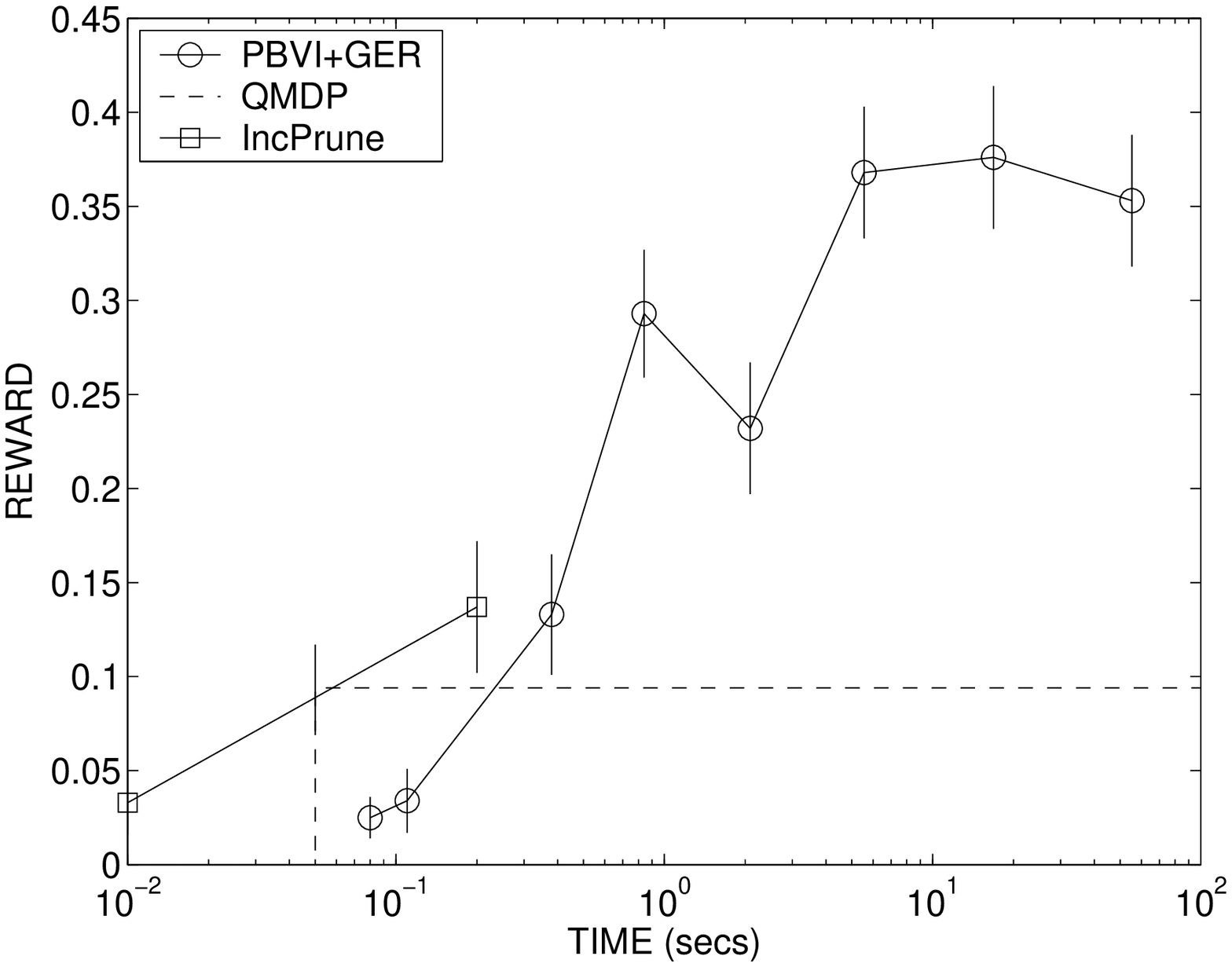}}
\caption{PBVI performance on well-known POMDP problems. Each figure shows the sum of discounted reward as a function of the computation time for a different problem domain.}
\label{fig_results}
\end{figure}

Figure~\ref{fig_results}a presents results for the Tiger-grid domain. Replicating earlier experiments by~\citeA{brafman97}, test runs terminate after 500 steps (there's an automatic reset every time the goal is reached) and results are averaged over 151 runs.

Figures~\ref{fig_results}b and~\ref{fig_results}c present results for the Hallway and Hallway2 domains, respectively. In this case, test runs are terminated when the goal is reached or after 251 steps (whichever occurs first), and the results are averaged over 251 runs. This is consistent with earlier experiments by~\citeA{littman95a}.

All three figures compare the performance of three different algorithms:
\begin{enumerate}
\item PBVI with Greedy Error Reduction (GER) belief point selection (Section~\ref{sec_belief_ger}).
\item QMDP~\cite{littman95a},
\item Incremental Pruning~\cite{cassandra97},
\end{enumerate}

The \textit{QMDP} heuristic~\cite{littman95a} takes into account partial observability at the current step, but assumes full observability on subsequent steps:
\begin{eqnarray}
\pi_{QMDP}(b) & = & \argmax_{a \in A} \sum_{s \in S} b(s)Q_{MDP}(s,a).\label{eqn_piQMDP}
\end{eqnarray}

The resulting policy has some ability to resolve uncertainty, but cannot benefit from long-term information gathering, or compare actions with different information potential. QMDP can be seen as providing a good performance baseline. For the three problems considered, it finds a policy extremely quickly, but the policy is clearly sub-optimal.

At the other end of the spectrum, the Incremental Pruning algorithm~\cite{zhang96,cassandra97} is a direct extension of the enumeration algorithm described above.  The principal insight is that the pruning of dominated $\alpha$-vectors (Eqn~\ref{eqn_prune}) can be interleaved directly with the cross-sum operator (Eqn~\ref{eqn_crosssum}). The resulting value function is the same, but the algorithm is more efficient because it discards unnecessary vectors earlier on. While Incremental Pruning algorithm can theoretically find an optimal policy, for the three problems considered here it would take far too long. In fact, only a few iterations of exact backups were completed in reasonable time. In all three problems, the resulting short-horizon policy was worse than the corresponding PBVI policy.

As shown in Figure~\ref{fig_results}, PBVI+GER provides a much better time/performance trade-off. It finds policies that are better than those obtained with QMDP, and does so in a matter of seconds, thereby demonstrating that it does not suffer from the same paralyzing complexity as Incremental Pruning.

While those who take a closer look at these results may be surprised to see that the performance of PBVI actually \textit{decreases} at some points (e.g., the ``dip'' in Fig.~\ref{fig_results}c), this is not unexpected. It is important to remember that the theoretical properties of PBVI only guarantee a bound on the estimate of the value function, but as shown here, this does not necessarily imply that the policy needs to improve monotonically. Nonetheless, as the value function converges, so will the policy (albeit at a slower rate).

\subsection{Tag Problem}
\label{sec_results_tag}

While the previous section establishes the good performance of PBVI on some well-known simulation problems, these are quite small and do not fully demonstrate the scalability of the algorithm.  To provide a better understanding of PBVI's effectiveness for large problems, this section presents results obtained when applying PBVI to the \textit{Tag} problem, a robot version of the popular game of lasertag.  In this problem, the agent must navigate its environment with the goal of searching for, and tagging, a moving
target~\cite{rosencrantz02}. Real-world versions of this problem can take many forms, and in Section~\ref{sec_robot} we present a similar problem domain where an interactive service robot must find an elderly patient roaming the corridors of a nursing home.

\begin{figure}[!ht]
\centerline{\includegraphics[height=5cm]{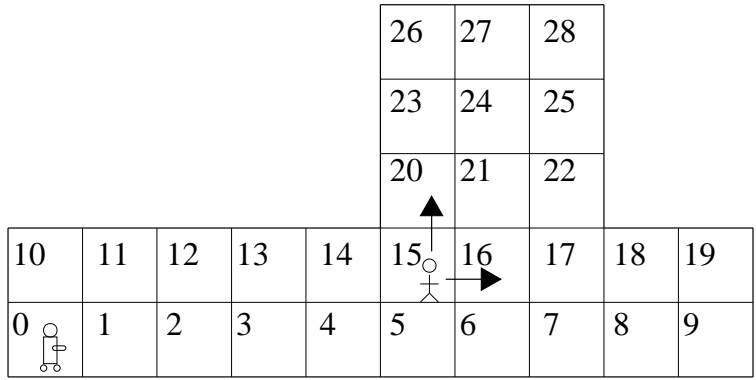}}
\caption{Spatial configuration of the domain}
\label{fig_tag}
\end{figure}

The synthetic scenario considered here is an order of magnitude larger (870 states) than most other POMDP benchmarks in the literature~\cite{cassandra99}.  When formulated as a POMDP problem, the goal is for the robot to optimize a policy allowing it to quickly find the person, assuming that the person moves (stochastically) according to a fixed policy. The spatial configuration of the environment used throughout this experiment is
illustrated in Figure~\ref{fig_tag}.

The state space is described by the cross-product of two position features,
$\mbox{\emph{Robot}}=\{s_0,\ldots,s_{29}\}$ and
$\mbox{\emph{Person}}=\{s_0,\ldots,s_{29},s_{found}\}$. Both 
start in independently-selected random positions, and the scenario
finishes when $\mbox{\emph{Person}}=s_{found}$.  The robot can
select from five actions: \textit{\{North, South, East, West,
Tag\}}. A reward of $-1$ is imposed for each motion action; the
\textit{Tag} action results in a $+10$ reward if the robot and person are in the same cell, or $-10$ otherwise. Throughout the scenario,
the Robot's position is fully observable, and a
\textit{Move} action has the predictable deterministic effect, e.~g.:
\begin{eqnarray}
Pr(Robot=s_{10} \mid Robot=s_0, North)=1,\nonumber
\end{eqnarray}
and so on for each adjacent cell and direction.
The position of the person, on the other hand, is completely unobservable unless both agents are in the same cell. Meanwhile at each step, the person (with omniscient knowledge) moves away from the robot with $Pr=0.8$ and stays in place with $Pr=0.2$, e.~g.:
\begin{eqnarray}
Pr(Person=s_{16} \mid Person=s_{15} \& Robot=s_0)=0.4 \nonumber \\
Pr(Person=s_{20} \mid Person=s_{15} \& Robot=s_0)=0.4 \nonumber \\
Pr(Person=s_{15} \mid Person=s_{15} \& Robot=s_0)=0.2. \nonumber
\end{eqnarray}

\begin{figure}[!ht]
\centerline{\includegraphics[height=6cm]{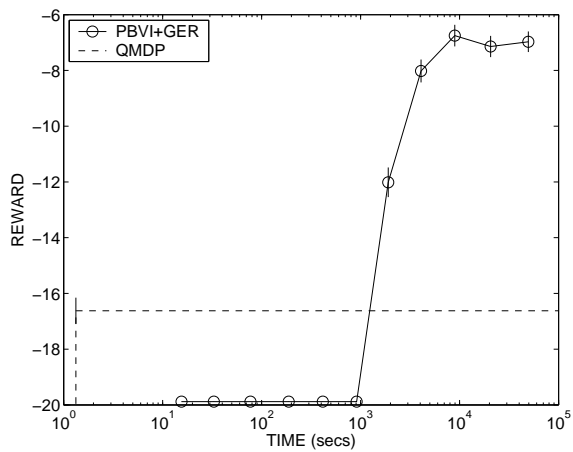}}
\caption{PBVI performance on Tag problem. We show the sum of discounted reward as a function of the computation time.}
\label{fig_resultsTag}
\end{figure}

Figure~\ref{fig_resultsTag} shows the performance of PBVI with Greedy Error Reduction on the Tag domain. Results are averaged over 1000 runs, using different (randomly chosen) start positions for each run. The QMDP approximation is also tested to provide a baseline comparison. The results show a gradual improvement in PBVI's performance as samples are added (each shown data point represents a new expansion of the belief set with value backups). It also confirms that computation time is directly related to the number of belief points.  PBVI requires fewer than 100 belief points to overcome QMDP, and the performance keeps on improving as more points are added. Performance appears to be converging with approximately 250 belief points. These results show that a PBVI-class algorithm can effectively tackle a problem with 870 states.

This problem is far beyond the reach of the Incremental Pruning algorithm. A single iteration of optimal value iteration on a problem of this size could produce over $10^{20}$ $\alpha$-vectors before pruning. Therefore, it was not applied.

This section describes one version of the Tag problem, which was used for simulation purposes in our work and that of others~\cite{braziunas04,poupart03,smith04,vlassis04}. In fact, the problem can be re-formulated in a variety of ways to accommodate different environments, person motion models, and observation models.  Section~\ref{sec_robot} discusses variations on this problem using more realistic robot and person models, and presents results validated onboard an independently developed robot simulator.

\subsection{Empirical Comparison of PBVI-Class Algorithms}
\label{sec_results_heuristics}

Having establish the good performance of PBVI+GER on a number of problems, we now consider empirical results for the different PBVI-class algorithms. This allows us to compare the effects of the various belief expansion heuristics.  We repeat the experiments on the Tiger-grid, Hallway, Hallway2 and Tag domains, as outlined above, but in this case we compare the performance of five different PBVI-class algorithms:
\begin{enumerate}
\item PBVI+RA: PBVI with belief points selected randomly from belief simplex (Section~\ref{sec_belief_ra}).
\item PBVI+SSRA: PBVI with belief points selected using stochastic simulation with random action (Section~\ref{sec_belief_ssra}).
\item PBVI+SSGA: PBVI with belief points selected using stochastic simulation with greedy action (Section~\ref{sec_belief_ssga}).
\item PBVI+SSEA: PBVI with belief points selected using stochastic simulation with exploratory action (Section~\ref{sec_belief_ssea}).
\item PBVI+GER: PBVI with belief points selected using greedy error reduction (Section~\ref{sec_belief_ger}).
\end{enumerate}

All PBVI-class algorithms can converge to the optimal value function given a sufficiently large set of belief points. But the rate at which they converge depends on their ability to generally pick useful points, and leave out the points containing less information.  Since the computation time is directly proportional to the number of belief points, the algorithm with the best performance is generally the one which can find a good solution with the fewest belief points.

\begin{figure}[!ht]
\centering
\subfigure[Tiger-grid]{\includegraphics[height=5.5cm]{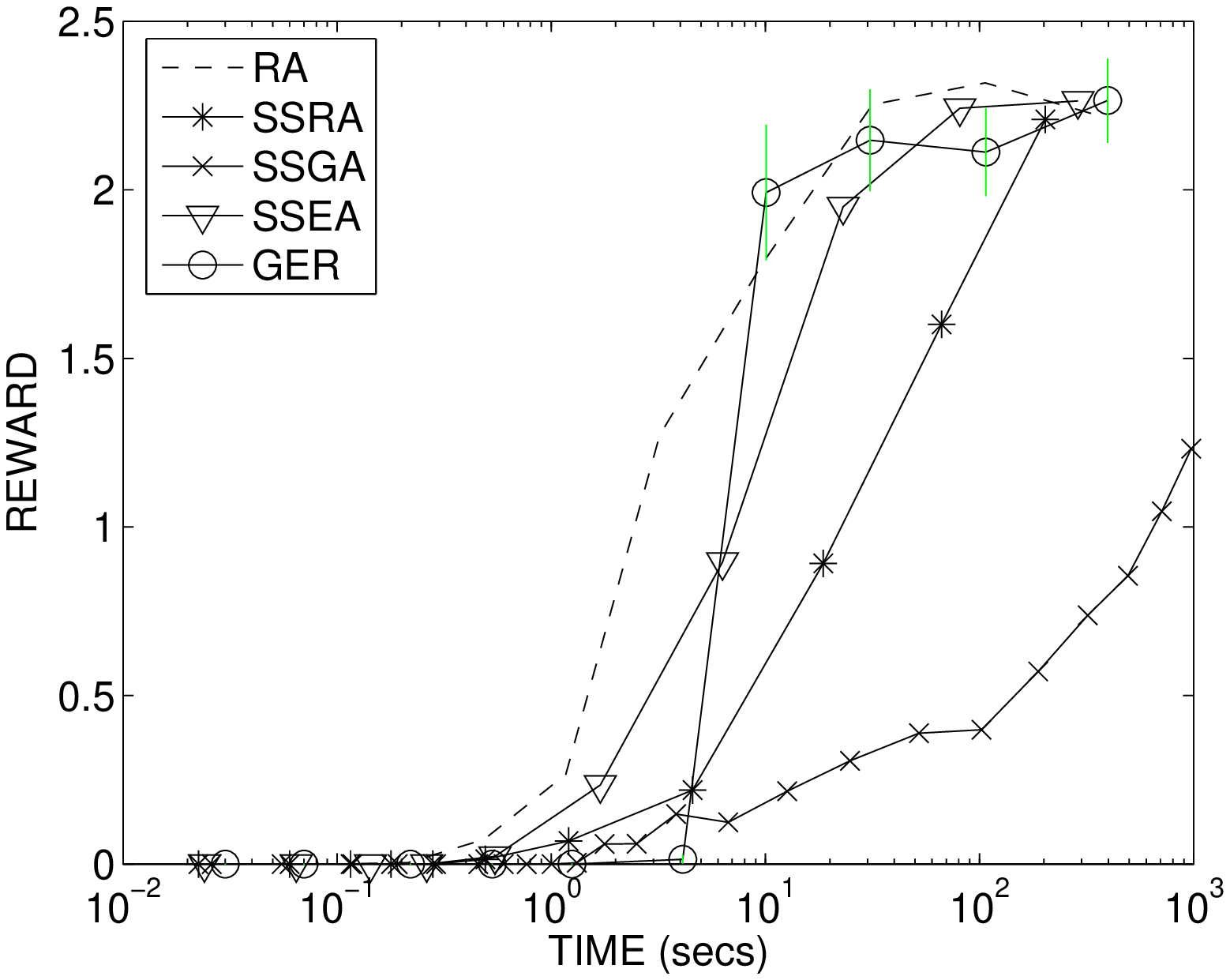}}
\hspace{0cm}
\subfigure[Hallway]{\includegraphics[height=5.5cm]{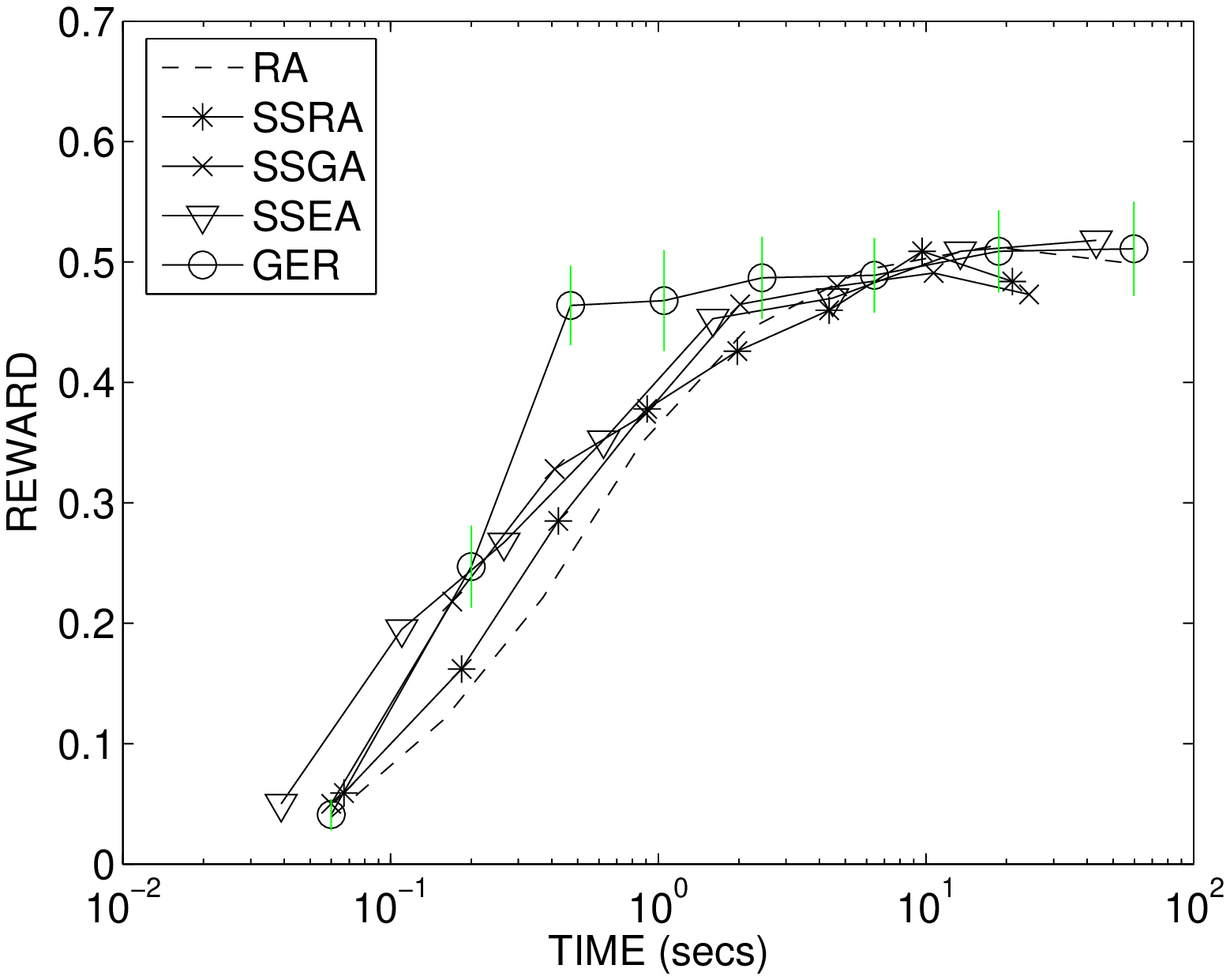}}
\hspace{0cm}
\subfigure[Hallway2]{\includegraphics[height=5.5cm]{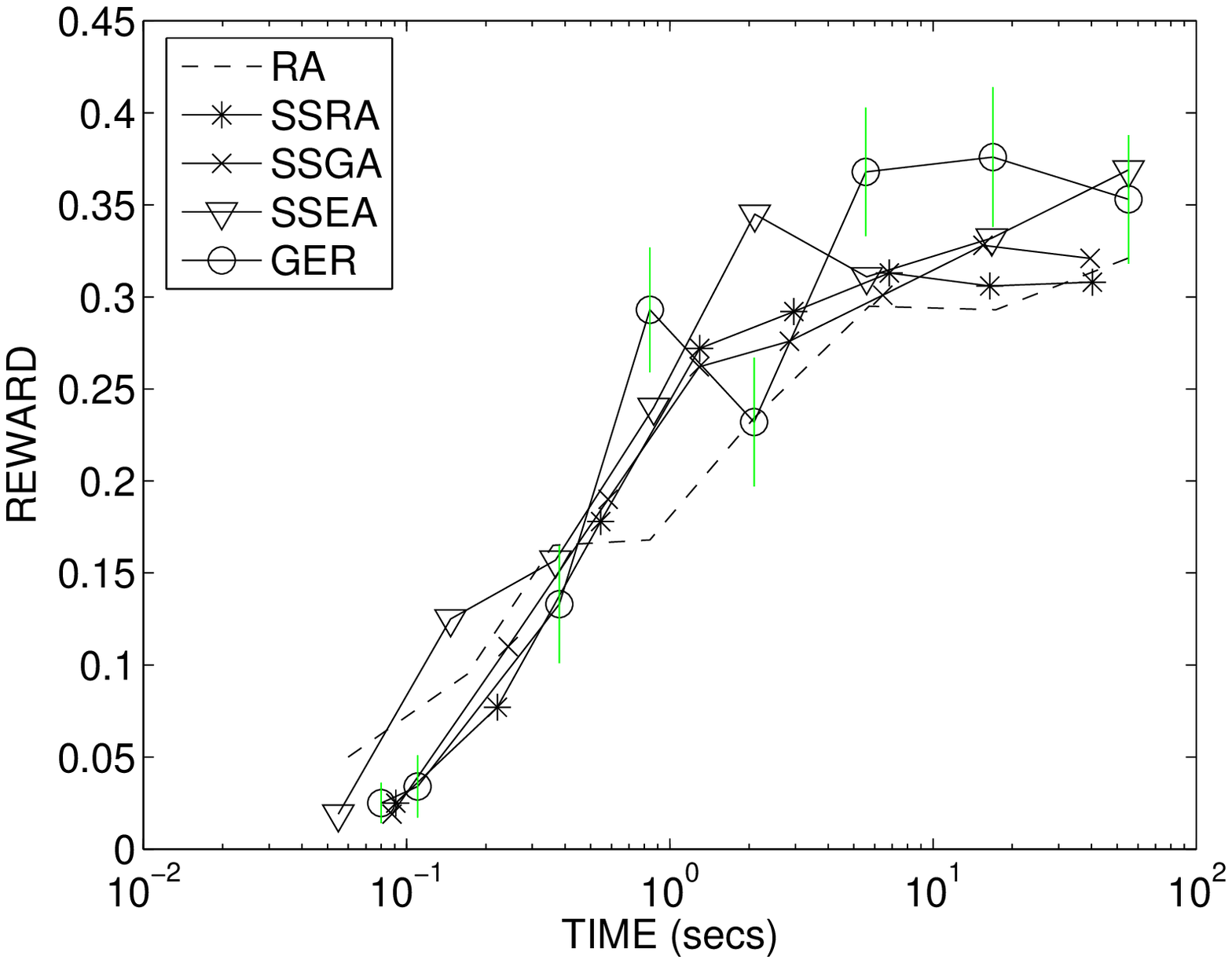}}
\hspace{0cm}
\subfigure[Tag]{\includegraphics[height=5.5cm]{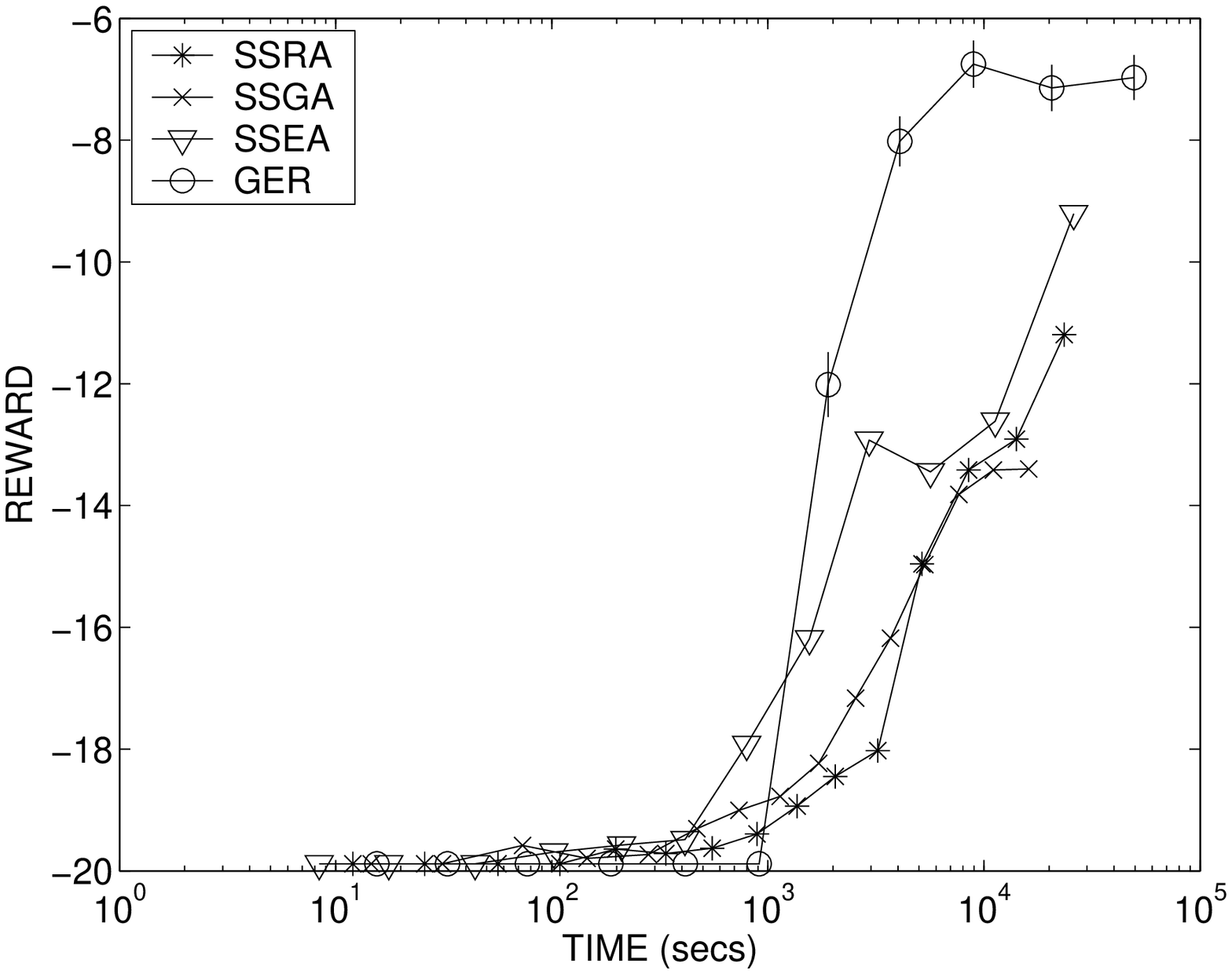}}
\caption{Belief expansion results showing execution performance as a function of the computation time.}
\label{fig_results2}
\end{figure}

Figure~\ref{fig_results2} shows a comparison between the performance of each of the five PBVI-class algorithms enumerated above on each of the four problem domains. In these pictures, we present performance results as a function of computation time.\footnote{Nearly identical graphs can be produced showing performance results as a function of the number of belief points. This confirms complexity analysis showing that the computation time is directly related to the number of belief points.}

As seen from these results, in the smallest domain---Tiger-grid---PBVI+GER is similar in performance to the random approach PBVI+RA\@.  In the Hallway domain, PBVI+GER reaches near-optimal performance earlier than the other algorithms. In Hallway2, it is unclear which of the five algorithms is best, though GER seems to converge earlier.


In the larger Tag domain, the situation is more interesting.  The PBVI+GER combination is clearly superior to the others.  There is reason to believe that PBVI+SSEA could match its performance, but would require on the order of twice as many points to do so. Nonetheless, PBVI+SSEA performs better than either PBVI+SSRA or PBVI+SSGA.  With the random heuristic (PBVI+RA), the reward did not improve regardless of how many belief points were added (4000+), and therefore we do not include it in the results. The results presented in Figure~\ref{fig_results2} suggest that the choice of belief points is crucial when dealing with large problems. In general, we believe that GER (and SSEA to a lesser degree) is superior to the other heuristics for solving domains with large numbers of action/observation pairs, because it has the ability to selectively chooses which branches of the reachability tree to explore.

As a side note, we were surprised by SSGA's poor performance (in comparison with SSRA) on the Tiger-grid and Tag domains.  This could be due to a poorly tuned greedy bias $\epsilon$, which we did not investigate at length. Future investigations using problems with a larger number of actions may shed better light on this issue.

In terms of computational requirement, GER is the most expensive to compute, followed by SSEA. However in all cases, the time to perform the belief expansion step is generally negligible ($<1\%$) compared to the cost of the value update steps. Therefore it seems best to use the more effective (though more expensive) heuristic.

The PBVI framework can accommodate a wide variety of strategies, past what is described in this paper. For example, one could extract belief points directly from sampled experimental traces. This will be the subject of future investigations.

\subsection{Comparative Analysis}
\label{sec_results_compare}

While the results outlined above show that PBVI-type algorithms are able to handle a wide spectrum of large-scale POMDP domains, it is not sufficient to compare the performance of PBVI only to QMDP and Incremental Pruning---the two ends of the spectrum---as done in Section~\ref{sec_results_simulation}. In fact there has been significant activity in recent years in the development of fast approximate POMDP algorithms, and so it is worthwhile to spend some time comparing the PBVI framework to these alternative approaches. This is made easy by the fact that many of these have been validated using the same set of problems as described above.

Table~\ref{table_results1} summarizes the performance of a large number of recent POMDP approximation algorithms, including PBVI, on the four target domains: Tiger-grid, Hallway, Hallway2, and Tag. The algorithms listed were selected based on the availability of comparable published results or available code, or in some cases because the algorithm could be re-implemented easily.

We compare their empirical performance, in terms of execution performance versus planning, on a set of simulation domains.  However as is often the case, these results show that there is not a single algorithm that is best for solving all problems.  We therefore also compile a summary of the attributes and characteristics of each algorithm, in an attempt to tell which algorithm may be best for what types of problems.
Table~\ref{table_results1} includes (whenever possible) the goal completion rates, sum of rewards, policy computation time, number of required belief points, and policy size (number of $\alpha$-vectors, or number of nodes in finite state controllers). The number of belief points and policy size are often identical, however the latter can be smaller if a single $\alpha$-vector is best for multiple belief points.

The results marked [*] were computed by us on a 3GHz Pentium 4; other results were likely computed on different platforms, and therefore time comparisons may be approximate at best. Nonetheless the number of samples and the size of the final policy are both useful indicators of computation time. The results reported for PBVI correspond to the earliest data point from Figures~\ref{fig_results} and~\ref{fig_resultsTag} where PBVI+GER achieves top performance.

Algorithms are listed in order of performance, starting with the algorithm(s) achieving the highest reward. All results assume a standard (not \textit{lookahead}) controller~\cite<see>[for definition]{hauskrecht00}.

Overall, the results indicate that some of the algorithms achieve sub-par performance in terms of expected reward. In the case of QMDP, this is because of fundamental limitations in the algorithm. While Incremental Pruning and the exact value-directed compression can theoretically reach optimal performance, they would require longer computation time to do so.  The grid method (see Tiger-grid results), BPI (see Tiger-grid, Hallway and Tag results) and PBUA (see Tag results) suffer from a similar problem, but offer much more graceful performance degradation. It is worth noting that none of these approaches assumes a known initial belief, so in effect they are solving harder problems. The results for BBSLS are not sufficiently extensive to comment at length, but it appears to be able to find reasonable policies with small controllers (see Tag results).

The remaining algorithms---HSVI, Perseus, and our own PBVI+GER---all offer comparable performance on these relatively large POMDP domains. HSVI seems to offer good control performance on the full range of tasks, but requires bigger controllers, and is therefore probably slower, especially on domains with high stochasticity (e.g., Tiger-grid, Hallway, Hallway2). The trade-offs between Perseus and PBVI+GER are less clear: the planning time, controller size and performance quality are quite comparable, and in fact the two approaches are very similar. Similarities and differences between the two approaches are explored further in Section~\ref{sec_robot}.

\begin{table}[!ht]
\begin{center}
\begin{tabular}{lccccc}
\hline
\hline
Method & Goal\% & Reward $\pm$ Conf.Int. & Time(s) & $|B|$ & $|\pi|$\\
\hline
\multicolumn{1}{l}{\textbf{Tiger-Grid (Maze33)}} \\
HSVI\scriptsize{~\cite{smith04}} & n.a. & 2.35 & 10341 & n.v. & 4860\\
Perseus\scriptsize{~\cite{vlassis04}} & n.a. & 2.34 & 104 & 10000 & 134\\
PBUA\scriptsize{~\cite{poon01}} & n.a. & 2.30 & 12116 & 660 & n.v.\\
PBVI+GER\scriptsize{[*]} & n.a. & 2.27 $\pm$ 0.13 & 397 & 512 & 508\\
BPI\scriptsize{~\cite{poupart03}} & n.a. & 1.81 & 163420 & n.a. & 1500\\
Grid\scriptsize{~\cite{brafman97}} & n.a. & 0.94 & n.v. & 174 & n.a.\\
QMDP\scriptsize{~\cite{littman95a}[*]} & n.a. & 0.276 & 0.02 & n.a. & 5\\
IncPrune\scriptsize{~\cite{cassandra97}[*]} & n.a. & 0.0 & 24hrs+ & n.a. & n.v.\\
Exact VDC\scriptsize{~\cite{poupart02}[*]} & n.a. & 0.0 & 24hrs+ & n.a. & n.v.\\
\hline
\multicolumn{2}{l}{\textbf{Hallway}}\\
PBUA\scriptsize{~\cite{poon01}} & 100 & 0.53 & 450 & 300 & n.v.\\
HSVI\scriptsize{~\cite{smith04}} & 100 & 0.52 & 10836 & n.v. & 1341\\
PBVI+GER\scriptsize{[*]} & 100 & 0.51 $\pm$ 0.03 & 19 & 64 & 64\\
Perseus\scriptsize{~\cite{vlassis04}} & n.v. & 0.51 & 35 & 10000 & 55\\
BPI\scriptsize{~\cite{poupart03}} & n.v. & 0.51 & 249730 & n.a. & 1500\\
QMDP\scriptsize{~\cite{littman95a}[*]} & 51 & 0.265 & 0.03 & n.a. & 5\\
Exact VDC\scriptsize{~\cite{poupart02}[*]} & 39 & 0.161 & 24hrs+ & n.a. & n.v. \\
IncPrune\scriptsize{~\cite{cassandra97}[*]} & 39 & 0.161 & 24hrs+ & n.a. & n.v.\\
\hline
\multicolumn{2}{l}{\textbf{Hallway2}}\\
PBVI+GER\scriptsize{[*]} & 100 & 0.37 $\pm$ 0.04 & 6 & 32 & 31\\
Perseus\scriptsize{~\cite{vlassis04}} & n.v. & 0.35 & 10 & 10000 & 56\\
HSVI\scriptsize{~\cite{smith04}} & 100 & 0.35 & 10010 & n.v. & 1571\\
PBUA\scriptsize{~\cite{poon01}} & 100 & 0.35 & 27898 & 1840 & n.v.\\
BPI\scriptsize{~\cite{poupart03}} & n.v. & 0.28 & 274280 & n.a. & 1500\\
Grid\scriptsize{~\cite{brafman97}} & 98 & n.v. & n.v. & 337 & n.a.\\
QMDP\scriptsize{~\cite{littman95a}[*]} & 22 & 0.109 & 1.44 & n.a. & 5\\
Exact VDC\scriptsize{~\cite{poupart02}[*]} & 48 & 0.137 & 24hrs+ & n.a. & n.v.\\
IncPrune\scriptsize{~\cite{cassandra97}[*]} & 48 & 0.137 & 24hrs+ & n.a. & n.v.\\
\hline
\multicolumn{2}{l}{\textbf{Tag}}\\
HSVI\scriptsize{~\cite{smith04}} & 100 & -6.37 & 10113 & n.v. & 1657\\
PBVI+GER\scriptsize{[*]} & 100 & -6.75 $\pm$ 0.39 & 8946 & 256 & 203\\
Perseus\scriptsize{~\cite{vlassis04}} & n.v. & -6.85 & 3076 & 10000 & 205\\
BBSLS\scriptsize{~\cite{braziunas04}} & n.v. & -8.31 & 100054 & n.a. & 30\\
BPI\scriptsize{~\cite{poupart03}} & n.v. & -9.18 & 59772 & n.a. & 940\\
QMDP\scriptsize{~\cite{littman95a}[*]} & 19 & -16.62 & 1.33 & n.a. & 5\\
PBUA\scriptsize{~\cite{poon01}[*]} & 0 & -19.9 & 24hrs+ & 4096 & n.v.\\
IncPrune\scriptsize{~\cite{cassandra97}[*]} & 0 & -19.9 & 24hrs+ & n.a. & n.v.\\
\hline
\hline
\scriptsize{n.a.=not applicable} & \multicolumn{2}{l}{\scriptsize{n.v.=not available}} & \multicolumn{3}{l}{\scriptsize{[*]=results computed by us}}\\
\\
\end{tabular}
\caption{Results of PBVI for standard POMDP domains}
\label{table_results1}
\end{center}
\end{table}
\clearpage

\subsection{Error Estimates}
\label{sec_results_error}

The results presented thus far suggest that the PBVI framework performs best when using the Greedy Error Reduction (GER) technique for selecting belief points. Under this scheme, to decide which belief points will be included, we estimate an error bound at a set of candidate points and then pick the one with the largest error estimate.  The error bound is estimated as described in Equation~\ref{eqn_error}. We now consider the question of how this estimate evolves as more and more points are added. The natural intuition is that with the first few points, error estimates will be very large, but as the density of the belief set increases, the error estimates will become much smaller.

\begin{figure}[!htb]
\centering
\subfigure{\includegraphics[height=6.6cm]{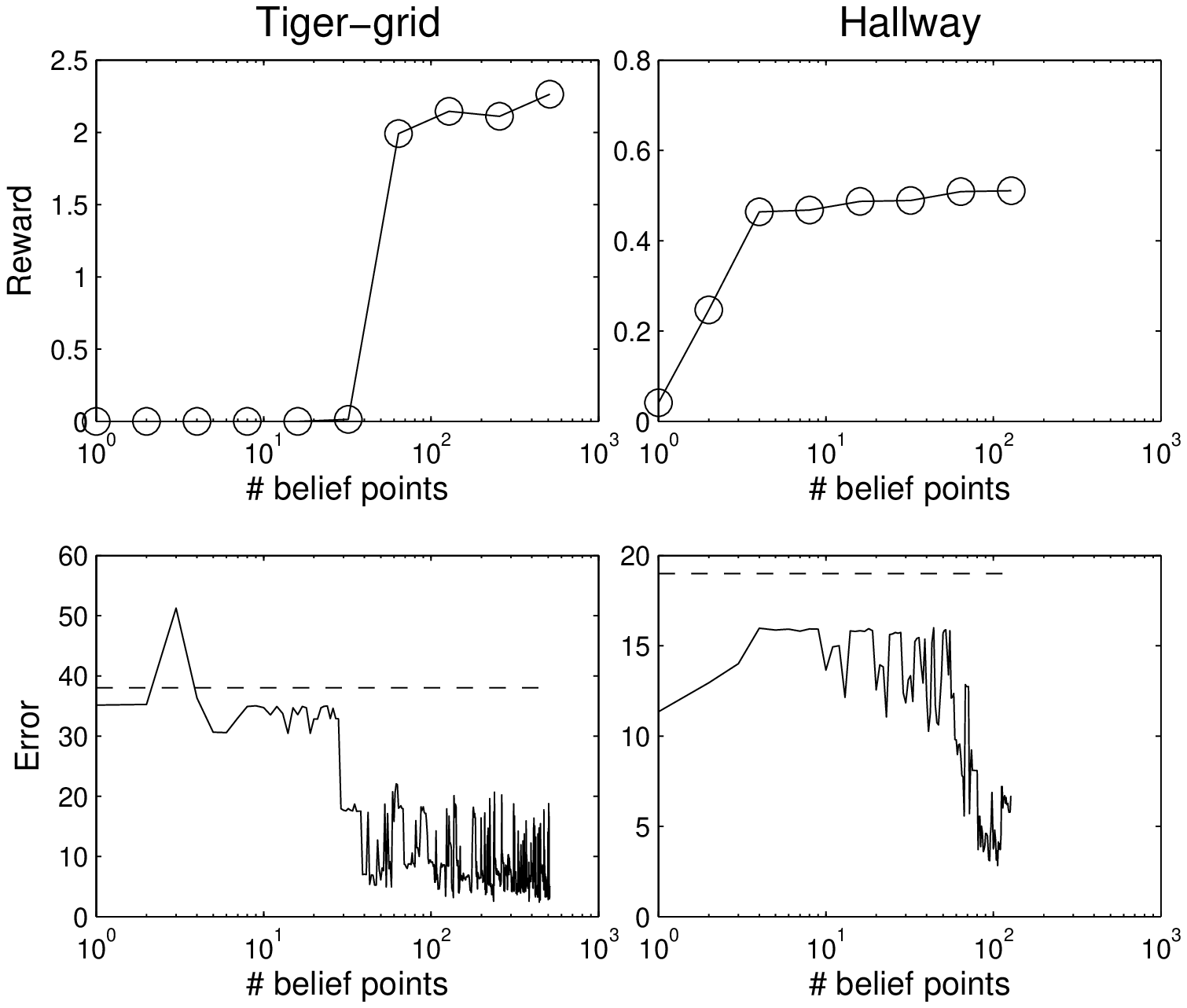}}
\hspace{-0.4cm}
\subfigure{\includegraphics[height=6.6cm]{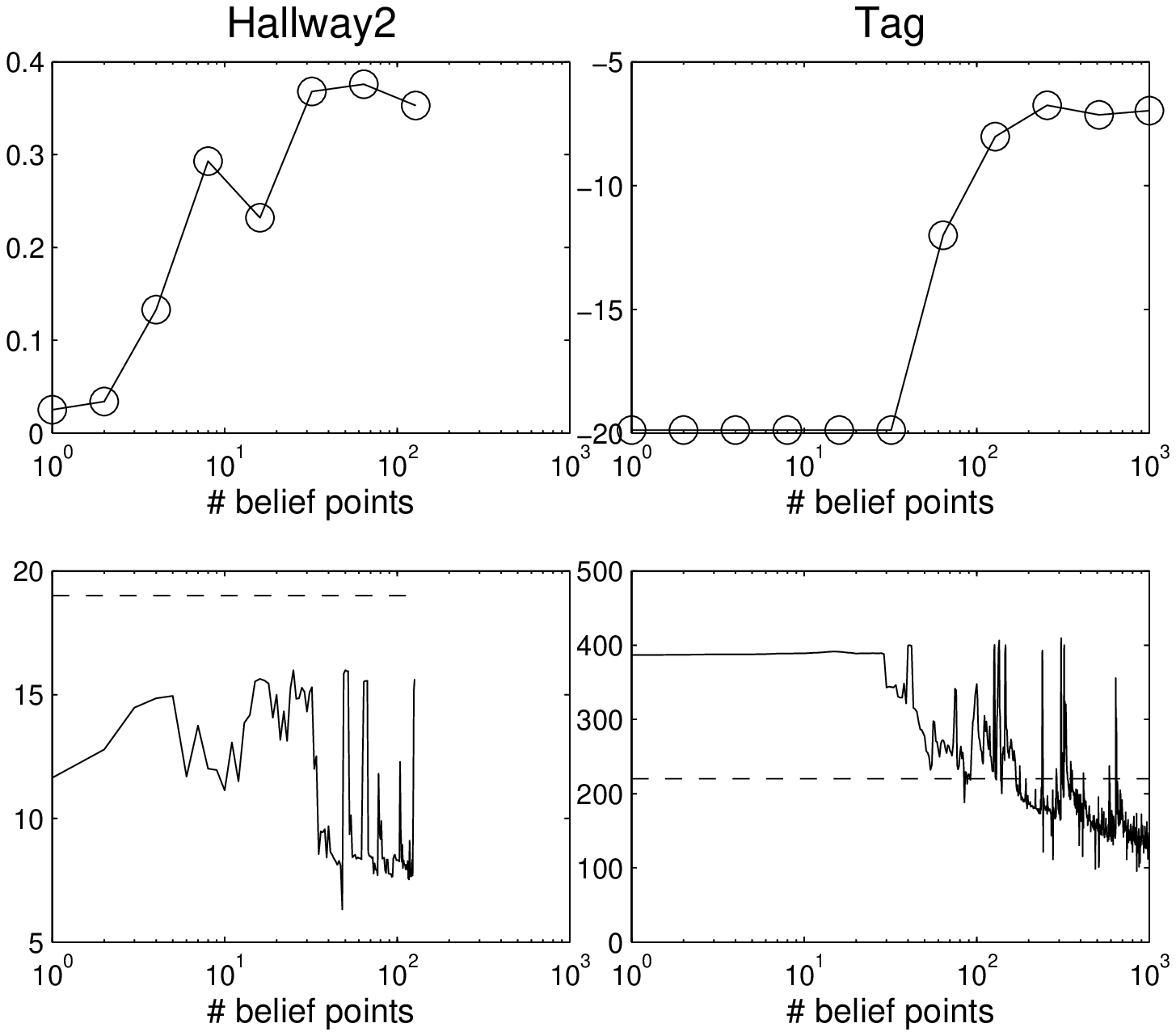}}
\hspace{0cm}
\caption{Sum of discounted reward (top graphs) and estimate of the bound on the error (bottom graphs) as a function of the number of selected belief points.}
\label{fig_error}
\end{figure}

Figure~\ref{fig_error} reconsiders the four target domains: Tiger-grid, Hallway, Hallway2 and Tag. In each case, we present both the reward performance as a function of the number of belief points (top row graphs), and the error estimate of each point selected according the order in which points were picked (bottom row graphs).  In addition, the bottom graphs also show (in dashed line) a trivial bound on the error $||V_t - V^*_t|| \leq \frac{R_{max}-R_{min}}{1-\gamma}$, valid for any t-step value function of an arbitrary policy. As expected, our bound is typically tighter than the trivial bound. In Tag, this only occurs once the number of belief points exceeds the number of states, which is not surprising, given that our bound depends on distance between reachable beliefs, and that all states are reachable beliefs in this domain. Overall, it seems that there is reasonably good correspondence between an improvement in performance and a decrease in our error estimates. We can conclude from this figure that even though the PBVI error is quite loose, it can in fact be informative in guiding exploration of the belief simplex.

We note that there is significant variance in our error estimates from one belief point to the next, as illustrated by the non-monotonic behavior of the curves in the bottom graphs of Figure~\ref{fig_error}. This behavior can be attributed to a few possibilities. First, there is the fact that the error estimate at a given belief is only approximate. And the value function used to calculate the error estimate is itself approximate. In addition, there is the fact that new belief points are always selected from the envelope of reachable beliefs, not from the set of all reachable beliefs. This suggests that GER could be improved by maintaining a deeper envelope of candidate belief points. Currently the envelope contains those points that are 1-step forward simulations from the points already selected. It may be useful to consider points 2--3 steps ahead. We predict this would reduce the jaggedness seen in Figure~\ref{fig_error}, and more importantly, also reduce the number of points necessary for good performance.  Of course, the tradeoff between the time spent selecting points and the time spent planning would have to be re-evaluated under this light.

\section{Robotic Applications}
\label{sec_robot}

\begin{figure}[!ht]
\centering
\subfigure{\includegraphics[height=6cm]{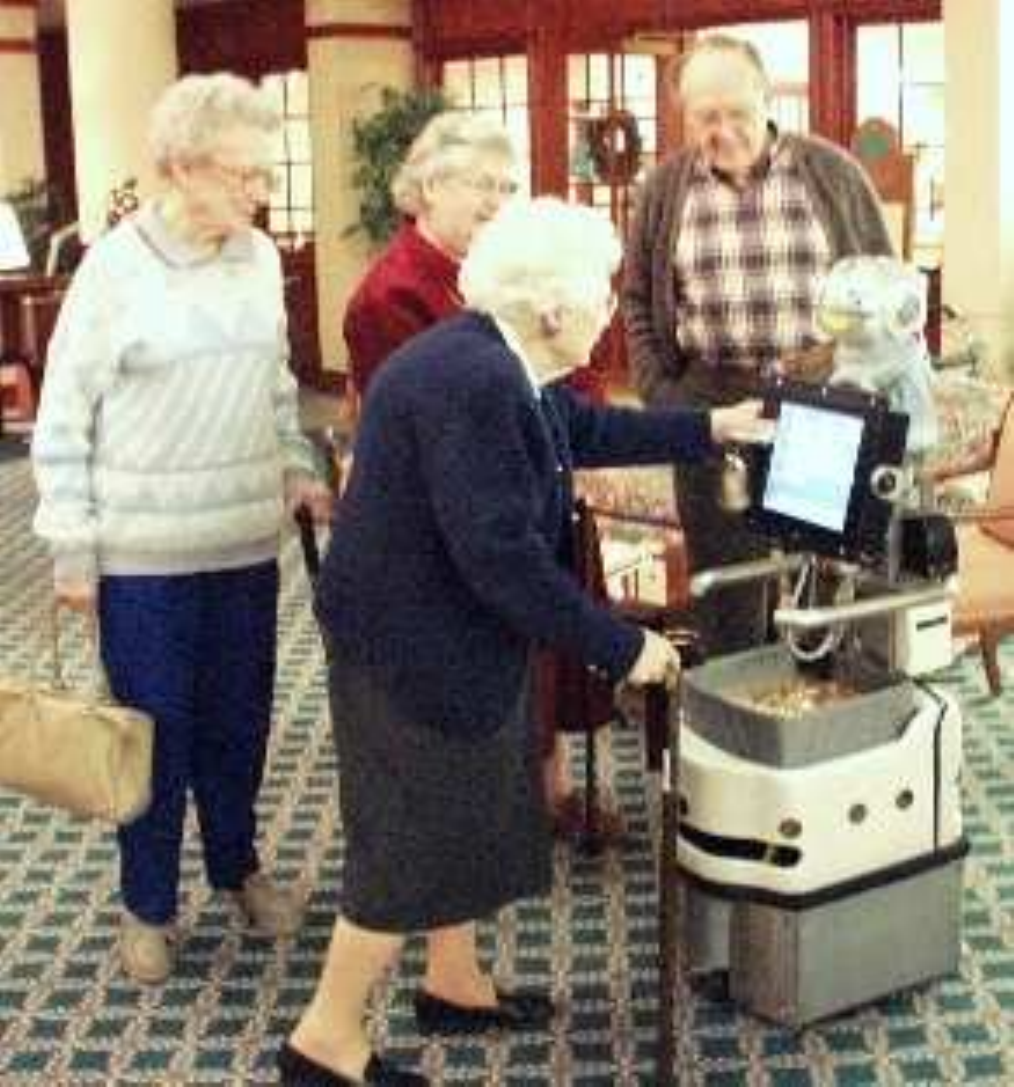}}
\hspace{1cm}
\caption{Pearl the Nursebot, interacting with elderly people at a nursing facility}
\label{fig_robots}
\end{figure}

The overall motivation behind the work described in this paper is the desire to provide high-quality robust planning for real-world autonomous systems, and in particular for robots. On the practical side, our search for a robust robot controller has been in large part guided by the Nursebot project~\cite{pineau03}.    The overall goal of the project is to develop personalized robotic technology that can play an active role in providing improved care and services to non-institutionalized elderly people. Pearl, shown in Figure~\ref{fig_robots}, is the main robotic platform used for this project.

\begin{figure}[!ht]
\centerline{\includegraphics[height=4cm]{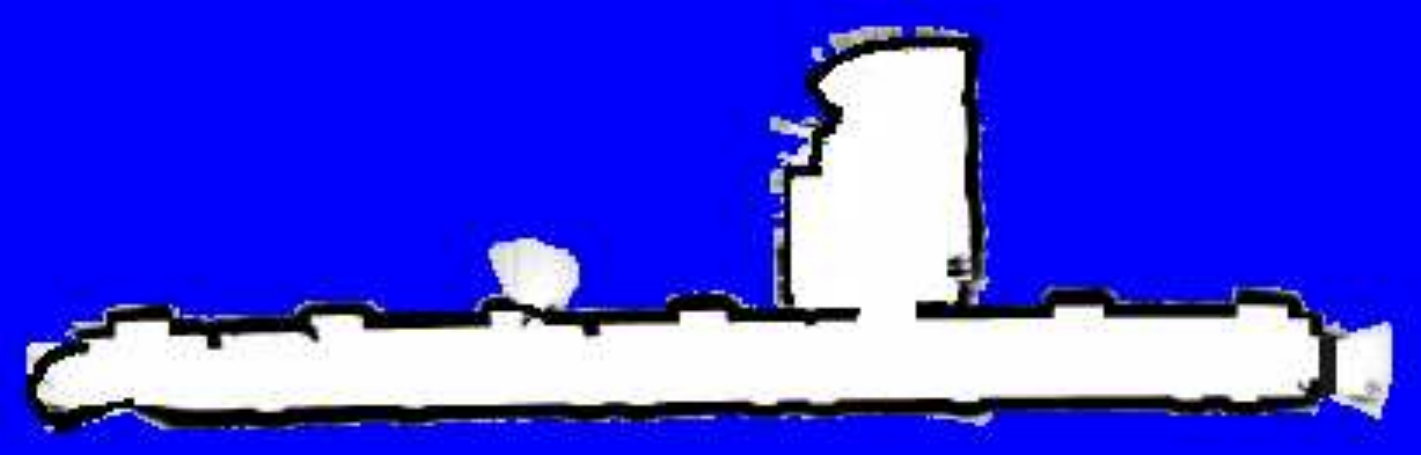}}
\caption{Map of the environment}
\label{fig_wean}
\end{figure}

From the many services a nursing-assistant robot could provide
~\cite{engelberger99,lacey98}, much of the work to date has focused on providing timely cognitive reminders (e.g., medications to take, appointments to attend, etc.) to elderly subjects~\cite{pollack02}. An important component of this task is finding the patient whenever it is time to issue a reminder. This task shares many similarities with the Tag problem presented in Section~\ref{sec_results_tag}. In this case, however, a robot-generated map of a real physical environment is used as the basis for the spatial configuration of the domain. This map is shown in Figure~\ref{fig_wean}. The white areas correspond to free space, the black lines indicate walls (or other obstacles) and the dark gray areas are not visible or accessible to the robot. One can easily imagine the patient's room and physiotherapy unit lying at either end of the corridor, with a common area shown in the upper-middle section.

The overall goal is for the robot to traverse the domain in order to find the missing patient and then deliver a message.  The robot must systematically explore the environment, reasoning about both spatial coverage and human motion patterns, in order to find the person.

\subsection{POMDP Modeling}
\label{sec_robot_model}

The problem domain is represented jointly by two state features: \textit{RobotPosition, PersonPosition}. Each feature is expressed through a discretization of the environment. Most of the experiments below assume a discretization of 2 meters, which means 26 discrete cells for each feature, for a total of 676 states.

It is assumed that the person and robot can move freely throughout this space.  The robot's motion is deterministically controlled by the choice of action (\textit{North, South, East, West}). The robot has a fifth action (\textit{DeliverMessage}), which concludes the scenario when used appropriately (i.e., when the robot and person are in the same location).

The person's motion is stochastic and falls in one of two modes. Part of the time, the person moves according to Brownian motion (e.g., moves in each cardinal direction with $Pr=0.1$, otherwise stays put). At other times, the person moves directly away from the robot. The Tag domain of Section~\ref{sec_results_tag} assumes that the person always moves always moves away the robot. This is not realistic when the person cannot see the robot.  The current experiment instead assumes that the person moves according to Brownian motion when the robot is far away, and moves away from the robot when it is closer (e.g., $<$ 4m). The person policy was designed this way to encourage the robot to find a robust policy.

In terms of state observability, there are two components: what the robot can sense about its own position, and what it can sense about the person's position.  In the first case, the assumption is that the robot knows its own position at all times. While this may seem like a generous (or optimistic) assumption, substantial experience with domains of this size and maps of this quality have demonstrated robust localization abilities~\cite{thrun00a}. This is especially true when planning operates at relatively coarse resolution (2 meters) compared to the localization precision (10 cm).  While exact position information is assumed for \textit{planning} in this domain, the \textit{execution} phase (during which we actually measure performance) does update the belief using full localization information, which includes positional uncertainty whenever appropriate.

Regarding the detection of the person, the assumption is that the robot has no knowledge of the person's position unless s/he is within a range of 2 meters. This is plausible given the robot's sensors. However, even in short-range, there is a small probability ($Pr=0.01$) that the robot will miss the person and therefore return a false negative.

In general, one could make sensible assumptions about the person's likely position (e.g., based on a knowledge of their daily activities), however we currently have no such information and therefore assume a uniform distribution over all initial positions. The person's subsequent movements are expressed through the motion model described above (i.e., a mix of Brownian motion and purposeful avoidance).

The reward function is straightforward: $R=-1$ for any motion action, $R=10$ when the robot decides to \textit{DeliverMessage} and it is in the same cell as the person, and $R=-100$ when the robot decides to \textit{DeliverMessage} in the person's absence.  The task terminates when the robot successfully delivers the message (i.e., $a=DeliverMessage$ and $s_{robot}=s_{person}$). We assume a discount factor of $0.95$.

We assume a known initial belief, $b_0$, consisting of a uniform distribution over all states. This is used both for selecting belief points during planning, and subsequently for executing and testing the final policy.

The initial map (Fig.~\ref{fig_wean}) of the domain was collected by a mobile robot, and slightly cleaned up by hand to remove artifacts (e.g., people walking by). We then assumed the model parameters described here, and applied PBVI planning to the problem as such. Value updates and belief point expansions were applied in alternation until (in simulation) the policy was able to find the person on $99\%$ of trials (trials were terminated when the person is found or after 100 execution steps). The final policy was implemented and tested onboard the publicly available CARMEN robot simulator~\cite{montemerlo03}.

\subsection{Comparative Evaluation of PBVI and Perseus}
\label{sec_robot_compare}

The subtask described here, with its 626 states, is beyond the capabilities of exact POMDP solvers.  Furthermore, as will be demonstrated below, MDP-type approximations are not equipped to handle uncertainty of the type exhibited in this task. The main purpose of our analysis is to evaluate the effectiveness of the point-based approach described in this paper to address this problem. While the results on the Tag domain (Section~\ref{sec_results_tag}) hint at the fact that PBVI and other algorithms may be able to handle this task, the more realistic map and modified motion model provide new challenges.

We begin our investigation by directly comparing the performance of PBVI (with GER belief points selection) with that of the Perseus algorithm on this complex robot domain. Perseus was described in Section~\ref{sec_related}; results presented below were produced using code provided by its authors~\cite{perseus05}. Results for both algorithms assume that a fixed POMDP model was generated by the robot simulator. This model is then stored and solved offline by each algorithm.

PBVI and Perseus each have a few parameters to set. PBVI requires: number of new belief points to add at each expansion ($B_{add}$) and planning horizon for each expansion ($H$). Perseus requires: number of belief points to generate during random walk ($B$) and the maximum planning time ($T$). Results presented below assume the following parameter settings: $B_{add}=30$, $H=25$, $B$=10,000, $T$=1500. Both algorithms were fairly robust to changes in these parameters.\footnote{A $\pm 25\%$ change in parameter value yielded sensibly similar results in terms of reward and number of $\alpha$ vectors, though of course the time, memory, and number of beliefs varied.}

\begin{figure}[!hbt]
\centering
\subfigure{\includegraphics[height=5cm]{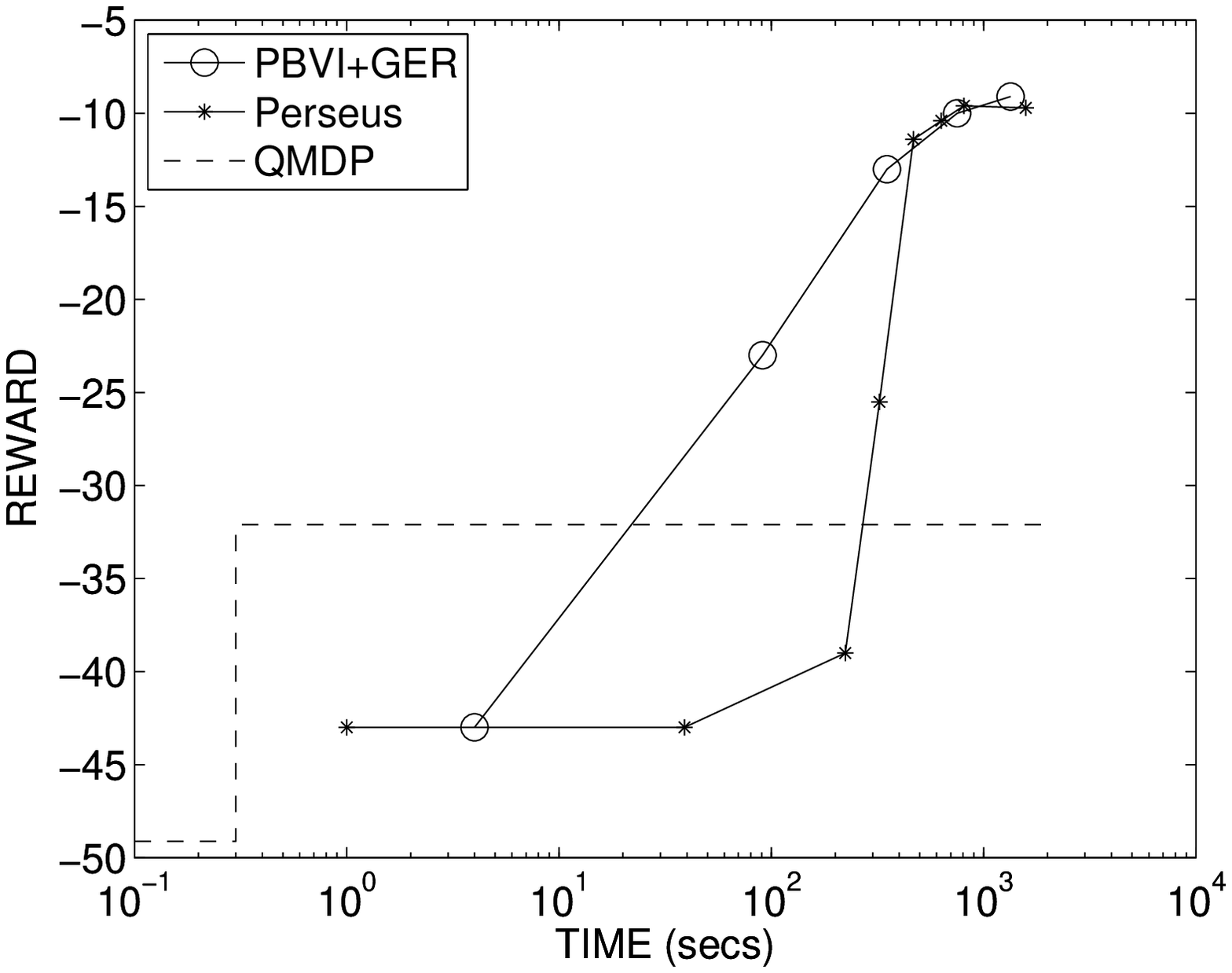}}
\subfigure{\includegraphics[height=5cm]{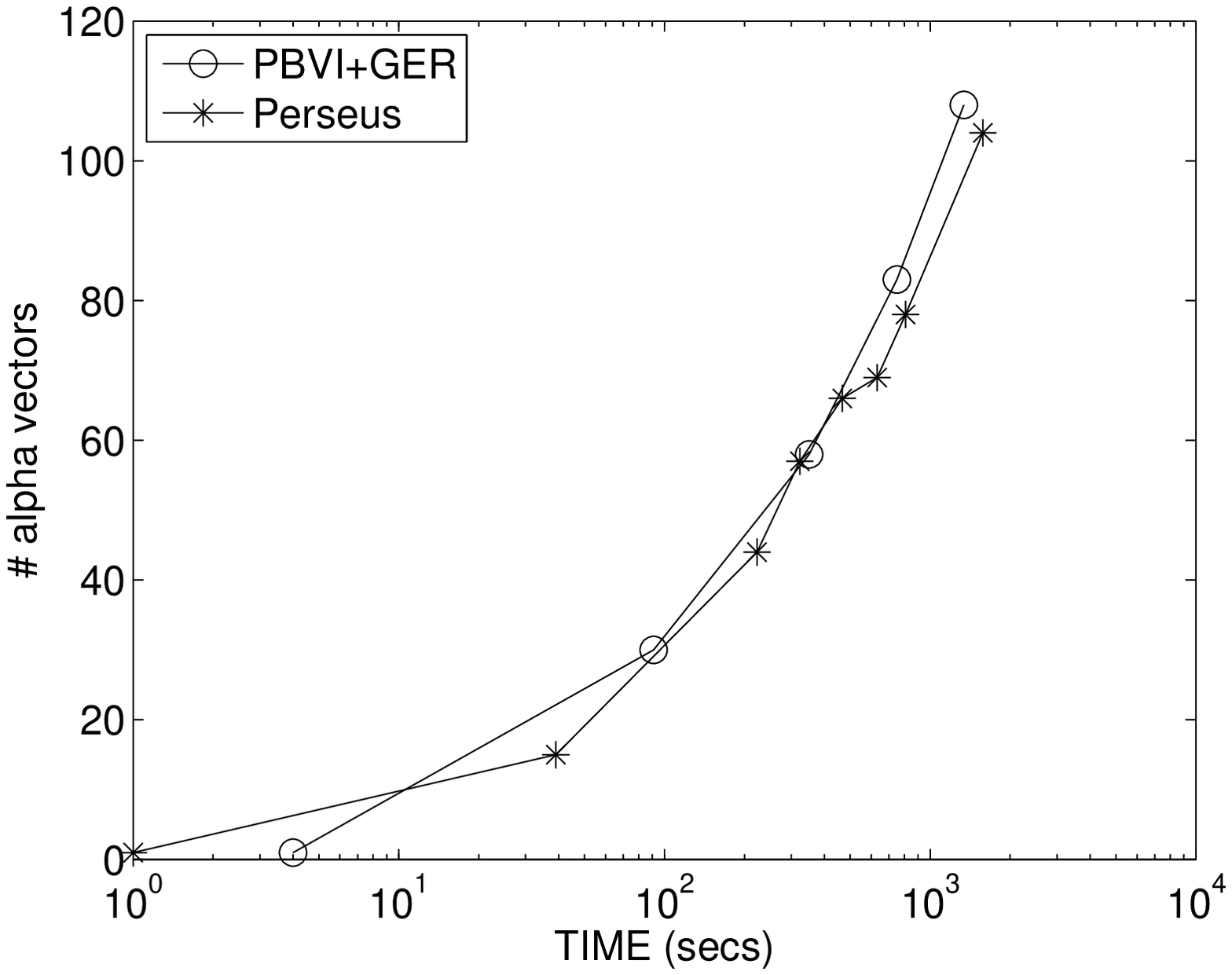}}
\hspace{2cm}
\subfigure{(a)}
\hspace{6cm}
\subfigure{(b)}
\hspace{5cm}
\subfigure{\includegraphics[height=5cm]{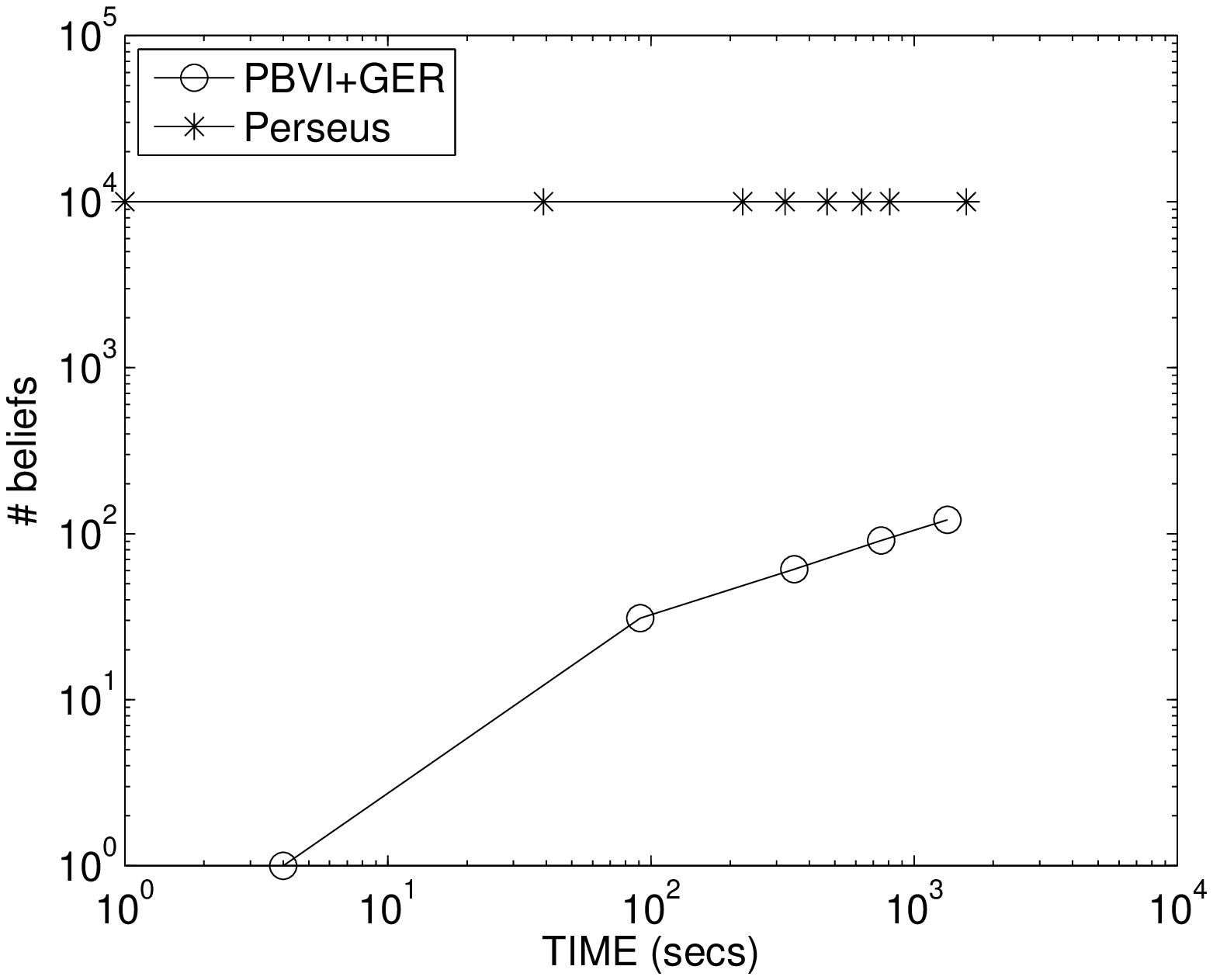}}
\subfigure{\includegraphics[height=5cm]{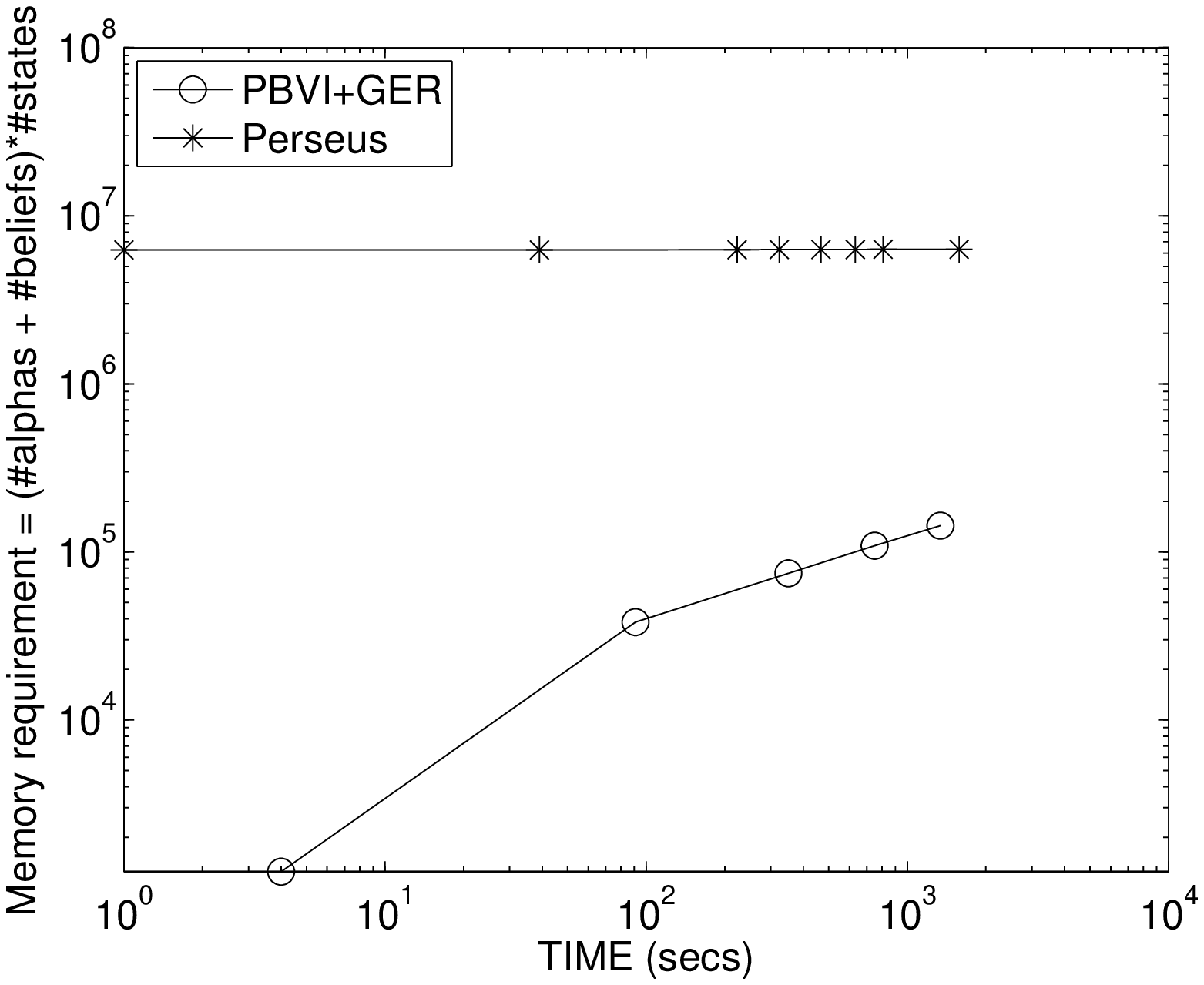}}
\hspace{8cm}
\subfigure{(c)}
\hspace{6cm}
\subfigure{(d)}
\caption{Comparison of PBVI and Perseus on robot simulation domain}
\label{fig_pbvi-perseus}
\end{figure}

Figure~\ref{fig_pbvi-perseus} summarizes the results of this experiment. These suggest a number of observations.
\begin{itemize} 
\item As shown in Figure~\ref{fig_pbvi-perseus}(a), both algorithms find the best solution in a similar time, but PBVI+GER has better anytime performance than Perseus (e.g., a much better policy is found given only 100 sec).
\item As shown in Figure~\ref{fig_pbvi-perseus}(b), both algorithms require a similar number of $\alpha$-vectors.
\item As shown in Figure~\ref{fig_pbvi-perseus}(c), PBVI+GER requires many fewer beliefs.
\item Because it requires fewer beliefs, PBVI+GER has much lower memory requirements; this is quantified in Figure~\ref{fig_pbvi-perseus}(d).
\end{itemize}

These new results suggest that PBVI and Perseus have similar performance if the objective is to find a near-optimal solution, and time and memory are not constrained.  In cases where one is willing to trade off accuracy for time, then PBVI may provide superior anytime performance. And in cases where memory is limited, PBVI's conservative approach with respect to belief point selection is advantageous.  Both these properties suggest that PBVI may scale better to very large domains.

\subsection{Experimental Results with Robot Simulator}
\label{sec_robot_results}

The results presented in above assume that the same POMDP model is used for planning and testing (i.e., to compute the reward in Figure~\ref{fig_pbvi-perseus}(a)).  This is useful to carry out a large number of experiments. The model however cannot entirely capture the dynamics of a realistic robot system, therefore there is some concern that the policy learned by point-based methods will not perform as well on a realistic robot. To verify the robustness of our approach, the final PBVI control policy was implemented and tested onboard the publicly available CARMEN robot simulator~\cite{montemerlo03}.

\begin{figure}[!ht]
\centering
\subfigure{(a) t=1}
\hspace{1cm}
\subfigure{\includegraphics[trim=0 40 25 5,clip=true,height=3.1cm]{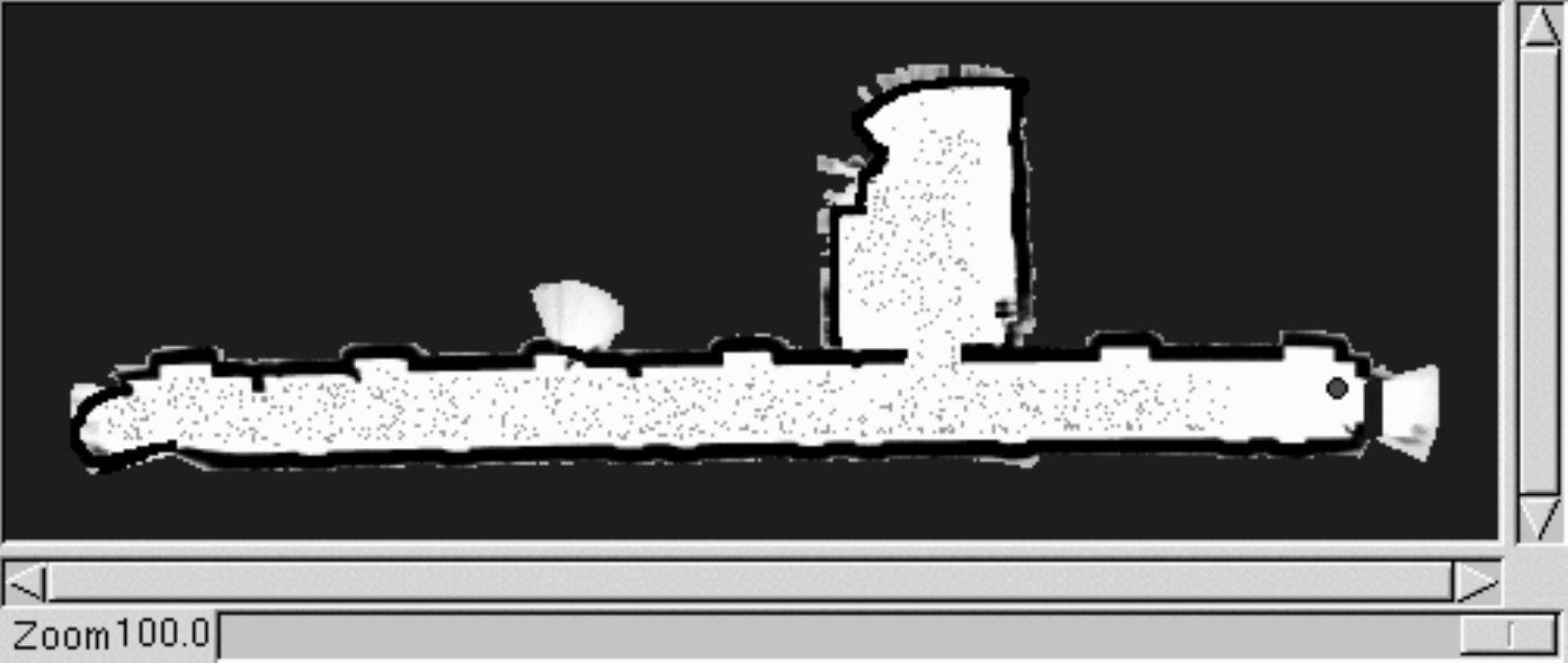}}
\hspace{4cm}
\subfigure{(b) t=7}
\hspace{1cm}
\subfigure{\includegraphics[trim=0 40 25 5,clip=true,height=3.1cm]{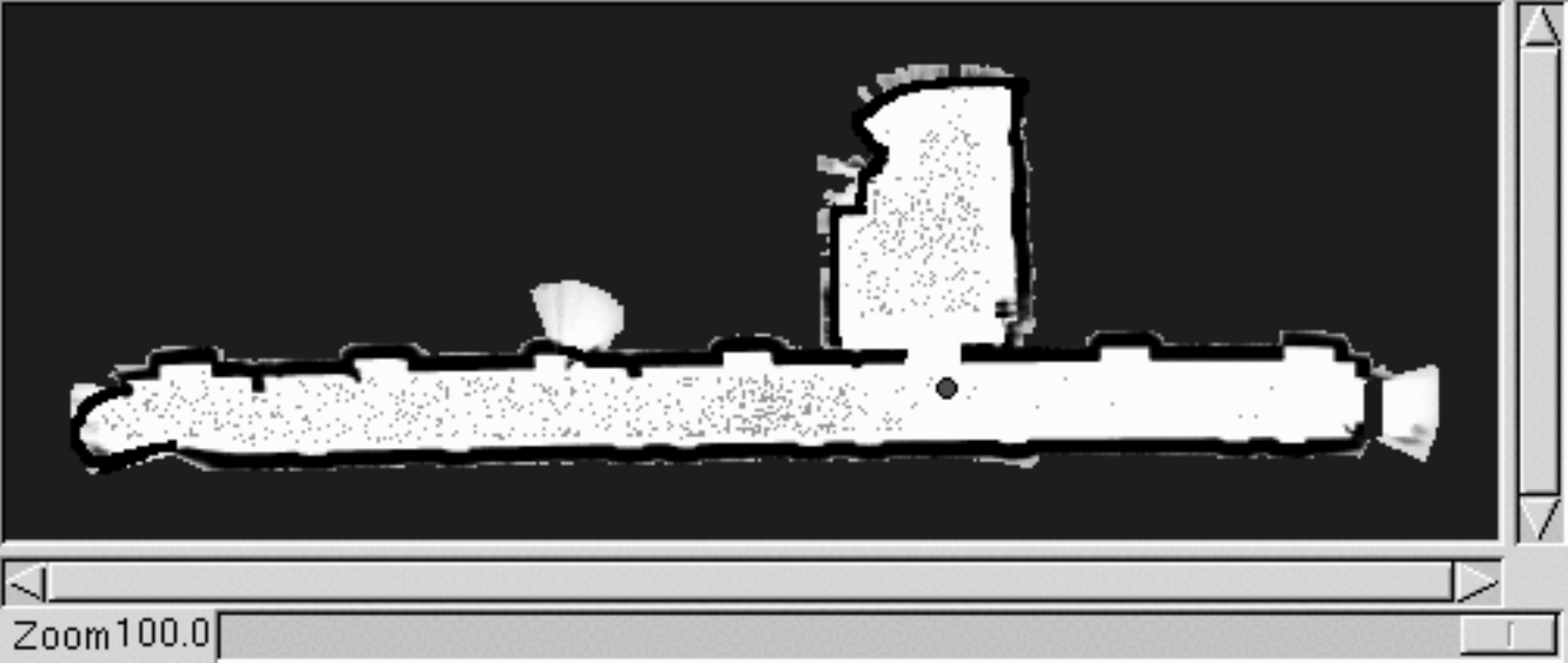}}
\hspace{4cm}
\subfigure{(c) t=12}
\hspace{1cm}
\subfigure{\includegraphics[trim=0 40 25 5,clip=true,height=3.1cm]{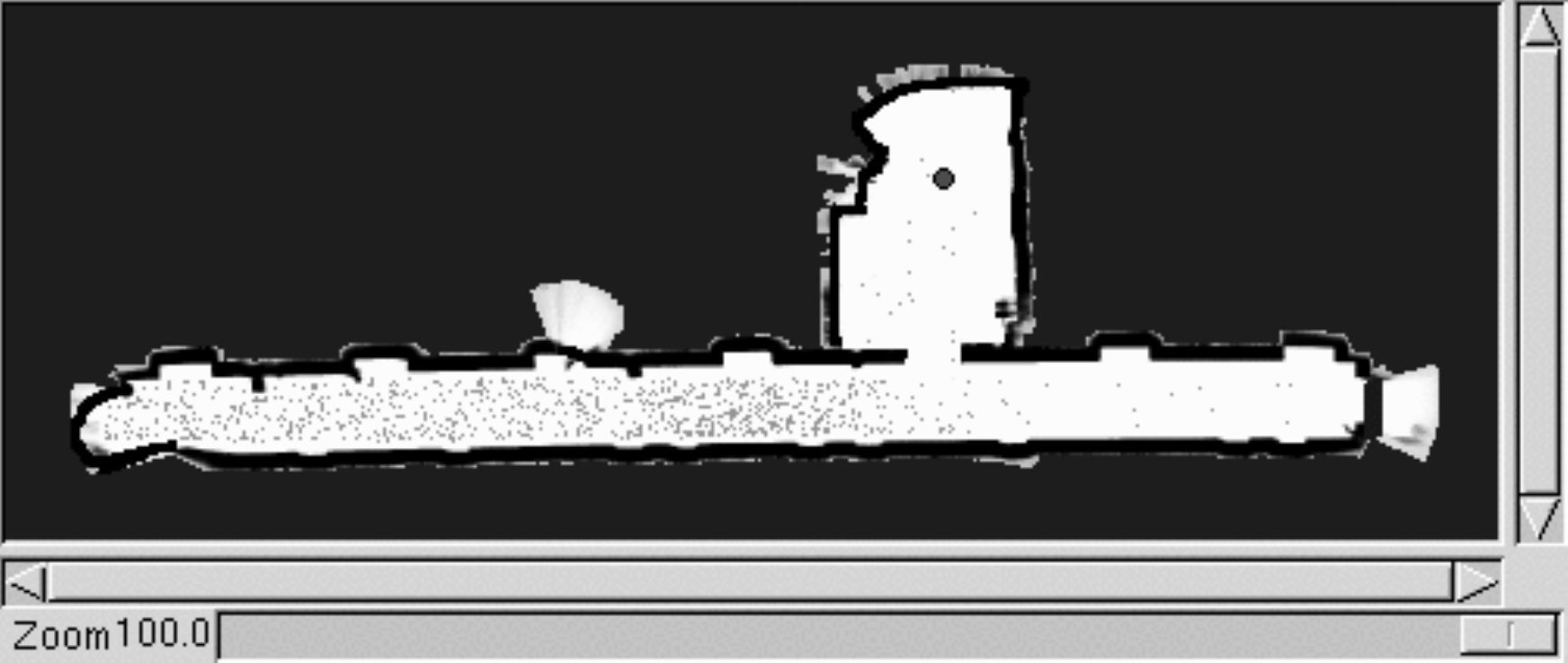}}
\hspace{4cm}
\subfigure{(d) t=17}
\hspace{1cm}
\subfigure{\includegraphics[trim=0 40 25 5,clip=true,height=3.1cm]{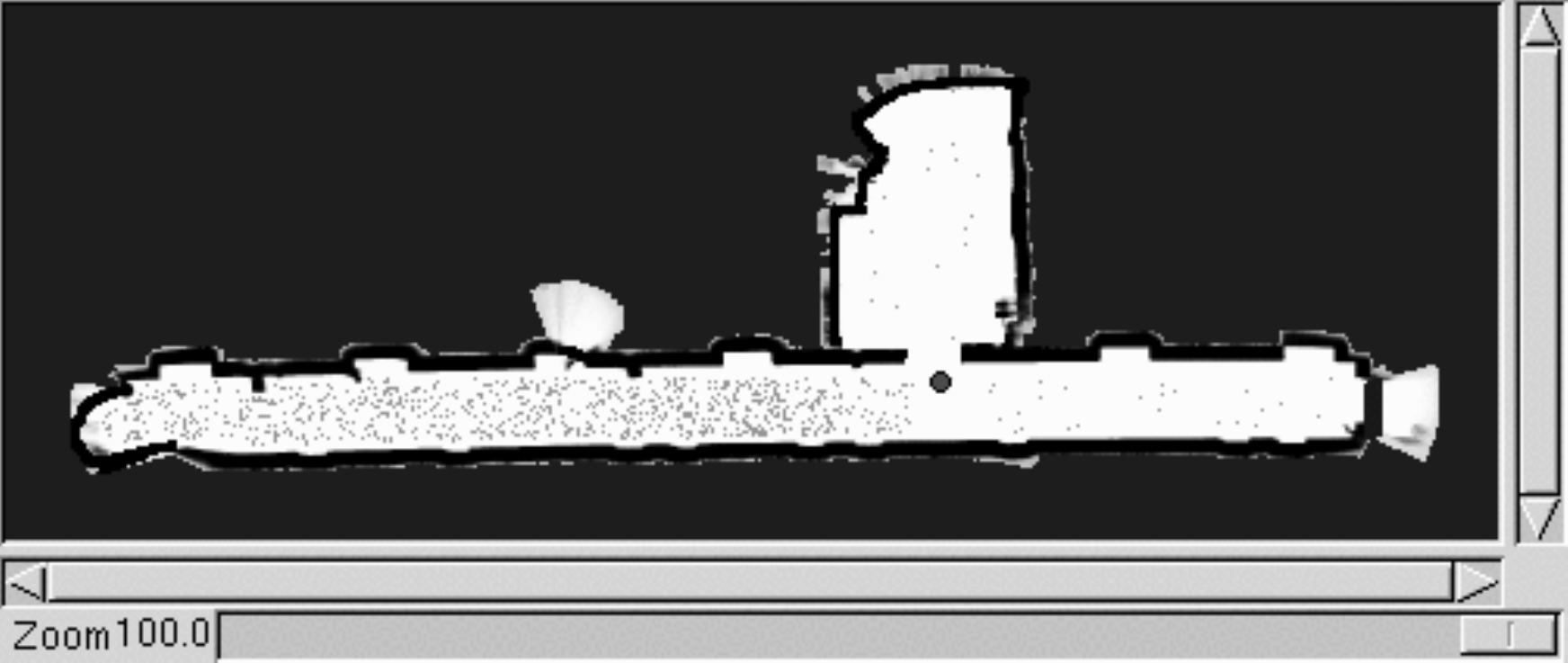}}
\hspace{4cm}
\subfigure{(e) t=29}
\hspace{1cm}
\subfigure{\includegraphics[trim=0 40 25 5,clip=true,height=3.1cm]{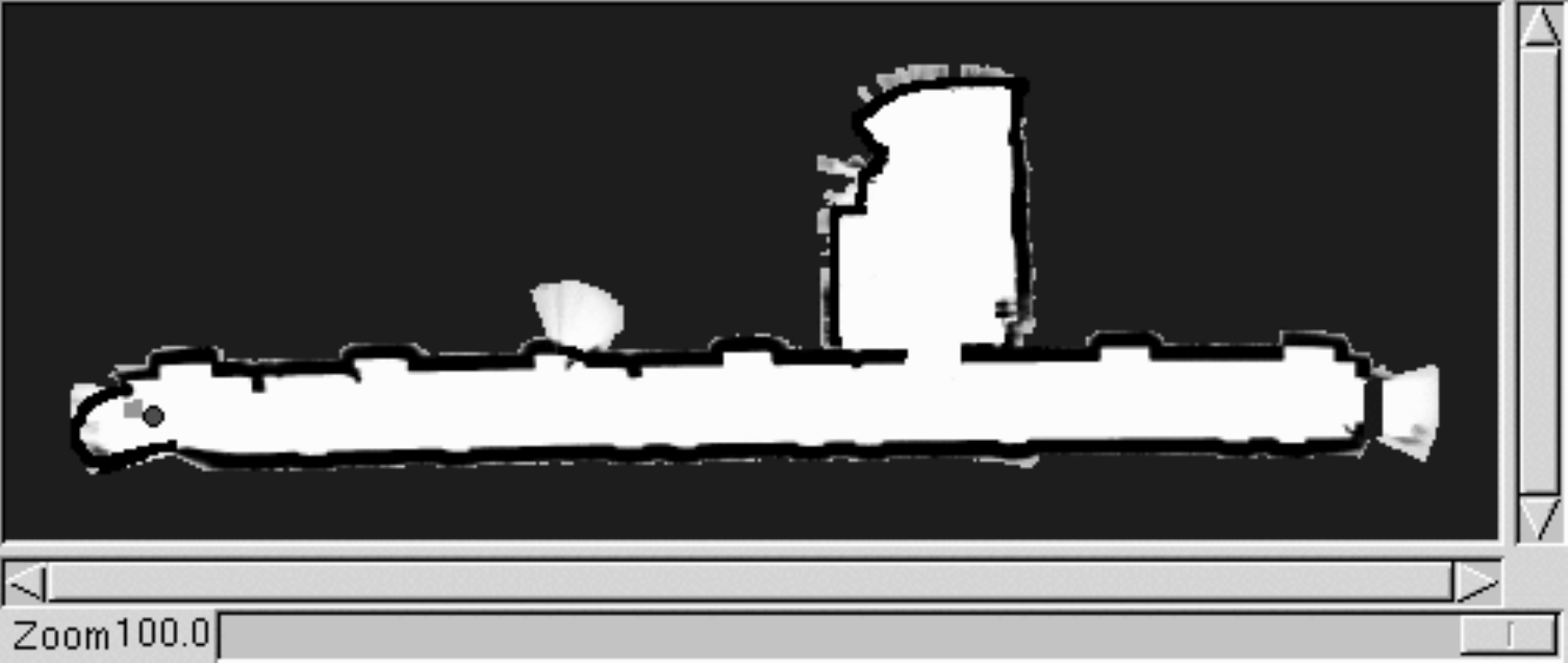}}
\caption{Example of a PBVI policy successfully finding the person}
\label{fig_s13-catch}
\end{figure}

The resulting policy is illustrated in Figure~\ref{fig_s13-catch}. This figure shows five snapshots obtained from a single run. In this particular scenario, the person starts at the far end of the left corridor. The person's location is not shown in any of the figures since it is not observable by the robot.  The figure instead shows the \textit{belief} over person positions, represented by a distribution of point samples (grey dots in Fig.~\ref{fig_s13-catch}). Each point represents a plausible hypothesis about the person's position. The figure shows the robot starting at the far right end of the corridor (Fig.~\ref{fig_s13-catch}a). The robot moves toward the left until the room's entrance (Fig.~\ref{fig_s13-catch}b). It then proceeds to check the entire room (Fig.~\ref{fig_s13-catch}c). Once relatively certain that the person is nowhere to be found, it exits the room (Fig.~\ref{fig_s13-catch}d), and moves down the left branch of the corridor, where it finally finds the person at the very end of the corridor (Fig.~\ref{fig_s13-catch}e).

This policy is optimized for any start position (for both the person and the robot). The scenario shown in Figure~\ref{fig_s13-catch} is one of the longer execution traces since the robot ends up searching the entire environment before finding the person. It is interesting to compare the choice of action between snapshots (b) and (d). The robot position in both is practically identical. Yet in (b) the robot chooses to go up into the room, whereas in (d) the robot chooses to move toward the left. This is a direct result of planning over \textit{beliefs}, rather than over \textit{states}.  The belief distribution over person positions is clearly different between those two cases, and therefore the policy specifies a very different course of action.

\begin{figure}[!ht]
\centering
\subfigure{(a) t=1}
\hspace{1cm}
\subfigure{\includegraphics[trim=0 40 25 5,clip=true,height=3.1cm]{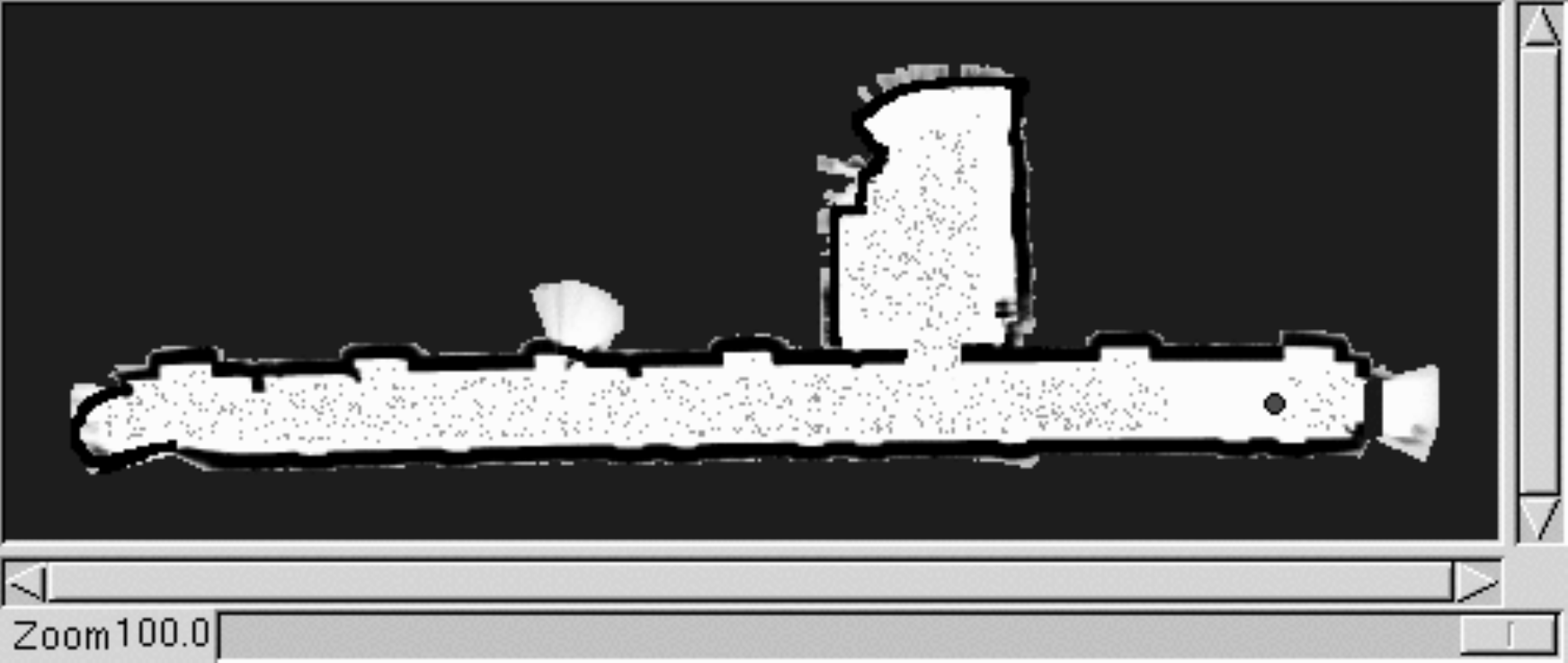}}
\hspace{4cm}
\subfigure{(b) t=7}
\hspace{1cm}
\subfigure{\includegraphics[trim=0 40 25 5,clip=true,height=3.1cm]{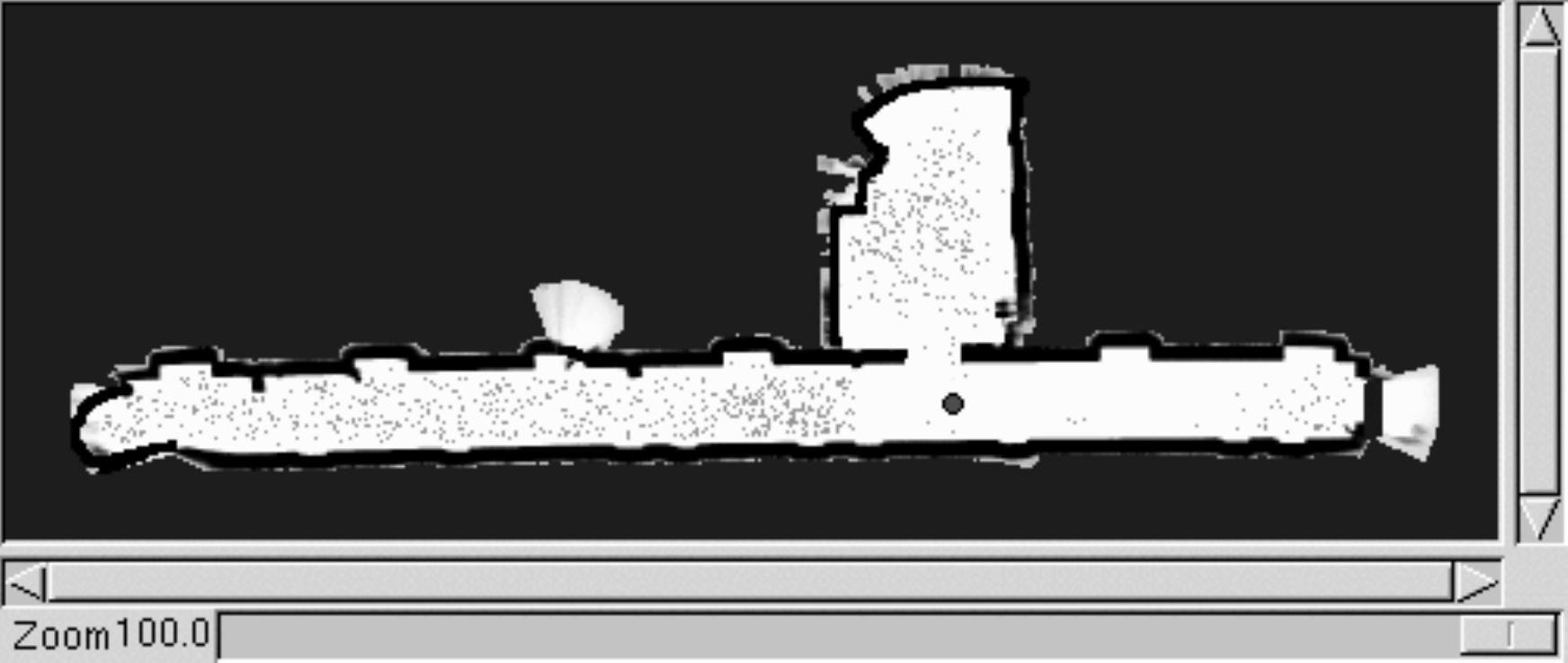}}
\hspace{4cm}
\subfigure{(c) t=17}
\hspace{1cm}
\subfigure{\includegraphics[trim=0 40 25 5,clip=true,height=3.1cm]{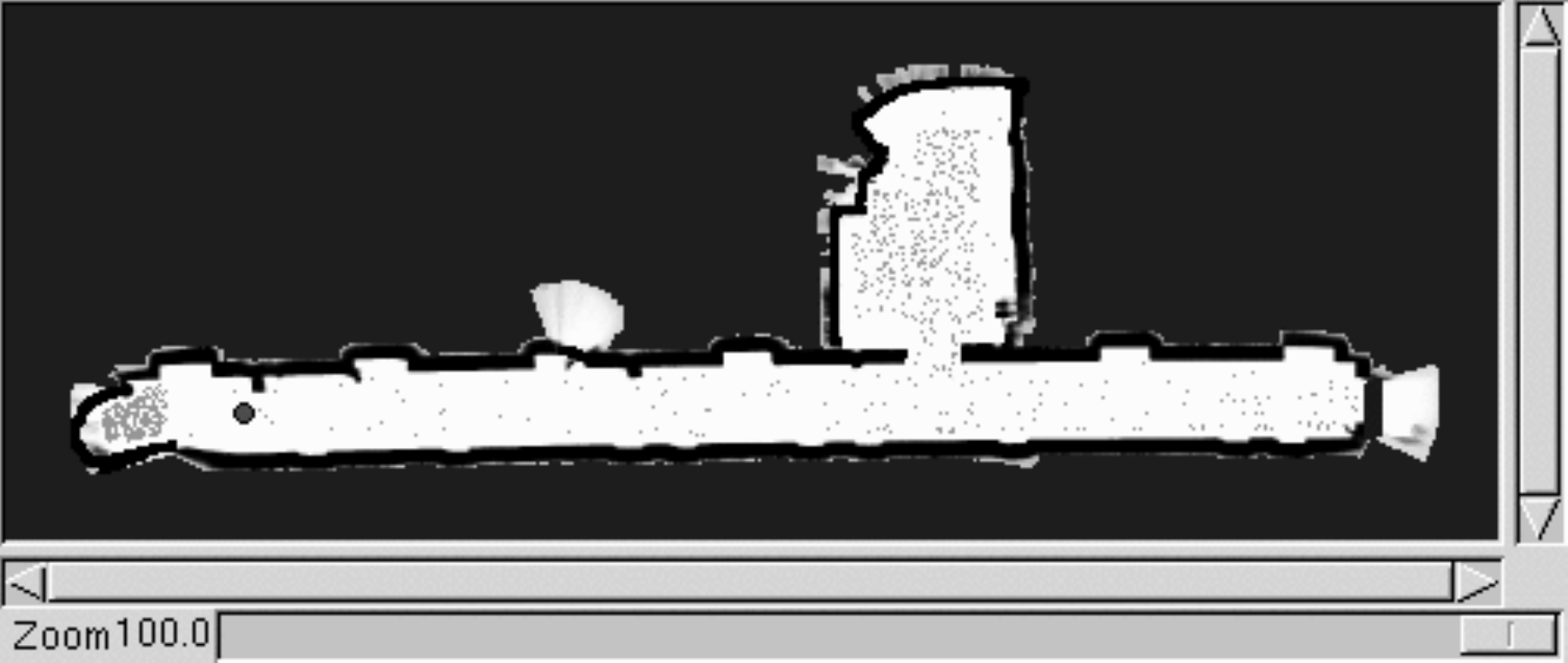}}
\hspace{4cm}
\subfigure{(d) t=27}
\hspace{1cm}
\subfigure{\includegraphics[trim=0 40 25 5,clip=true,height=3.1cm]{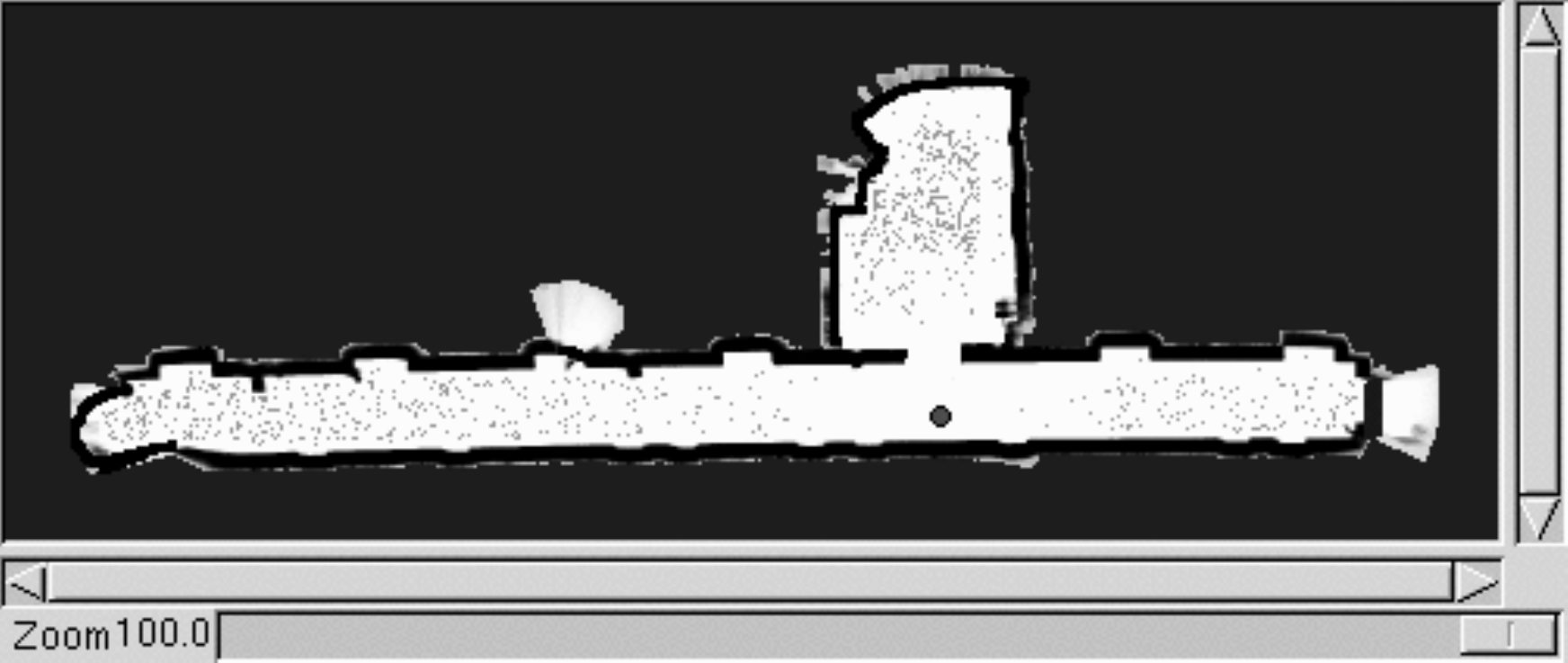}}
\caption{Example of a QMDP policy failing to find the person}
\label{fig_qmdp-miss}
\end{figure}

Figure~\ref{fig_qmdp-miss} looks at the policy obtained when solving this same problem using the QMDP heuristic. Four snapshots are offered from different stages of a specific scenario, assuming the person started on the far left side and the robot on the far right side (Fig.~\ref{fig_qmdp-miss}a).  After proceeding to the room entrance (Fig.~\ref{fig_qmdp-miss}b), the robot continues down the corridor until it \textit{almost} reaches the end (Fig.~\ref{fig_qmdp-miss}c). It then turns around and comes back toward the room entrance, where it stations itself (Fig.~\ref{fig_qmdp-miss}d) until the scenario is forcibly terminated.  As a result, the robot cannot find the person when s/he is at the left edge of the corridor or in the room. What's more, because of the running-away behavior adopted by the subject, even when the person starts elsewhere in the corridor, as the robot approaches the person will gradually retreat to the left and similarly escape from the robot.

Even though QMDP does not explicitly plan over \textit{beliefs}, it can generate different policy actions for cases where the state is identical but the belief is different. This is seen when comparing Figure~\ref{fig_qmdp-miss} (b) and (d).  In both of these, the robot is identically located, however the belief over person positions is different. In (b), most of the probability mass is to the left of the robot, therefore it travels in that direction. In (d), the probability mass is distributed evenly between the three branches (left corridor, room, right corridor). The robot is equally pulled in all directions and therefore stops there. This scenario illustrates some of the strength of QMDP\@. Namely, there are many cases where it is not necessary to explicitly reduce uncertainty. However, it also shows that more sophisticated approaches are needed to handle some cases.

These results show that PBVI can perform outside the bounds of simple maze domains, and is able to handle realistic problem domains. In particular, throughout this evaluation, the robot simulator was in no way constrained to behave as described in our POMDP model (Sec.~\ref{sec_robot_model}). This means that the robot's actions often had stochastic effects, the robot's position was not always fully observable, and that belief tracking had to be performed asynchronously (i.e., not always a straightforward ordering of actions and observations). Despite this misalignment between the model assumed for planning, and the execution environment, the control policy optimized by PBVI could successfully be used to complete the task.

\clearpage

\clearpage

\section{Discussion}
\label{sec_contributions}

This paper describes a class of anytime point-based POMDP algorithms called PBVI, which combines point-based value updates with strategic selection of belief points, to solve large POMDPs. Further extensions to the PBVI framework, whereby value updates are applied to groups of belief points according to their spatial distribution, are described by~\cite{pineau04a}.  The main contributions pertaining to the PBVI framework are now summarized.

\textbf{Scalability}. The PBVI framework is an important step towards truly scalable POMDP solutions. This is achieved by bounding the policy size through the selection of a small set of belief points.

\textbf{Anytime planning}. PBVI-class algorithms alternates between steps of value updating and steps of belief point selection. As new points are added, the solution improves, at the expense of increased computational time. The trade-off can be controlled by adjusting the number of points.  The algorithm can be terminated either when a satisfactory solution is found, or when planning time is elapsed.

\textbf{Bounded error}. We provide a theoretical bound on the error of the approximation introduced in the PBVI framework. This result holds for a range of belief point selection methods, and lead directly to the development of a new PBVI-type algorithm: PBVI+GER, where estimates of the error bound are used directly to select belief points. Furthermore we find that the bounds can be useful in assessing when to stop adding belief points.

\textbf{Exploration}. We proposed a set of new point selection heuristics, which explore the tree of reachable beliefs to select useful belief points. The most successful technique described, Greedy Error Reduction (GER), uses an estimate of the error bound on candidate belief points to select the most useful points.

\textbf{Improved empirical performance}. PBVI has demonstrated the ability to reduce planning time for a number of well-known POMDP problems, including Tiger-grid, Hallway, and Hallway2. By operating on a set of discrete points, PBVI algorithms can perform polynomial-time value updates, thereby overcoming the curse of history that paralyzes exact algorithms.  The GER technique used to select points allows us to solve large problems with fewer belief points than alternative approaches.

\textbf{New problem domain}. PBVI was applied to a new POMDP planning domain (Tag), for which it generated an approximate solution that outperformed baseline algorithms QMDP and Incremental Pruning.  This new domain has since been adopted as a test case for other algorithms~\cite{vlassis04,smith04,braziunas04,poupart03}. This fosters an increased ease of comparison between new techniques. Further comparative analysis was provided in Section~\ref{sec_robot_compare} highlighting similarities and differences between PBVI and Perseus.

\textbf{Demonstrated performance}. PBVI was applied in the context of a robotic search-and-rescue type scenario, where a mobile robot is required to search its environment and find a non-stationary individual. PBVI's performance was evaluated using a realistic, independently-developed, robot simulator.

A significant portion of this paper is dedicated to a thorough comparative analysis of point-based methods.  This includes evaluating a range of point-based selection methods, as well as evaluating mechanisms for ordering value updates.  The comparison of point-based selection techniques suggest that the GER method presented in Section~\ref{sec_belief_ger} is superior to more naive techniques. In terms of ordering of value updates, the randomized strategy which is used in the Perseus algorithm appears effective to accelerate planning. A natural next step would be to combine the GER belief selection heuristic with Perseus's random value updates. We performed experiments along these lines, but did not achieve any significant speed-up over the current performance of PBVI or Perseus (e.g., as reported in Figure~\ref{fig_pbvi-perseus}(a)). It is likely that when belief points are chosen carefully (as in GER), each of these points needs to be updated systematically and therefore there is no additional benefit to using randomized value updates.

Looking towards the future, it is important to remember that while we have demonstrated the ability to solve problems which are large by POMDP standards, many real-world domains far exceed the complex of domains considered in this paper. In particular, it is not unusual for a problem to be expressed through a number of multi-valued state \textit{features}, in which case the number of states grows exponentially with the number of features. This is of concern because each belief point and each $\alpha$-vector has dimensionality $|S|$ (where $|S|$ is the number of states) and all dimensions are updated simultaneously. This is an important issue to address to improve the scalability of point-based value approaches in general.

There are various existing attempts at overcoming the curse of dimensionality in POMDPs.
Some of these---e.~g. the belief compression techniques by~\citeA{roy02a}---cannot be incorporated within the PBVI framework without compromising its theoretical properties (as discussed in Section~\ref{sec_pbvi}). Others, in particular the exact compression algorithm by~\citeA{poupart02}, can be combined with PBVI\@. However, preliminary experiments in this direction have yielded little performance improvement. There is reason to believe that approximate value compression would yield better results, but again at the expense of forgoing PBVI's theoretical properties. The challenge therefore is to devise function-approximation techniques that both reduce the dimensionality effectively, while maintaining the convexity properties of the solution.

A secondary (but no less important) issue concerning the scalability of PBVI pertains to the number of belief points necessary to obtain a good solution.  While problems addressed thus far can usually be solved with $O(|S|)$ belief points, this need not be true. In the worse case, the number of belief points necessary may be exponential in the plan length. The PBVI framework can accommodate a wide variety of strategies for generating belief points, and the Greedy Error Reduction technique seems particularly effective. However this is unlikely to be the definitive answer to belief point selection. In more general terms, this relates closely to the well-known issue of exploration versus exploitation, which arises across a wide array of problem-solving techniques.

These promising opportunities for future research aside, the PBVI framework has already
pushed the envelope of POMDP problems that can be solved with existing
computational resources. As the field of POMDPs matures, finding ways
of computing policies efficiently will likely continue to be a major
bottleneck. We hope that point based algorithms such as the PBVI
will play a leading role in the search for more efficient algorithms.

\section*{Acknowledgments}
The authors wish to thank Craig Boutilier, Michael Littman, Andrew
Moore and Matthew Mason for many thoughtful comments and discussions
regarding this work. We also thank Darius Braziunas, Pascal Poupart,
Trey Smith and Nikos Vlassis, for conversations regarding their
algorithms and results. The contributions of Michael Montemerlo and
Nicholas Roy in conducting the empirical robot evaluations are
gratefully acknowledged. Finally, we thank three anonymous reviewers and one dedicated editor (Sridhar Mahadevan) whose
feedback significantly improved the paper. This work was funded
through the DARPA MARS program and NSF's ITR program (Project:
"Robotic Assistants for the Elderly", PI: J. Dunbar-Jacob).

\bibliographystyle{theapa}
\bibliography{../papers}

\end{document}